\documentclass{article}
\usepackage[affil-it]{authblk}
\usepackage{bookmark}
\usepackage{geometry}
\usepackage[toc,page]{appendix}
\usepackage{graphicx} 
\usepackage{amsmath}
\usepackage{amssymb}
\usepackage{amsthm}
\title{Acoustic Waveform Inversion with Image-to-Image Schrödinger Bridges}
\author{A.S. Stankevich\thanks{\texttt{stankevich.as@phystech.edu}; Corresponding Author} \,, I.B. Petrov} 
\affil{Moscow Institute of Physics and Technology, info@mipt.ru}
\date{April 30, 2025}

\usepackage{booktabs, tabularx}
\usepackage{multirow}   
\newcolumntype{Y}{>{\raggedright\arraybackslash}X} 
\newcolumntype{C}[1]{>{\centering\let\newline\\\arraybackslash\hspace{0pt}}m{#1}}

\usepackage{hyperref}

\usepackage{cleveref}

\usepackage{algorithm}
\usepackage{algpseudocode}
\usepackage{mathtools}

\usepackage[font=scriptsize]{caption}
\usepackage{subcaption} 

\usepackage[
    style=ieee,
    sorting=none,
    bibencoding=utf8,
    backend=biber,
    autolang=other,
    date=year,
    url=false,
    doi=false,
    isbn=false
]{biblatex}
\usepackage{csquotes}
\nocite{*}
\usepackage{siunitx}
\newcommand{\norm}[1]{\left\lVert#1\right\rVert}
\newcommand*\Laplace{\mathop{}\!\mathbin\bigtriangleup}
\newcommand{\vect}[1]{\boldsymbol{\mathbf{#1}}}
\DeclareMathOperator*{\argmin}{arg\,min}

\theoremstyle{plain}
\newtheorem*{proposition}{Proposition}

\begin{document}

\maketitle

\begin{abstract}

    Recent developments in application of deep learning models to acoustic Full Waveform Inversion (FWI) 
    are marked by the use of diffusion models as prior distributions for 
    Bayesian-like inference procedures. The advantage of these methods is the ability to generate high-resolution samples, 
    which are otherwise unattainable with classical inversion methods or other deep learning-based 
    solutions. However, the iterative and stochastic nature of sampling from diffusion models along with 
    heuristic nature of output control remain limiting factors for their applicability
    For instance, an optimal way to include the approximate velocity model into diffusion-based inversion scheme remains unclear,
    even though it is considered an essential part of FWI pipeline. 
    We address the issue by employing a Schrödinger Bridge that interpolates between the distributions 
    of ground truth and smoothed velocity models. Thus, the inference process that starts from
    an approximate velocity model is guaranteed to arrive at a sample from the distribution of reference velocity models 
    in a finite time. To facilitate the learning of nonlinear drifts that transfer 
    samples between distributions and to enable controlled inference given the seismic data, 
    we extend the concept of Image-to-Image Schrödinger Bridge (\(\text{I}^2\text{SB}\)) to conditional sampling, 
    resulting in a conditional Image-to-Image Schrödinger Bridge (c\(\text{I}^2\text{SB}\)) framework for acoustic inversion.  
    To validate our method, we assess its effectiveness in reconstructing the reference 
    velocity model from its smoothed approximation, coupled with the observed seismic signal of fixed shape. 
    Our experiments demonstrate that the proposed solution outperforms our reimplementation of 
    conditional diffusion model suggested in earlier works, while requiring only a few neural function evaluations (NFEs) 
    to achieve sample fidelity superior to that attained with supervised learning-based approach. 
    The supplementary code implementing the algorithms described in this paper can be found in 
    the repository \url{https://github.com/stankevich-mipt/seismic_inversion_via_I2SB}

\end{abstract}

{
\section{Introduction}

The general purpose of seismic inversion procedures is to reveal the properties of the subsurface medium given results from non-destructive 
scanning \cite{russell1988introduction}. Typical inverse problems encountered in seismology are ill-posed and underconstrained, as the data coverage 
is fundamentally limited due to the nature of the scanning setup. Thus, they are usually solved numerically using various physics-based approximations of 
continuous media and regularization techniques that guarantee the existense of a solution (\cite{schuster2017seismic, robertsson2007introduction}).

A significant part of seismic inversion studies is performed in the acoustic approximation. 
The dynamics of the acoustic medium is governed by a second-order partial-differential equation (PDE) parameterised by a single function of 
acoustic velocity \cite{alford1974accuracy}. As such, the process of seismic scanning is formulated as an initial-boundary value problem for the acoustic PDE. 
Inverting the data of acoustic seismic scanning involves recovering the spatial distribution of the medium's velocity using the waveform data 
recorded at the location of signal receivers.

Recent advances in signal processing with deep learning algorithms have encouraged researchers to actively investigate data-driven approaches 
to seismic inversion \cite{adler2021deep, mousavi2024applications}. A common strategy for supervised frameworks is 
to employ convolutional neural networks that take either preprocessed (\cite{araya2018deep, zhang2020adjoint, zhang2021deep}) or raw 
(\cite{wu2019inversionnet, yang2019deep, zhang2020data}) waveform data as input and map it directly to corresponding velocity images  
The training data for such models in the majority of cases is handcrafted -- finite-difference computational engines emulate the 
dynamics of acoustic media parameterised by manually generated velocity fields, obtaining matching seismogram-impedance image pairs.

Diffusion models (\cite{ho2020ddpm, yang2023diffusion}) have shown themselves as a promising tool for solving imaging inverse problems. 
Namely, the diffusion model generation process could be altered in order to perform posterior sampling 
conditioned on measurements using Bayesian inference \cite{song2020score-based-modelling-SDE}. This aforementioned feature requires 
evaluation of the conditional score, a task that can be done either in zero-shot fashion by imposing a statistical hypothesis 
on the structure of observed data (\cite{li2022srdiff, lugmayr2022repaint, kawar2022denoising}), or by internal means of deep learning models 
(\cite{ho2022classifierfreediffusionguidance, rombach2022highresolutionimagesynthesislatent}).  
Regardless of the specifics, neural network here could be viewed as an implicit prior disentangled from the 
data acquisition model. 

To date, such insight has already been employed by several researchers on the topic of deep-learning applications 
in seismic inversion (\cite{wang2023apriorregularizedfwiusinggenerativediffusionmodels,wang2024controllable, zhang2024diffusionvel}). 
Since the procedure of waveform inversion commonly progresses from coarse to fine resolution and is known to be sensitive to
initial approximation, researchers naturally gravitate towards conditioning models on smoothed velocity models as 
auxiliary input. However, to the best of our knowledge, conditioning methods present in the literature on acoustic inversion 
with diffusion-based models are so far purely heuristic. The goal of this paper is to demonstrate a data-driven seismic 
inversion technique that incorporates smoothed velocity models into the diffusion-based workflow on a theoretical basis. 
Our contribution is as follows:
\begin{itemize}
    \item We have developed a novel acoustic waveform inversion scheme based on the problem statement 
    described in the conditional Diffusion Schrödinger Bridge paper (cDSB, \cite{shi2022conditional}).
    To facilitate the process of generative model building, we have adapted the training pipeline proposed 
    for a specific case of Schrödinger Bridge problem statement described by Liu et al 
    \cite{liu2023i2sbimagetoimageschrodingerbridge} towards conditional simulation, 
    resulting in a conditional Image-to-Image Schrödinger Bridge (c\(\text{I}^2\text{SB}\))
    \item We have designed an end-to-end training procedure similar to classifier-free diffusion guidance 
    \cite{ho2022classifierfreediffusionguidance} that allows our model to learn nonlinear drifts for both conditional and unconditional 
    settings within the scope of the same framework (fig. \ref{fig:acoustic_seismic_inversion_with_i2sb_pipeline}).  
    We have demonstrated that model trained in this way achieves a sample diversity/fidelity tradeoff, which can be controlled through 
    a mixing weight parameter.   
    \item We have assessed the performance of the proposed scheme using the OpenFWI data collection 
    \cite{DBLP:journals/corr/abs-2111-02926}. 
    Our experiments have shown that the developed method outperforms several solutions introduced in previous works on the subject.

\end{itemize}

\begin{figure}[h!]

    \begin{minipage}[t]{0.45\textwidth}\vspace{0pt}
        \centering
        \caption*{c\(\text{I}^2\text{SB}\) training}
        \vspace{0.6cm}
        \includegraphics[width=1.2\textwidth]{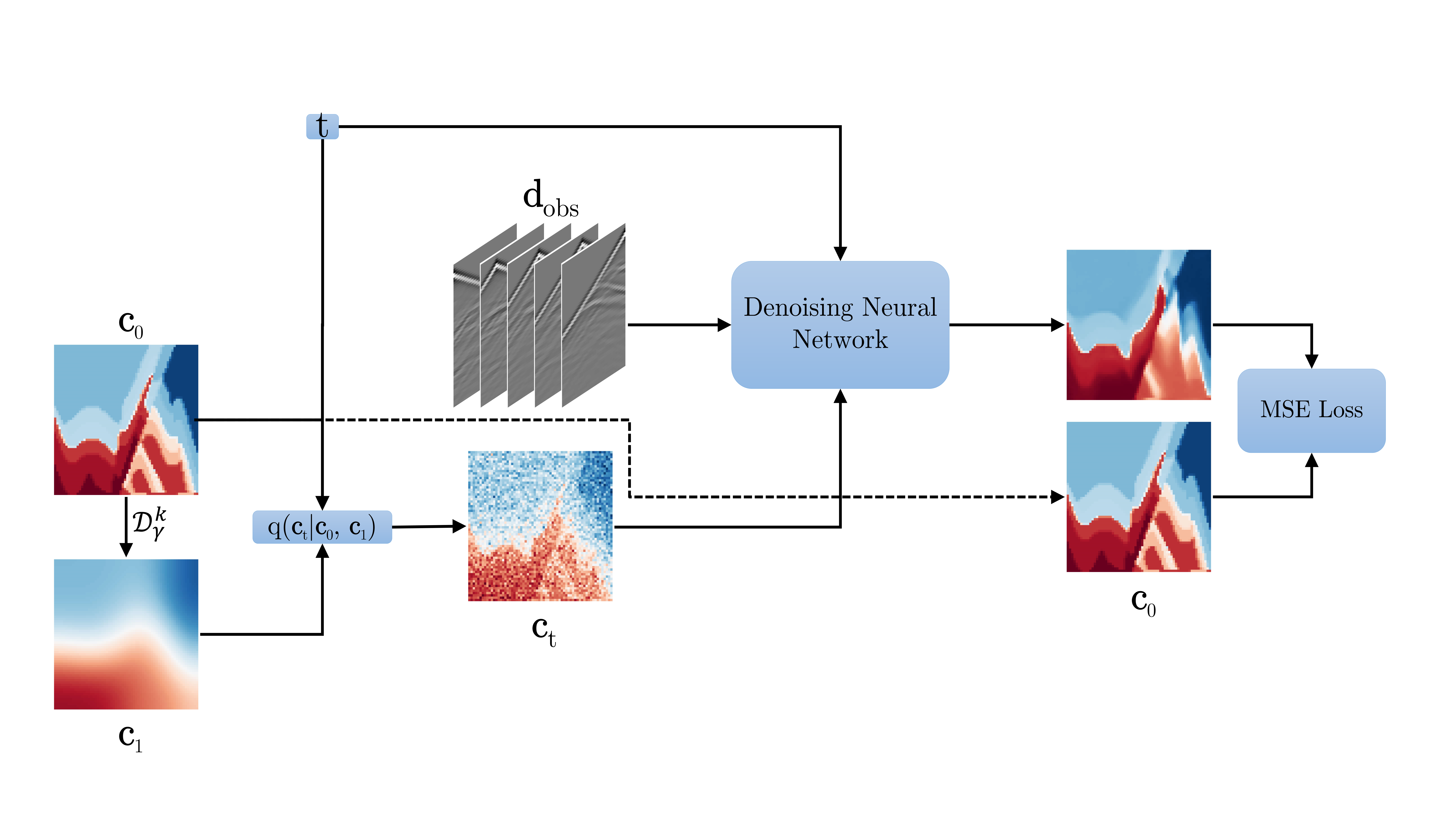}        
        \label{fig:i2sb_training_scheme}
    \end{minipage}
    \hfill
    \begin{minipage}[t]{0.45\textwidth}\vspace{0pt}
        \centering
        \caption*{c\(\text{I}^2\text{SB}\) inference}
        \includegraphics[width=1.2\textwidth]{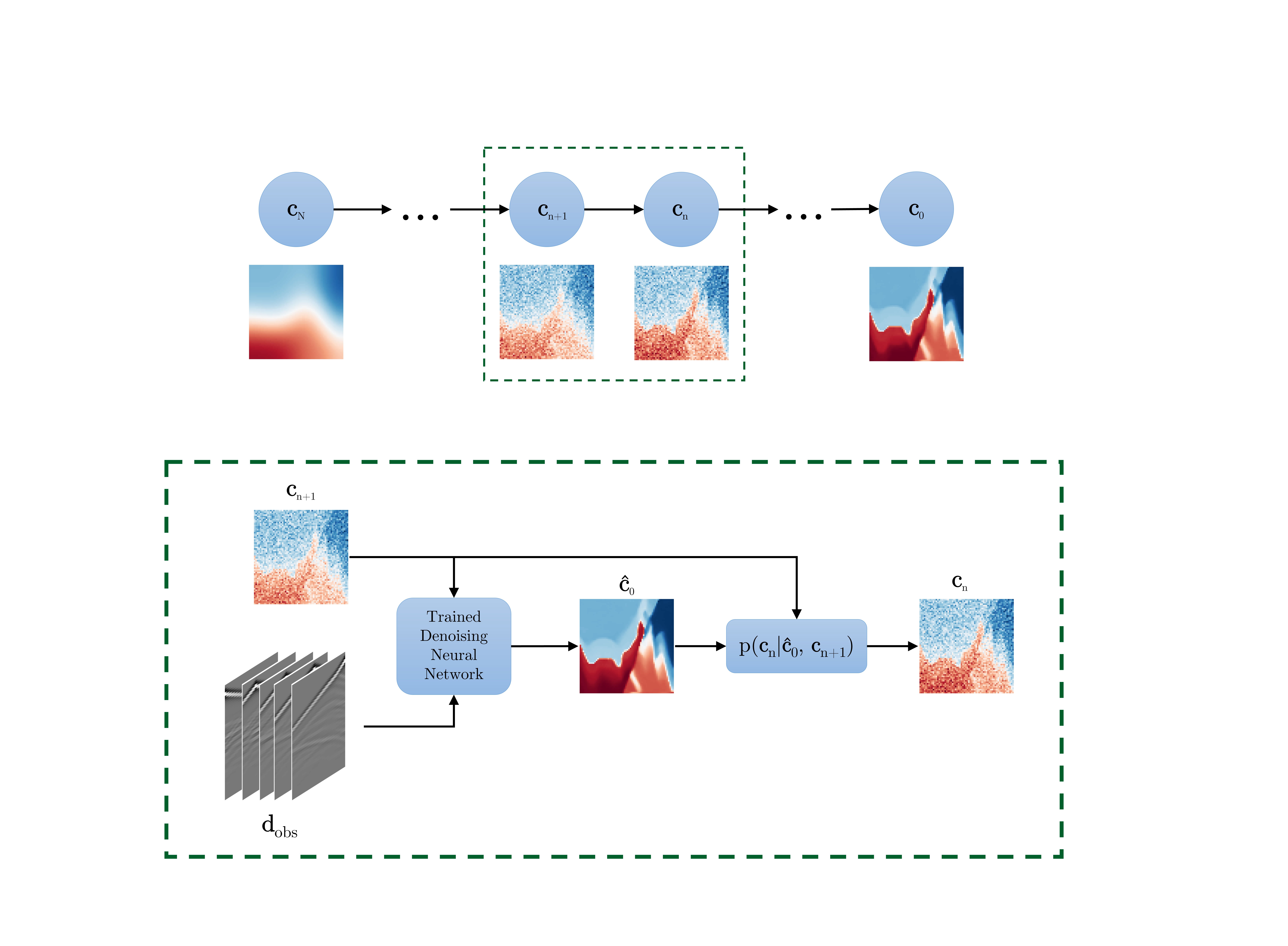}
        \label{fig:i2sb_inference_scheme}
    \end{minipage}
    
    \caption{Conditional Image-to-Image Schrödinger Bridge in application to acoustic seismic inversion}
    \label{fig:acoustic_seismic_inversion_with_i2sb_pipeline}
\end{figure}
}

\section{Background}

\subsection{Acoustic Waveform Inversion with Deep Learning}

Consider a continuous medium governed by the 2D acoustic wave equation
\begin{equation} \label{eq:2d_acoustic_PDEs}
    \Laplace{\vect{p}} - \dfrac{1}{\vect{c}_0^2} \dfrac{\partial^2 \vect{p}}{\partial t^2} = \vect{s}
\end{equation}
Here \(\Laplace{}\) is the 2D Laplace operator, \(\vect{p}(x,y,t) \) is the pressure field, 
\( \vect{s}(x,y,t)\) is the source function, and \(\vect{c}_0(x,y)\) is the velocity field. 
The conventional regularized full waveform inversion (FWI, \cite{virieux2009overview}) problem 
statement is formulated as an optimization task  
\begin{equation} \label{eq:acoustic_fwi_problem_statement}
    \min\limits_{\vect{c}_0} \, J(\vect{d}_{\text{model}},  \vect{d}_{\text{obs}}) + \lambda R(\vect{c}_0)
    \quad \text{s.t.} \quad \vect{d}_{\text{model}} = F(\vect{c}_0) 
\end{equation}
where J measures the discrepancy between modelled $\vect{d}_{\text{model}}$ 
and observed $\vect{d}_{\text{obs}}$ seismic data, \(F\) is the forward modelling operator, 
and $R(\vect{c}_0)$ is the regularization term scaled by the $\lambda$ multiplier. 
The purpose of regularization is to ensure the existence of a solution.
In other words, the goal of full waveform inversion 
is to estimate the value of ''pseudoinverse'' of \(F\) applied to the observed data 
\begin{equation} \label{eq:acoustic_fwi_problem_statement_pseudoinverse}
    \vect{c}_0^* = \hat{F}^{-1} (\vect{d}_{\text{obs}})
\end{equation}

Supervised learning-based approaches to acoustic waveform inversion seek $\hat{F}^{-1}_{\vect{\theta}}$ -- 
a parametric approximation of $\hat{F}^{-1}$. The implementation of $\hat{F}^{-1}_{\vect{\theta}}$ commonly 
involves convolutional neural networks, as both seismograms and 2D velocity models resemble multichanneled images. 
The tuning of parameters $\vect{\theta}$ is carried 
out using gradient optimization on a training dataset, which contains paired instances of velocity models 
$\vect{c}^i_{\text{train}}$ and corresponding observed data $\vect{d}^i_{\text{train}}$.
Once the parameter fitting is done, $\vect{c}^i_{\text{test}} = \hat{F}^{-1}_{\vect{\theta}^*} \left( \vect{d}^i_{\text{test}} \right)$ 
yields the reconstructed velocity model for seismogram $\vect{d}^i_{\text{test}}$.

\subsection{Score-based Generative Modelling (SGM)}

SGM \cite{song2020score-based-modelling-SDE} is a framework for generative modelling which considers the morphing 
process of probability distributions from the perspective of solutions to stochastic differential equations (SDEs).
The main object of interest for SGM is a forward SDE of the form
\begin{equation} \label{eq:forward_sde_general_form}
    d \mathbf{x} = \mathbf{f} \left( \mathbf{x}, t\right) dt + \sqrt{\beta_t} d \mathbf{w}, 
    \quad \mathbf{x} \in \mathbb{R}^d, \quad t \in [0, 1]
\end{equation}
With the proper choice of \(\beta_t\) and \(f_t\) - the latter has to be linear on $\mathbf{x}$ - the terminal 
distributions of noising equation approaches standard Gaussian regardless of the initial distribution, i.e., 
\(\vect{x}_1 \sim \mathcal{N} (0, 1) \, \forall \vect{x}_0 \sim p_{\text{data}}\). A notable property of the equation 
\eqref{eq:forward_sde_general_form} is the existence of its time-reversal \cite{anderson1982reverse}, defined by
\begin{equation} \label{eq:reverse_sde_general_form}
    d \mathbf{x} = -\left[
        \mathbf{f} \left( \mathbf{x}, t\right) - \beta(t) \nabla_{\mathbf{x}_t} \log p_t \left( \mathbf{x}_t\right)
    \right]dt + \sqrt{\beta(t)} d \overline{\mathbf{w}}
\end{equation}
Forward and backward SDEs share the same marginal probability densities, since path 
measures induced by equations \labelcref{eq:forward_sde_general_form,eq:reverse_sde_general_form} 
are equal almost surely. Thus, it is possible to construct generative process by means of  
denoising score matching \cite{vincent2011connection}
\begin{equation}\label{eq:denoising_score_matching}
    \min\limits_{\vect{\theta}} \norm{s_{\vect{\theta}} \left(\vect{x}_t, t \right) - 
    \sigma_t \nabla \log p (\vect{x}_t, t| \vect{x}_0)}
\end{equation} 
Here the output of parametric model $s_{\vect{\theta}} \left(\vect{x}_t, t \right)$ is 
is regressed towards the score $\nabla_{\vect{x}_t}  \log p (\vect{x}_t, t| \vect{x}_0)$ 
of the forward process \eqref{eq:forward_sde_general_form},
which can be effectively computed at runtime. Once trained, the score estimator can be substituted 
into the discretization of \eqref{eq:reverse_sde_general_form} in place of the unknown term 
$\nabla_{\mathbf{x}_t} \log p_t \left( \mathbf{x}_t\right)$, resulting in recursive procedure for 
sampling from the considered distribution.   

\subsection{Schrödinger Bridge (SB)}

Dynamic SB expression (\cite{schrodinger1932theorie, pavon1991free, leonard2013survey}) describes
an entropy-regularized optimal transport problem for two different distributions $p_{\text{data}}$ and 
$p_{\text{prior}}$ on a finite time interval $t \in [0, T]$
\begin{equation} \label{eq:general_dynamic_sb_problem_statement}
    \min\limits_{\mathbb{Q} \in \mathcal{P} \left(p_{\text{data}}, p_{\text{prior}} \right)}
    D_{\text{KL}} \left( \mathbb{Q} || \mathbb{P}\right)
\end{equation}
In the statement above $\mathbb{Q}$ belongs to a set of path measures having $p_{\text{data}}$ and $p_{\text{prior}}$ as 
its marginal densities, and $\mathbb{P}$ is a reference measure. If $\mathbb{P}$ is a 
path measure associated with stochastic process \eqref{eq:forward_sde_general_form}, the optimality condition for
\eqref{eq:general_dynamic_sb_problem_statement} is provided by a set of PDEs coupled 
through their boundary conditions \cite{chen2021stochastic, chen2021likelihood}. To wit, let 
$\Psi (\vect{x}_t, t), \hat{\Psi} (\vect{x}_t, t) \in \, C^{2, 1} (\mathbb{R}^d, [0, T])$
be the solution of the following system of equations
\begin{subequations} \label{eq:kfp_equations_sb}
    \begin{equation} \label{eq:kfp_forward_equation_sb}
        \dfrac{\partial \Psi (\vect{x}, t)}{\partial t} = 
        -\nabla_{\vect{x}} \Psi^{T} \vect{f} - \beta_t \dfrac{1}{2} \Laplace \Psi 
    \end{equation}
    \begin{equation} \label{eq:kfp_backward_equation_sb}
        \dfrac{\partial \hat{\Psi} (\vect{x}, t)}{\partial t} = 
         -\nabla_{\vect{x}} \cdot (\hat{\Psi}\vect{f}) + \beta_t \dfrac{1}{2} \Laplace \hat{\Psi}    
    \end{equation}
    \begin{equation}
        \textit{s.t.} \quad \Psi (\vect{x}, 0)\hat{\Psi} (\vect{x}, 0) = p_{\text{data}} \quad  
        \Psi (\vect{x}, T)\hat{\Psi} (\vect{x}, T) = p_{\text{prior}}
    \end{equation}
\end{subequations}
In this case, the solution to optimization problem \eqref{eq:general_dynamic_sb_problem_statement} can be 
expressed by the path measure of either the forward \eqref{eq:sb_forward_sde}, or,
equivalently, the backward \eqref{eq:sb_backward_sde} SDEs 
\begin{subequations} \label{eq:sb_equations}
    \begin{equation} \label{eq:sb_forward_sde}
        d \vect{x}_t = \left[\vect{f}_t + \beta_t \nabla_{\vect{x}_t} \log \Psi (\vect{x}_t, t) \right] dt + \sqrt{\beta_t} 
        d\vect{w}_t 
    \end{equation}
    \begin{equation} \label{eq:sb_backward_sde} 
        d \vect{x}_t = \left[\vect{f}_t - \beta_t \nabla_{\vect{x}_t} \log \hat{\Psi} (\vect{x}_t, t) \right] dt + \sqrt{\beta_t} 
        d\overline{\vect{w}}_t 
    \end{equation}
\end{subequations} 
Similarly to SGM, path measures induced by stochastic processes \eqref{eq:sb_forward_sde} and \eqref{eq:sb_backward_sde} 
are equal almost surely. Equations \eqref{eq:sb_equations} can be viewed as a nonlinear generalization of 
forward and backward SGM equations \eqref{eq:forward_sde_general_form} \eqref{eq:reverse_sde_general_form}, 
since the inclusion of nonlinear drift allows diffusion to transfer samples beyond Gaussian priors on a finite time horizon. 

\subsection{Conditional Diffusion Schrödinger Bridge}

Consider the discrete equivalent of noising process \eqref{eq:forward_sde_general_form} represented by 
a Markov chain $\vect{x}_{0, N} = \left\{\vect{x}_k \right\}_{k=0}^N \in \mathcal{X} = (\mathbb{R}^d)^{N+1}$
with joint density 
\begin{equation} \label{eq:noising_markov_chain_dsb}
    p(\vect{x}_{0:N}) = p_0 (\vect{x}_0) \prod_{k=0}^{N-1} p_{k+1|k}(\vect{x}_{k+1}|\vect{x}_{k})
\end{equation}
where $p_0 = p_{\text{data}}$ and $p_{k+1|k}$ are Markov transition densities, with
$p_{k+1}(\vect{x}_{k+1}) = \int p_{k+1|k} (\vect{x}_{k+1} | \vect{x}_k) p_k(\vect{x}|_k) d\vect{x}_k$.
De Bortoli et al. \cite{debortoli2021diffusion} formulated the dynamic Schrödinger Bridge problem for forward density
$p(\vect{x}_{0:N})$ of the noising chain \eqref{eq:noising_markov_chain_dsb}.
\begin{equation} \label{eq:sb_n_state_markov_chain}
    \pi^* = \argmin \limits_{\pi} \left\{
        D_{\text{KL}} (\pi || p(\vect{x}_{0:N})): \pi_0 = p_{\text{data}}, \pi_N = p_{\text{prior}}
    \right\}
\end{equation}
that admits a solution in form of Iterative Proportional Fitting procedure adapted towards score 
matching ideas. Now, suppose that we are interested in sampling from a posterior distribution
\begin{equation*}
    p(\vect{x}|\vect{y}^{\text{obs}}) \propto p_\text{data}(\vect{x}) g (\vect{y}^{\text{obs}}|\vect{x})
\end{equation*} using an N+1 - step discrete Schrödinger Bridge,
assuming that it is possible to sample $(\vect{x}, \vect{y}) \sim p_{\text{data}}(\vect{x}) g(\vect{y}|\vect{x})$. 
A straightforward approach would be to consider the SB problem \eqref{eq:sb_n_state_markov_chain}
with $p_{\text{data}}(\vect{x})$ replaced by the posterior $p(\vect{x}|\vect{y}^{\text{obs}})$, i.e.
\begin{equation}\label{eq:conditional_i2sb_original_statement}
    \pi^* = \argmin \limits_{\pi} \left\{ 
        D_{\text{KL}}(\pi || p_{\vect{y}^{\text{obs}}}): \pi_0 = p(\cdot|\vect{y}^{\text{obs}}), \pi_N = p_\text{prior}  
    \right\}
\end{equation}
where $
    p_{\vect{y}^{\text{obs}}}(\vect{x}_{0, N}) \coloneq 
    p(\vect{x}_0|\vect{y}^{\text{obs}}) \prod_{k=0}^{N-1} p_{k+1|k} (\vect{x}_{k+1}|\vect{x}_k)
$ 
is the forward noising process. The former, however, 
is explicitly dependent on $p(\vect{x}_0 | \vect{y}^{\text{obs}})$, which is intractable under our 
assumptions. 

This complication can be addressed by solving an amortized problem instead. 
Let $
    p_{\text{join}} (\vect{x}, \vect{y}) = 
    p_{\text{data}} (\vect{x}) g(\vect{y}|\vect{x}) = 
    p(\vect{x}|\vect{y}) p_{\text{obs}}(\vect{y})$, 
and $
    p_{\text{jprior}}(\vect{x}, \vect{y}) = 
    p_{\text{prior}} (\vect{x}) p_{\text{obs}} (\vect{y})
$ where $
    p_{\text{obs}} (\vect{y}) =
     \int p_{\text{data}} (\vect{x}) g(\vect{y}|\vect{x}) d\vect{x}
$. We are interested in finding the transition kernel
$\pi^{\text{cond}, *} = (\pi_{\vect{y}}^{\text{cond}, *})_{\vect{y} \in \mathcal{Y}}$ where 
$\pi_{\vect{y}}^{c, *}$ defines a distribution on 
$\mathcal{X} = (\mathbb{R}^d)^{N+1}$ for each $\vect{y} \in \mathcal{Y}$ satisfying
\begin{equation}\label{eq:conditional_i2sb_amortized_statement}
    \pi^{\text{cond}, *} = \argmin \limits_{\pi} \left\{ 
        \mathbb{E}_{\vect{y} \sim p_{\text{obs}}} [
            D_{\text{KL}}(\pi_{\vect{y}}^{\text{cond}} | p_Y)
        ]: \pi_0^{\text{cond}} \otimes p_{\text{obs}} = p_{\text{join}}, \pi_N^{\text{cond}} \otimes p_{\text{obs}} = p_\text{jprior}  
    \right\}
\end{equation}
which is an averaged version of \eqref{eq:conditional_i2sb_original_statement} over the distribution $p_{\text{obs}} (\vect{y})$
of $\vect{y}$. Once $\pi^{{\text{cond}}, *}$ is known, sampling $\vect{x}_N \sim p_{\text{prior}}$, then 
$\vect{x}_k|\vect{x}_{k+1} \sim \pi^{\text{cond}, *}_{\vect{y}, k|k+1} (\vect{x}_k | \vect{x}_{k+1})$ for $k = N-1, \ldots, 0$ yields 
$\vect{x}_0 \sim p(\vect{x}|\vect{y})$ given $Y \sim p(\vect{y}^\text{obs})$  

Problem \eqref{eq:conditional_i2sb_amortized_statement} could be reformulated as a SB on extended space, thus, guaranteeing 
the existence and uniqueness of the SB solution.
\begin{proposition} \textnormal{(Shi et al., \cite{shi2022conditional})} Consider the following SB problem 
\begin{equation} \label{eq:i2sb_extended_space}
    \overline{\pi}^* = \argmin_{\overline{\pi}}  \left\{
        D_{\text{KL}}(\overline{\pi} |\overline{p}): 
        \text{s.t.} \, \overline{\pi}_0 = p_{\text{join}}, \overline{\pi}_N = p_{\text{jprior}}
    \right\}   
\end{equation}
where $\overline{p}(\vect{x}_{0:N}, \vect{y}_{0:N}) \coloneq p_{\vect{y}_0} (\vect{x}_{0:N}) 
\overline{p}_{\text{obs}}(\vect{y}_{0:N})$ 
with $\overline{p}_{\text{obs}}(\vect{y}_{0:N}) \coloneq \prod_{k=0}^{N-1} \delta_{\vect{y}_k} (\vect{y}_{k+1})$, 
and $p_{\vect{y}_0}$ is the forward process of \eqref{eq:conditional_i2sb_original_statement}.
If $D_{\text{KL}}(\overline{\pi}^* | \overline{p}) < +\infty$, then 
$\overline{\pi}^* =  \pi^{\text{cond}, *} \otimes \overline{p}_{\text{obs}}$, where $\pi^{c, *}$ 
solves \eqref{eq:conditional_i2sb_amortized_statement}.
\end{proposition} Compared to \eqref{eq:conditional_i2sb_original_statement}, both distributions $p_{\text{jprior}}$
and $p_{\text{join}}$ could be sampled analytically, thus, enabling numerical methods of SB problem solving, 
such as the one introduced in \cite{debortoli2021diffusion}.

\subsection{Image-to-Image Schrödinger Bridge (\(\text{I}^2\text{SB}\))}

Liu et al. \cite{liu2023i2sbimagetoimageschrodingerbridge} demonstrate that for 
certain cases of SB problem statement the necessity to solve system \eqref{eq:kfp_equations_sb}
could be circumvented altogether. Such conclusion could be drawn from comparison 
between \eqref{eq:kfp_equations_sb} and Fokker-Plank equation for Ito process 
\eqref{eq:forward_sde_general_form}, yielding an the alternative formulation 
for \eqref{eq:sb_equations}, with $\nabla_{\vect{x}_t} \log \hat{\Psi} (\vect{x}_t, t)$ and
$\nabla_{\vect{x}_t} \log \Psi (\vect{x}_t, t)$ as score functions of the following 
linear SDEs, respectively   
\begin{subequations} \label{eq:linear_sde_for_sb}
    \begin{equation} \label{eq:linear_sde_for_sb_backward_potential}
        d \vect{x}_t = \vect{f}_t (\vect{x}_t, t) dt + \sqrt{\beta_t} d\vect{w}_t \quad 
        \vect{x}_0 = \hat{\Psi}(\cdot, 0)
    \end{equation}
    \begin{equation} \label{eq:linear_sde_for_sb_forward_potential}
        d \vect{x}_t = \vect{f}_t (\vect{x}_t, t) dt + \sqrt{\beta_t} d\hat{\vect{w}}_t \quad 
        \vect{x}_T = \Psi(\cdot, T)
    \end{equation}        
\end{subequations}
Hence, if the values $\Psi(\cdot, T)$ and $\hat{\Psi}(\cdot, 0)$ are tractable, 
nonlinear drift $\nabla_{\vect{x}_t} \log \hat{\Psi}$ transporting samples 
from $p_{\text{prior}}$ to $p_{\text{data}}$ could be learned as a score 
of linear SDE, identical to the one present in SGM framework. 
The former could be achieved by assuming 
that data distribution is the \textit{Dirac delta} centered at $a \in \mathbb{R}^d$,
i.e., $p_\text{data}(\cdot, 0) = \delta_{a}(\cdot)$. In that case,
\begin{equation} \label{eq:schrodinger_potential_simplified_form}
    \hat{\Psi}(\cdot, 0) = \delta_{a}(\cdot) \quad \Psi(\cdot, T) = \dfrac{p_\text{prior}}{\hat{\Psi}(\cdot, T)}
\end{equation}
Thus, the model under consideration learns a mixture of scores of degenerate distributions centered on the elements of the training dataset.
Additionally, the paper \cite{liu2023i2sbimagetoimageschrodingerbridge} derives the analytic posterior for equations 
\eqref{eq:sb_equations} given a boundary pair ($\vect{x}_0, \vect{x}_T$) under the assumption of $f \coloneq 0$. Namely,
\begin{subequations} \label{eq:sb_posterior_analytical_form}
    \begin{equation}
        q ( \vect{x}_t |\vect{x}_0, \vect{x}_1) = \mathcal{N} \left(
            \vect{x}_t; \vect{\mu}_t(\vect{x}_0, \vect{x}_1), \Sigma_t 
         \right)
    \end{equation}
    \begin{equation}
        \vect{\mu}_t = \dfrac{\overline{\sigma}_t^2}{\overline{\sigma}_t^2 + \sigma_t^2} \vect{x}_0 +
                      \dfrac{\sigma_t^2}{\overline{\sigma}_t^2 + \sigma_t^2} \vect{x}_1
        \quad 
        \Sigma_t = \dfrac{\sigma_t^2 \overline{\sigma}_t^2}{\sigma_t^2 + \overline{\sigma}_t^2} \cdot I                   
    \end{equation}
\end{subequations}
where $\sigma_t^2 = \int_0^t \beta_\tau d\tau$ and  $\overline{\sigma}_t^2 = \int_t^T \beta_\tau d\tau$ are variances accumulated 
from two sides of the diffusion bridge. In addition, such posterior marginalizes the recursive posterior sampling in DDPM \cite{ho2020ddpm}
\begin{equation}\label{eq:sb_posterior_with_ddpm_posterior}
    q \left(\vect{x}_n | \vect{x}_0, \vect{x}_N \right) = 
    \int \prod_{k=n}^{N-1} p_{\text{DDPM}} (\vect{x}_k | \vect{x}_0, \vect{x}_{k+1}) d \vect{x}_{k+1}  
\end{equation}
The above circumstances facilitate SB-based generative model building, 
since \cref{eq:sb_posterior_analytical_form} allows for analytical sampling of $\vect{x}_t$ 
during training, when $\vect{x}_0$ and $\vect{x}_T$ are available. Moreover,  
\cref{eq:sb_posterior_with_ddpm_posterior} ensures that running convenional DDPM
starting from $\vect{x}_T$ during inference induces marginal density of SB paths.  

Finally, in the limit of vanishing stochasticity, i.e. $\beta_t \rightarrow 0$, the SDE between 
$(\vect{x}_0, \vect{x}_T)$ reduces to optimal transport ODE
\begin{equation}
    d\vect{x}_t = \vect{v}_t(\vect{x}_t | \vect{x}_0) dt 
    \quad 
    \vect{v}_t(\vect{x}_t | \vect{x}_0) = \dfrac{\beta_t}{\sigma_t^2} \left( \vect{x}_t - \vect{x}_0 \right)
\end{equation} 
which, similarly to probability flow ODE \cite{song2020score-based-modelling-SDE}, 
could be utilized for deterministic sampling and likelihood estimation.

\section{Problem Statement and Suggested Method}

Consider the acoustic FWI problem statement 
(\cref{eq:acoustic_fwi_problem_statement,eq:acoustic_fwi_problem_statement_pseudoinverse})
coupled with additional information provided by a smooth velocity model $\vect{c}_{\text{smooth}}$ of the medium under investigation. 
\begin{equation}
    \begin{cases}
        \hfil \min\limits_{\vect{c}_0} \, J(\vect{d}_{\text{model}},  \vect{d}_{\text{obs}}) + \lambda R(\vect{c}) \\
        \text{s.t.} \quad \vect{d}_{\text{model}} = F(\vect{c}_0),\quad \vect{c}_1 = \vect{c}_{\text{smooth}}
    \end{cases}
\end{equation}
$\vect{c}_{\text{smooth}}$ is believed to be reasonably close to ground truth velocity model $\vect{c}_0^*$,  
yet lacking high-frequency details. Hence, in realistic inversion scenarios $\vect{c}_{\text{smooth}}$ 
is commonly employed as a starting point for nonlinear optimization. 

We propose a novel way to utilize such piece of information in context of recently 
proposed diffusion-based deep learning approach to acoustic waveform inversion. 
Our method is two-staged: the first stage consists of fitting the parameters of c\(\text{I}^2\text{SB}\)
model on training data. Once the optimal set of parameters is found, 
$\vect{c}_0^* = \hat{F}^{-1}_{\vect{\theta}^*} (\vect{d}_{\text{obs}}, \vect{c}_{\text{smooth}})$ 
is obtained by running the inference process of such model.

The transition from \(\text{I}^2\text{SB}\) to c\(\text{I}^2\text{SB}\) proceeds as follows. 
According to \cite{shi2022conditional}, continuous dynamic SB problem statement matching 
the one described by equation \eqref{eq:conditional_i2sb_amortized_statement} 
is formulated as   
\begin{equation} \label{eq:continuous_conditional_i2sb_extended_space}
    \Pi^* = \argmin_{\Pi}  \left\{
        D_{\text{KL}}(\Pi |\mathbb{P}): 
        \Pi \in \mathcal{P} (\mathcal{C}), \Pi_0 = p_{\text{join}}, \Pi_T = p_{\text{jprior}}
    \right\}   
\end{equation}
where reference measure $\mathbb{P}$ is associated with
\begin{equation}\label{eq:continuous_conditional_i2sb_extended_space_ref_measure}
    \begin{cases}
        d \vect{x} = f(\vect{x}, t) dt + \sqrt{\beta}_t d \vect{w}_t \\
        d \vect{y} = 0 
    \end{cases}
\end{equation}
Schrödinger potentials for such dynamics are defined by a set of hyperbolic PDEs
coupled with their boundary conditions \cite{chen2020stochasticcontrolliaisonsrichard}
\begin{subequations} \label{eq:kfp_equations_conditional_sb}
    \begin{equation} \label{eq:kfp_forward_equation_conditional_sb}
        \dfrac{\partial \Psi_t (\vect{x}|\vect{y})}{\partial t} = 
        -\vect{f}_t (\vect{x}) \cdot \nabla_{\vect{x}} \Psi_t^{T} (\vect{x}|\vect{y})  
        -\beta_t \dfrac{1}{2} \Laplace \Psi_t (\vect{x}|\vect{y})
    \end{equation}
    \begin{equation} \label{eq:kfp_backward_equation_conditional_sb}
        \dfrac{\partial \hat{\Psi}_t (\vect{x} | \vect{y})}{\partial t} = 
         -\nabla_{\vect{x}} \cdot (\hat{\Psi}\vect{f}) (\vect{x}|\vect{y}) 
         + \beta_t \dfrac{1}{2} \Laplace \hat{\Psi}_t (\vect{x}|\vect{y})    
    \end{equation}
    \begin{equation}
        \textit{s.t.} \quad \Psi_0 (\vect{x}, \vect{y}) \hat{\Psi}_0 (\vect{x},\vect{y}) = 
        p_{\text{join}} \quad  
        \Psi_T (\vect{x}, \vect{y})\hat{\Psi}_T (\vect{x}, \vect{y}) = p_{\text{jprior}}
    \end{equation}
\end{subequations}
The solution of \eqref{eq:continuous_conditional_i2sb_extended_space} is a
path measure associated with either the forward \eqref{eq:conditional_sb_forward_sde}, or 
the backward \eqref{eq:conditional_sb_backward_sde} SDE with gradients of potentials introduced 
in \eqref{eq:kfp_equations_conditional_sb} added as nonlinear drift terms

\begin{subequations} \label{eq:conditional_sb_equations}
    \begin{equation} \label{eq:conditional_sb_forward_sde}
        \begin{cases}
            d \vect{x}_t = \left[
                \vect{f}_t + \beta_t \nabla_{\vect{x}_t} \log \Psi (\vect{x}_t, t | \vect{y}) 
            \right] dt + \sqrt{\beta_t} d\vect{w}_t \\
            d \vect{y}_t = 0 \\
            (\vect{x}_0, \vect{y}_0) \sim p_{\text{join}}
        \end{cases} 
    \end{equation}
    \begin{equation} \label{eq:conditional_sb_backward_sde} 
        \begin{cases}
            d \vect{x}_t = \left[
                \vect{f}_t - \beta_t \nabla_{\vect{x}_t} \log \hat{\Psi} (\vect{x}_t, t | \vect{y}) 
            \right] dt + \sqrt{\beta_t}  d\overline{\vect{w}}_t \\
            d \vect{y}_t = 0 \\
            (\vect{x}_T, \vect{y}_T) \sim p_{\text{jprior}}
        \end{cases}
    \end{equation}
\end{subequations}

Let $\vect{x}_0 \sim p_{\text{data}}$ be the selected element of the training set. 
If $p_{\text{data}} = \delta_{\vect{x}_0} (\cdot)$, then 
$p_\text{obs} (\vect{y}) = g(\vect{y}| \vect{x}_0)$ and
$p_\text{join}^{\vect{x}_0}(\vect{x}, \vect{y}) = \delta_{\vect{x}_0} (\vect{x}) g (\vect{y}| \vect{x}_0)$. 
Consider the joint prior  $p_{\text{jprior}}$ given by $p_{\text{jprior}}^{\vect{x}_0} (\vect{x}, \vect{y}) = p_{\text{dist}} (\vect{x} | \vect{x}_0)g(\vect{y}| \vect{x}_0)$.
Similarly to vanilla  \(\text{I}^2\text{SB}\), we will approximate 
the scores of the aggregated set of distribution pairs 
$\left(p_\text{join}^{\vect{x}_0}, p_{\text{jprior}}^{\vect{x}_0} \right)$
within the scope of the same deep learning model. 
To do so, we line for line apply the derivation process of image-to-image Schrödinger Bridge to 
the system \eqref{eq:conditional_sb_equations}.
After adapting the notation towards the acoustic inversion problem statement, 
we arrive at algorithm \ref{alg:conditional_i2sb_training}.
It solves a particular case of conditional half-bridge problem for general 
amortized conditional SB statement \eqref{eq:conditional_i2sb_amortized_statement},
in which $\vect{c}_1$ continuously drifts towards $\vect{c}_0$ while $\vect{d}_{\text{obs}}$ remains unchanged
throughout the process. 

\begin{algorithm}[H]
    \caption{Conditional \(\text{I}^2\text{SB}\) training}
    \begin{algorithmic}[1]
    
    \Require{
        $\vect{c}_{\vect{\theta}} (\vect{c}_t, t, \vect{d}_{\text{obs}})$ - denoising neural network, 
        $\beta_t$ - noise schedule for eq. \eqref{eq:sb_equations} with $f \coloneq 0$,
        $p_{\text{join}}(\vect{c}_0, \vect{d}_{\text{obs}})$ - training dataset, 
        $p_{\text{dist}} (\vect{c}_1 | \vect{c}_0)$ - distortion operator,
        $p_{\text{uncond}}$ - probability of unconditional training  
    } 
    
    \Repeat
    \State Sample \(
        t \sim \mathcal{U}\left([0, 1] \right), 
        \left( \vect{c}_0, \vect{d}_{\text{obs}} \right) 
        \sim p_{\text{data}}(\vect{c}_0, \vect{d}_{\text{obs}}),
        \vect{c}_1 \sim p_{\text{dist}} (\vect{c}_1| \vect{c}_0)
    \)
    \State Sample $\vect{c}_t \sim q \left(\vect{c}_t | \vect{c}_0, \vect{c}_1 \right)$ 
           according to \eqref{eq:sb_posterior_analytical_form}
    \State Take gradient step on $\vect{\theta}$ w.r.t $\norm{
        \vect{c}_0 - \vect{c}_{\vect{\theta}} (\vect{c}_t, t, \vect{d}_{\text{obs}})
    }_2^2$
    \Until{converges}
    \end{algorithmic}
    \label{alg:conditional_i2sb_training}
\end{algorithm}

Excluding observed data $\vect{d}_\text{obs}$ from neural network inputs in alg. \labelcref{alg:conditional_i2sb_training} 
results in the exact \(\text{I}^2\text{SB}\) training procedure proposed in \cite{liu2023i2sbimagetoimageschrodingerbridge}.
Thus, it is possible to learn both conditional and unconditional nonlinear drifts 
with the same model (alg. \labelcref{alg:guided_conditional_i2sb_training}).
During our experiments we have discovered that assigning higher weights for loss terms that include conditional 
input is beneficial for overall model performance. This behaviour is connected to the prior being 
informative enough to derive $\vect{c}_0$ purely from $\vect{c_1}$. Hence, joint c\(\text{I}^2\text{SB}\) 
training procedure without additional weighting tends to converge towards 
the local minimum in which neural network ignores the conditional input altogether. 
\begin{algorithm}[H]
    \caption{Guided Conditional \(\text{I}^2\text{SB}\) training}
    \begin{algorithmic}[1]
    
    \Require{
        $\vect{c}_{\vect{\theta}} (\vect{c}_t, t, \vect{d}_{\text{obs}})$ - denoising neural network, 
        $\beta_t$ - noise schedule for eq. \eqref{eq:sb_equations} with $f \coloneq 0$,
        $p_{\text{join}}(\vect{c}_0, \vect{d}_{\text{obs}})$ - training dataset, 
        $p_{\text{dist}} (\vect{c}_1 | \vect{c}_0)$ - distortion operator,
        $p_{\text{uncond}}$ - probability of unconditional training,
        $w_{\text{cond}}$ - loss weighting coefficient in conditional setting 
    } 
    
    \Repeat
    \State Sample \(
        t \sim \mathcal{U}\left([0, 1] \right), 
        \left( \vect{c}_0, \vect{d}_{\text{obs}} \right) 
        \sim p_{\text{data}}(\vect{c}_0, \vect{d}_{\text{obs}}),
        \vect{c}_1 \sim p_{\text{dist}} (\vect{c}_1| \vect{c}_0)
    \)
    \State Sample $\vect{c}_t \sim q \left(\vect{c}_t | \vect{c}_0, \vect{c}_1 \right)$ 
           according to \eqref{eq:sb_posterior_analytical_form}
    \State Sample $\xi \sim \text{Bin}(p_{\text{uncond}})$.
    \State  If $\xi = 1$, set $w=0, \, \vect{d}_{\text{obs}} = \vect{0}$, else 
    set $w=w_{\text{cond}}$
    \State Take gradient step on $\vect{\theta}$ w.r.t $ (1 + w) \cdot \norm{
        \vect{c}_0 - \vect{c}_{\vect{\theta}} (\vect{c}_t, t, \vect{d}_{\text{obs}})
    }_2^2$
    \Until{converges}
    \end{algorithmic}
    \label{alg:guided_conditional_i2sb_training}
\end{algorithm}

Following the analogy with classifier-free guidance, we propose a modification to \(\text{I}^2\text{SB}\)
sampling scheme which employs weighted combination of conditional and unconditional predictions
(alg. \ref{alg:guided_conditional_i2sb_sampling}).
Intuitively such scheme should reproduce the effect of low-temperature sampling for GANs \cite{brock2018large}, 
as similar property was observed in SGM setting \cite{ho2022classifierfreediffusionguidance},
although we leave rigorous theoretical treatment of the issue for future studies.  

\begin{algorithm}
    \caption{Guided Conditional \(\text{I}^2\text{SB}\) sampling}
    \begin{algorithmic}[1]
    
    \State \textbf{Input}  $\vect{c}_{\vect{\theta}} $ - pretrained denoising neural network, 
    $\vect{c}_{\text{smooth}}$ - smoothed velocity model, 
    $\{ t_n\}_{n=1}^{N}$ - time interval discretization, $\eta$ - guidance scale,
    $\vect{d}_{\text{obs}}$ - observed seismic data
    \State $\vect{c}_N \leftarrow \vect{c}_{\text{smooth}}$
    \For{$n=N$ \textbf{to} $n=1$}
    \State Predict $\tilde{\vect{c}}_0 =  \eta \cdot \vect{c}_{\vect{\theta}} 
    \left(\vect{c}_n, t_n, \vect{d}_{\text{obs}}\right) + (1 - \eta) \cdot 
    \vect{c}_{\vect{\theta}} \left(\vect{c}_n, t_n, \vect{0}\right) $
    \State Sample $\vect{c}_{n-1} \sim p(\vect{c}_{n-1}|\tilde{\vect{c}}_0, \vect{c}_{n})$ using DDPM posterior 
    \EndFor
    
    \State \Return $\vect{c}_0$
    \end{algorithmic}
    \label{alg:guided_conditional_i2sb_sampling}
\end{algorithm}

\section{Related works}

Wang et al. \cite{wang2023apriorregularizedfwiusinggenerativediffusionmodels} proposed incorporating diffusion-based 
regularization into the workflow of acoustic FWI by complementing each denoising step of pre-trained diffusion model 
with a few iterations of gradient optimization. As a result, the authors claim a level of sample fidelity 
unattainable with conventional FWI, as well as measurement consistency.  

Wang et al. \cite{wang2024controllable} demonstrated that the output of a diffusion model trained on synthetic 
acoustic velocity fields from the OpenFWI data collection could be manipulated by conditioning on 
different types of information that might be available during seismic studies. 
For this purpose, a conditional diffusion model was trained using classifier-free diffusion guidance. The authors report high degree of 
perceptual similarity to intended velocity fields on validation samples, along with inconsequential deterioration of reconstruction quality  
on out-of-distribution data. 

The paper \cite{zhang2024diffusionvel} showcases an alternative way of including multimodal prior information obtained with seismic 
studies into the DDPM sampling scheme. Namely, Zhang et al. utilize an ensemble of conditional diffusion models to obtain multiple
predictions of the original velocity model at each denoising step. Each model works with an individual mode of the available data. 
A linear combination of predictions with a weighting schedule is employed to proceed to the next step.
Reconstruction quality is evaluated on several datasets from the OpenFWI collection, demonstrating superior performance
compared to InversionNet and VelocityGAN frameworks \cite{wu2019inversionnet, zhang2020data}. 
Since the latter were trained with waveform data exclusively, the research indicates that auxiliary 
information could indeed improve data-driven inversion procedures.

A key limitation of sampling schemes introduced above is their reliance on the standard Gaussian distribution as the terminal
diffusion point, despite its lack of connection with the data available to the model at inference stage. 
Thus, the forward noising process has to run for a sufficiently long time to transfer data points 
from the original distribution to the standard Gaussian noise. Additionally, the numerical approximation of SDE \eqref{eq:forward_sde_general_form} 
is only accurate when the discretization timestep is small. This property puts a lower bound on the amount of NFEs required for cSGMs-based inference to perform well.
On the other hand, generative models built on the basis of Schrödinger bridges rely neither on limit property
of specific class of noising diffusions, nor on the tractability of 
the prior distribution \cite{debortoli2021diffusion}. These features are critical for us, since 
they: 1) allow us to construct a theoretically sound foundation for including smoothed velocity fields into diffusion-based acoustic model building; 
2) reduce the total computational cost of inference, as Schrödinger bridge-based models generally achieve sample fidelity similar 
to SGM-based models, but with less NFEs. 
{
\section{Results}

\subsection{Experiment Setup}

\subsubsection{Model} \label{subsubsection:model}

{
The architecture chosen for c\(\text{I}^2\text{SB}\) numerical experiments is a variation of the 
attention-augmented U-Net with timestep embedding proposed in \cite{DBLP:journals/corr/abs-2105-05233} 
(see fig. \ref{fig:attention_augmented_unet_schematic}). We introduce two models to investigate 
the effect of parameter scaling on framework performance: 
our \textbf{ci2sb-seismic-small} and \textbf{ci2sb-seismic-large} models are comprised of
approx. $9.5 \cdot 10^5$ and $6.3 \cdot 10^7$ parameters, respectively. Model inputs are  $B \times C \times H \times W$ tensors;  
the first channel of each tensor is reserved for noisy samples $\vect{c}_t$ at timestep $t$, 
while the remaining channels contain conditional information. SB is conditioned on seismic data 
$\vect{d}_{\text{data}}$ with each shot being reduced to the shape of $\vect{c}_t$ via linear interpolation.
To implement classifier-free guidance \cite{ho2022classifierfreediffusionguidance}, 
we occasionally zero-mask conditional inputs with set probability $p_{\text{uncond}} = 0.5$.

Linear drift for equations \eqref{eq:linear_sde_for_sb} is set to 
\(f \coloneq 0 \), whereas the noise schedule $\beta_t$ is selected to be shrinking at both 
boundaries (\cite{chen2021likelihood, debortoli2021diffusion, liu2023i2sbimagetoimageschrodingerbridge}). 
Time interval is uniformly split into 1000 sampling steps. 
Similarly to \cite{karras2022elucidatingdesignspacediffusionbased} UNet is parameterised
to estimate \(\vect{c}_0\) at each denoising iteration. However, network preconditioning and objective rescaling proposed 
in the same paper did not result in performance gain.
We experimented with different values of reweighting coefficient; setting $w_\text{cond} = 100$ turned out to be 
a reliable choice that enhances the performance while keeping the training process stable. 

\begin{figure}[h!]
    \centering
    \captionsetup{justification=centering}
    \includegraphics[width=120mm]{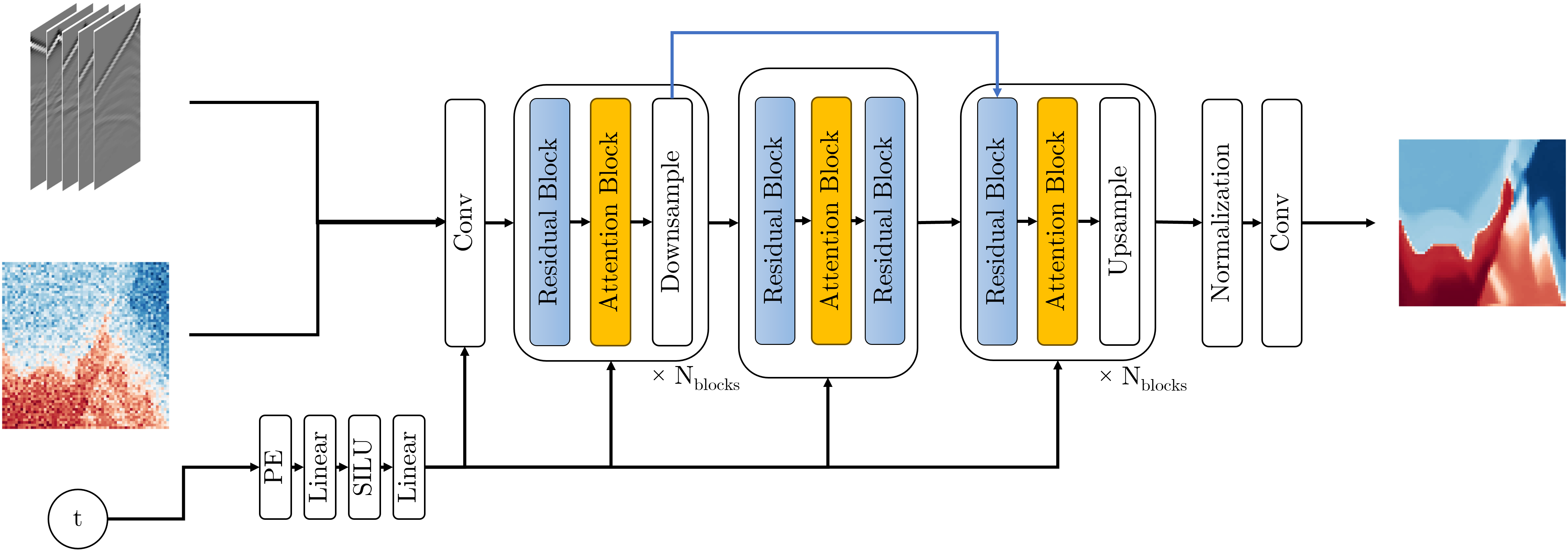}
    \caption{
        A schematic of attention-augmented UNet network used for numerical studies throughout the study. 
        Original architecture was proposed in  \cite{DBLP:journals/corr/abs-2105-05233} 
    }
    \label{fig:attention_augmented_unet_schematic}
\end{figure}

}

\subsubsection{Data} \label{subsubsection:data}

{
The current study is primarily focused on OpenFWI \cite{DBLP:journals/corr/abs-2111-02926} 
- an open-access collection of synthetic seismic data. OpenFWI consists of several dataset batches, 
each corresponding to specific domain of geological interest. Dataset entries are represented by 
procedurally generated velocity maps and matching waveform data obtained through numerical simulation of 
acoustic media dynamics (see fig. \ref{fig:openfwi_velocity_map_samples}). The closure of 
the problem statement for numerical modelling is achieved by imposing a free-surface boundary condition on 
the upper edge and non-reflective boundary condition on the other edges of area under investigation. 
Modelling is carried out by a finite difference scheme on spatially staggered grid, 
with order of approximation being second in time and fourth in space. 
Synthetic waveform data is recorded at a sampling rate of 1000 Hz with equidistant receivers located 
at the upper edge of the medium. The scanning setup employs 5 source terms 
scattered over the top boundary of the medium; each emits 15 Hz Ricker wavelet
\begin{figure}[h!]
    \centering
    \captionsetup{justification=centering}
    \includegraphics[width=120mm]{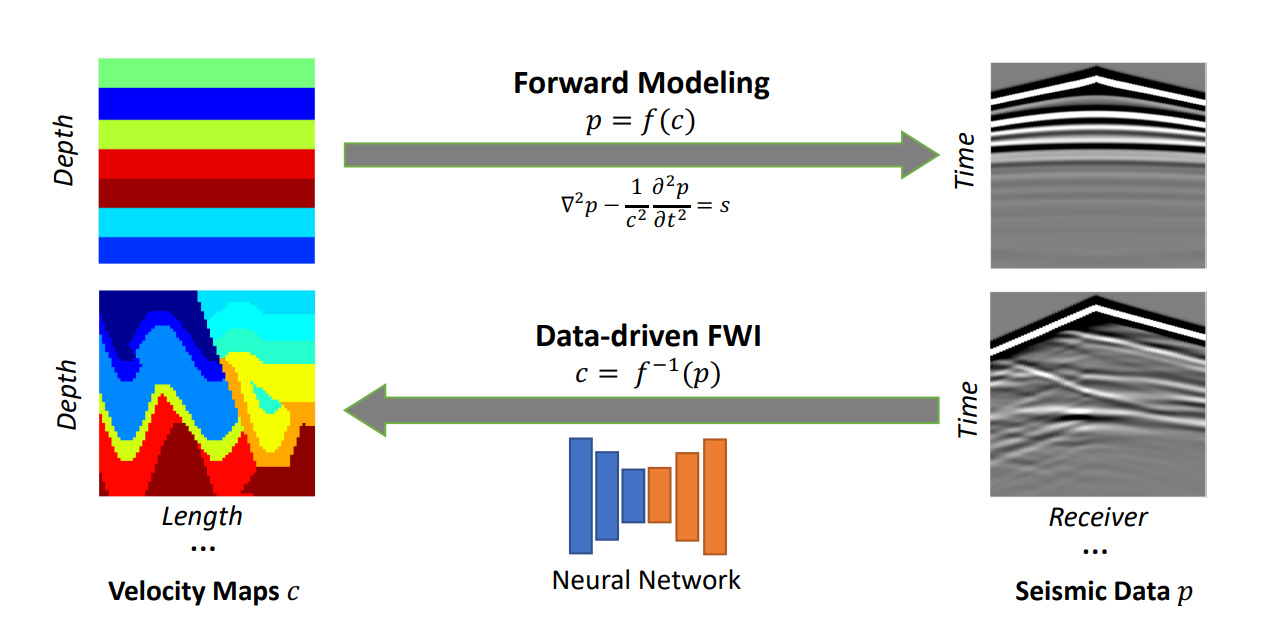}
    \caption{OpenFWI velocity map and seismic data samples \cite{DBLP:journals/corr/abs-2111-02926}}
    \label{fig:openfwi_velocity_map_samples}
\end{figure}

For training and evaluation purposes we utilize velocity maps and seismic signals from "Vel", 
"Fault", and "Style" families. Train/validation split ratio for each dataset is kept unchanged from the original paper 
(tab. \labelcref{tab:openfwi_dataset_instance_summary}). Velocity map tensors are resized 
to the shape \(1\times64\times64\) and scaled to range [-1, 1.]. In turn, seismograms are
subjected to logarithmic transformation, normalized to range [-1, 1.], and resized with linear interpolation 
to be of the shape \(5\times64\times64\).

\begin{table}[h!]
    \centering
    \resizebox{0.9\textwidth}{!}{
        \resizebox{\columnwidth}{!}{
            \begin{tabular}{c|c|cccc}
            \hline
                Group             & 
                Dataset           & 
                Size              & 
                \#Train/\#Test    & 
                Seismic Data Size & 
                Velocity Map Size 
            \\ \hline
                Vel Family        & 
                \begin{tabular}[c]{@{}c@{}}FlatVel-A/B\\ CurveVel-A/B\end{tabular} & 
                \begin{tabular}[c]{@{}c@{}}43 GB\\ 43 GB\end{tabular} & 
                \begin{tabular}[c]{@{}c@{}}24K / 6K\\ 24K / 6K\end{tabular} & 
                \begin{tabular}[c]{@{}c@{}} $5 \times 1000 \times 1 \times 70$ \\ $5 \times 1000 \times 1 \times 70 $\end{tabular} & 
                \begin{tabular}[c]{@{}c@{}} $70 \times 1 \times 70 $\\ $70 \times 1 \times 70$ \end{tabular} 
            \\
                Fault Family      & 
                \begin{tabular}[c]{@{}c@{}}FlatFault-A/B\\ CurveFault-A/B\end{tabular} &
                \begin{tabular}[c]{@{}c@{}}77 GB\\ 77 GB\end{tabular} & 
                \begin{tabular}[c]{@{}c@{}}48K / 6K\\ 48K / 6K\end{tabular} & 
                \begin{tabular}[c]{@{}c@{}}$5 \times 1000 \times 1 \times 70 $\\ $5 \times 1000 \times 1 \times 70$\end{tabular} & 
                \begin{tabular}[c]{@{}c@{}}$70 \times 1 \times 70 $\\ $70 \times 1 \times 70 $ \end{tabular}       
            \\
                Style Family      & 
                Style-A/B         & 
                95 GB             & 
                60K / 7K          & 
                $5 \times 1000 \times 1 \times 70 $ & 
                $70 \times 1 \times 70 $
            \\
            \hline
    
            \end{tabular}
        }
    }
    \caption{Summary of OpenFWI dataset instances used in the study}
    \label{tab:openfwi_dataset_instance_summary}
\end{table}
}

{
\subsubsection{Distortion Operator}

The training of  c\(\text{I}^2\text{SB}\) assumes the prior distribution at the end of the 
time interval given by \(p(\vect{c}_0, \vect{c}_1) = p_{\text{data}}(\vect{c}_0)p_{\text{dist}}(\vect{c}_1|\vect{c}_0)\). 
We provide \(p_{\text{dist}}(\cdot |\vect{c}_0) \) as a parametric stochastic operator 
\begin{equation} 
    \vect{c}_1 = \mathcal{D}_k^{\gamma} (\vect{c}_0) =\mathcal{S}_k \left(\gamma \cdot \vect{z} + (1 - \gamma) \cdot \vect{c}_0\right), 
    \quad \vect{z} \sim \mathcal{N}(\vect{0}, \vect{I})
\end{equation}
Here \( \mathcal{S}_k\) is the spatial Gaussian filter with kernel size \(k \times k\).
We set $k$ to be in the integer range from 8 to 16, while \(\gamma\) varies uniformly from 0.0 to 0.2. 
The design choice for $\mathcal{D}_k^{\gamma}$ complies with the concept that initial guesses lack fine-grained details present in
the original velocity models. It also ensures high variance of endpoint distributions (fig. \ref{fig:degradation_operator_samples}). 
Note, however, that suggested construction is generic and does not have any basis in real-world data. 
}

\begin{figure}[h!]
    \centering
    \begin{minipage}[b]{0.3\textwidth}
        \centering
        \includegraphics[width=\textwidth]{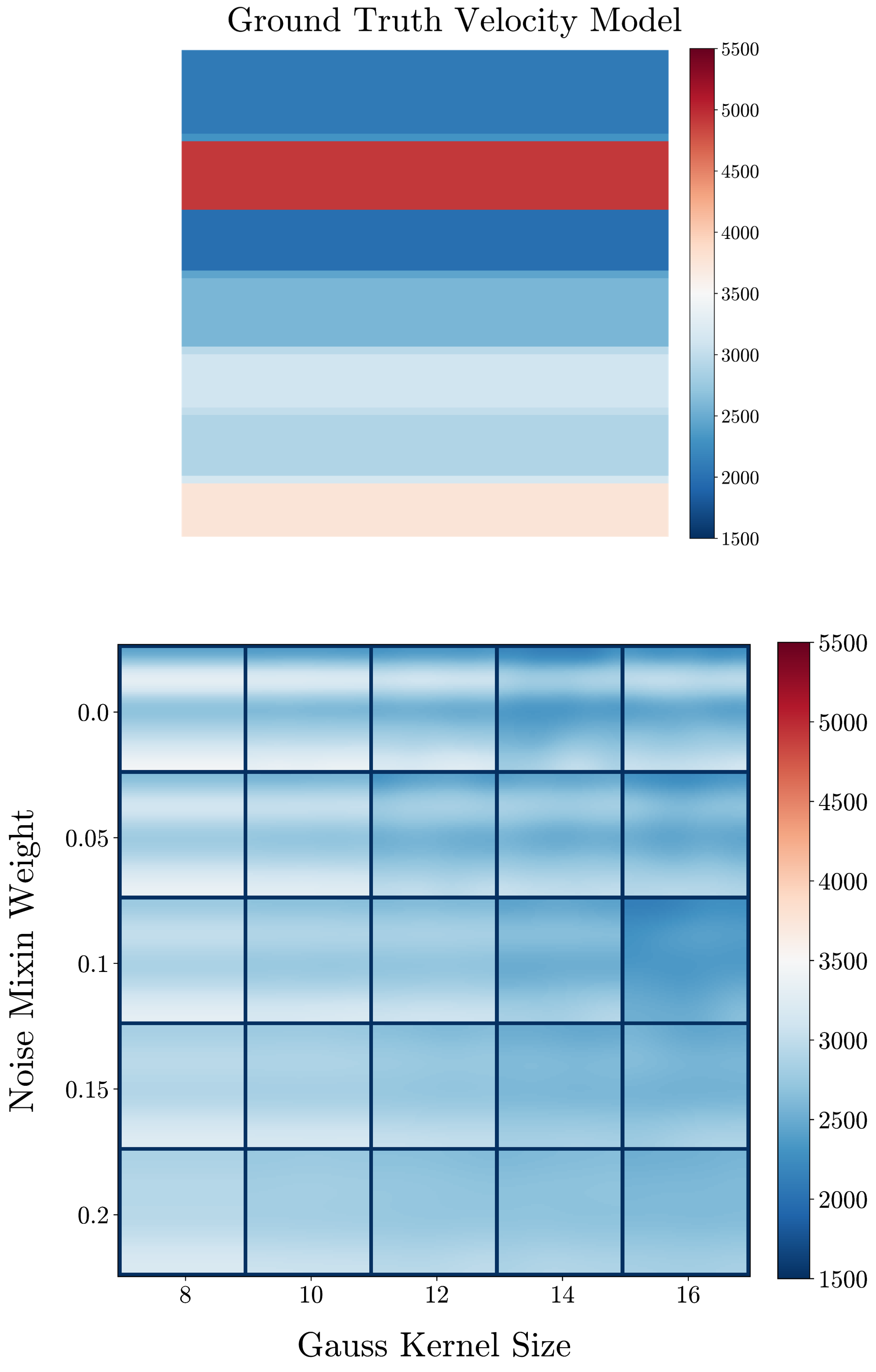}
        \subcaption{FlatVel\_B}
        \label{fig:degradation_operator_flatvel_b}
    \end{minipage}
    \begin{minipage}[b]{0.305\textwidth}
        \centering
        \includegraphics[width=\textwidth]{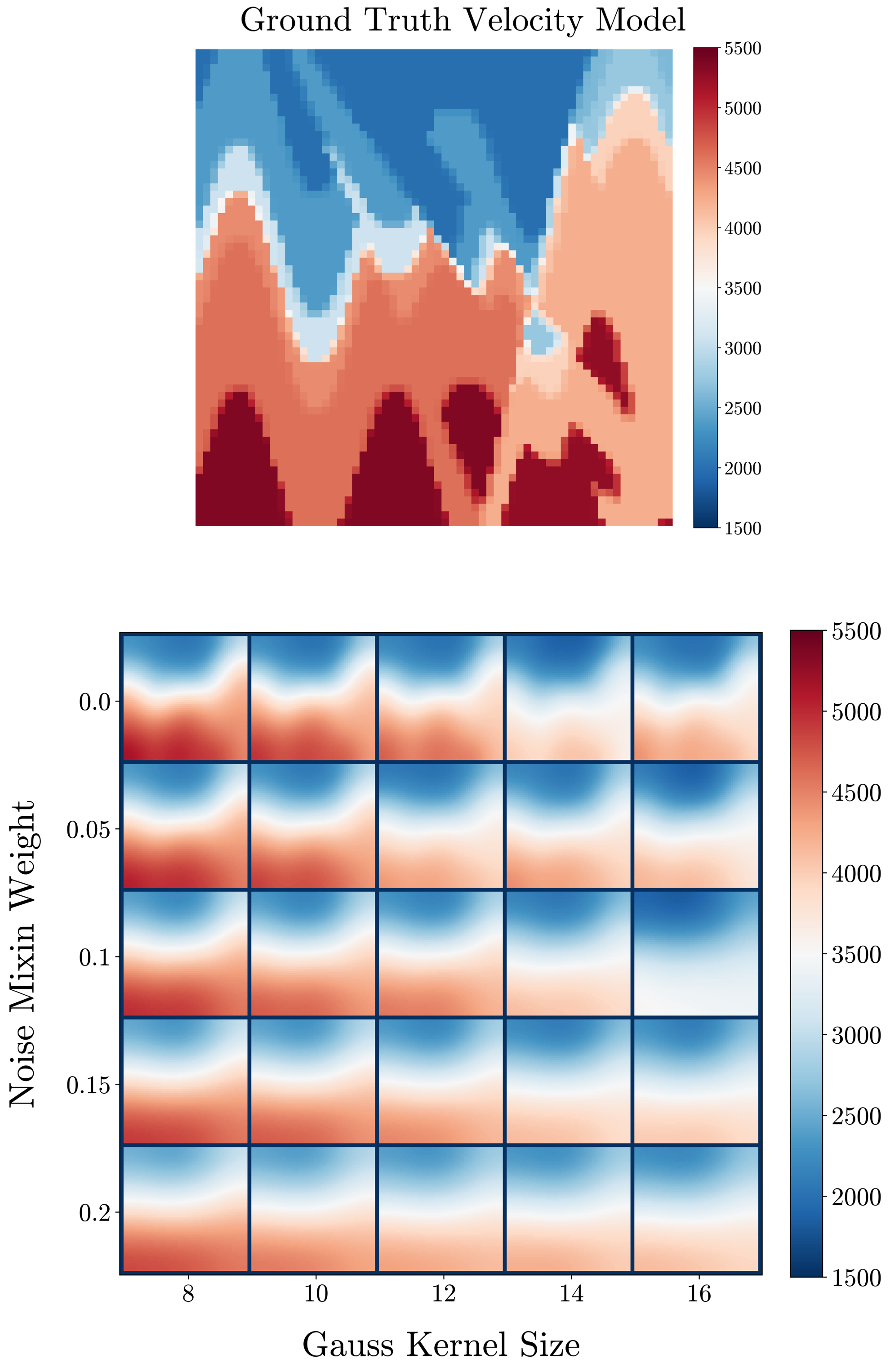}
        \subcaption{CurveFault\_B}
        \label{fig:degradation_operator_curvefault_b}
    \end{minipage}
    \begin{minipage}[b]{0.3\textwidth}
        \centering
        \includegraphics[width=\textwidth]{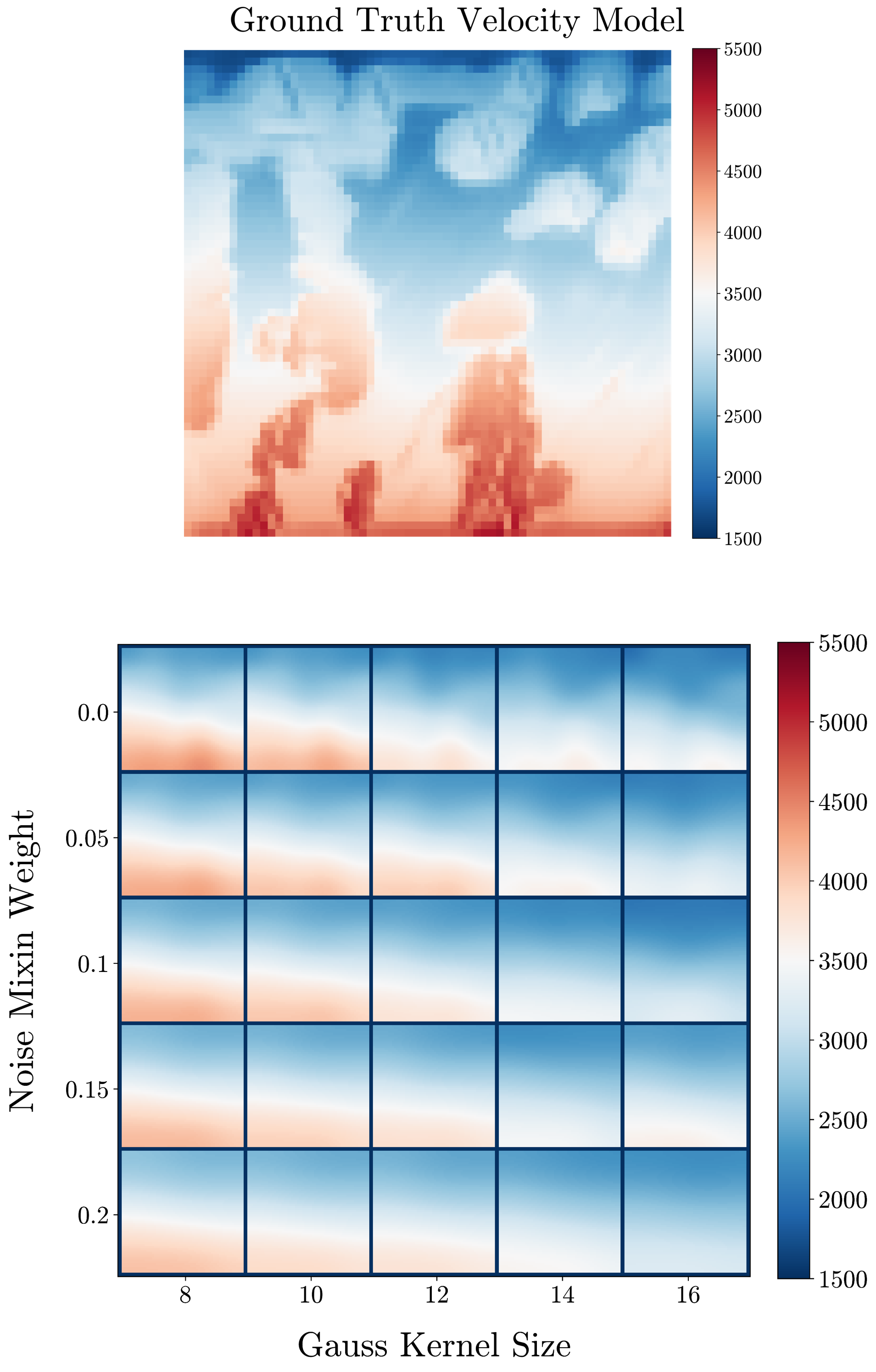}
        \subcaption{Style\_B}
        \label{fig:degradation_operator_style_b}
    \end{minipage}
    \caption{Initial guess distributions obtained via application of degradation operator to velocity model samples from different datasets}
    \label{fig:degradation_operator_samples}
\end{figure}
}

\subsubsection{Baselines}

We contrast the suggested method with our reimplementation of two data-driven seismic inversion 
frameworks; namely, InversionNet \cite{wu2019inversionnet}, which accepts 
both seismic data and smoothed velocity models as inputs, 
as well as the conditional SGM model suggested in \cite{wang2024controllable}. 

Our implementation of InversionNet is straightforward -- we utilize the same network architecture 
and data processing pipeline proposed in Sections \labelcref{subsubsection:model,subsubsection:data}, 
yet pass a constant timestep value for each network call. 
Given $\vect{d}_{\text{obs}}$ and $\vect{c}_{1}$, InversionNet predicts 
$\tilde{\vect{c}}_0$, which is regressed towards reference velocity distribution $\vect{c}_0$
with MSE loss during training stage.  
  
Our cSGM implementation largely follows the "Improved Denoising Diffusion Probabilistic Models" article
\cite{DBLP:journals/corr/abs-2102-09672}. We employ the 1000 step cosine noise schedule proposed in the original 
paper, but opt out of learning output variance, exclusively estimating  $\vect{c}_0$ at each 
denoising iteration. Conditional inputs for SGM are both seismic data $\vect{d}_{\text{obs}}$ and 
smoothed velocity models $\vect{c}_{1}$, concatenated along channel axis. Consequently, the neural network 
architecture used in \labelcref{subsubsection:model} is altered to include additional input channel.

\subsubsection{Evaluation}
 
For the pivotal experiment, we fit deep learning models
on a single data piece obtained by merging dataset entries described in Table 
\ref{tab:openfwi_dataset_instance_summary}. As proposed in \cite{DBLP:journals/corr/abs-2111-02926},
we utilize validation data to assess inversion quality by measuring the values of MAE, MSE, 
and SSIM metrics between reconstructed and ground truth velocity models separately for 
all dataset instances included in the validation subset.   
Our experiments suggest that diffusion-based algorithms achieve the best possible values for regression metrics 
with deterministic sampling schemes; specifically, our cSGM implementation uses
DDIM sampling with $\sigma_t = 0$ proposed in \cite{song2020denoising}, while \(\text{I}^2\text{SB}\) 
replaces posteriors with their means during inference. By default, it is assumed that sampling procedures 
for cSGM and c\(\text{I}^2\text{SB}\) utilize 50 network calls each; i.e, the NFE parameter is set to 50.
A more detailed study on the impact of such parameter on inversion quality could be found in section \ref{sec:discussion}. 

\subsection{Experiment Results}

\begin{figure}[p]

    \begin{minipage}[b]{0.45\textwidth}
        \centering
        \includegraphics[width=\textwidth]{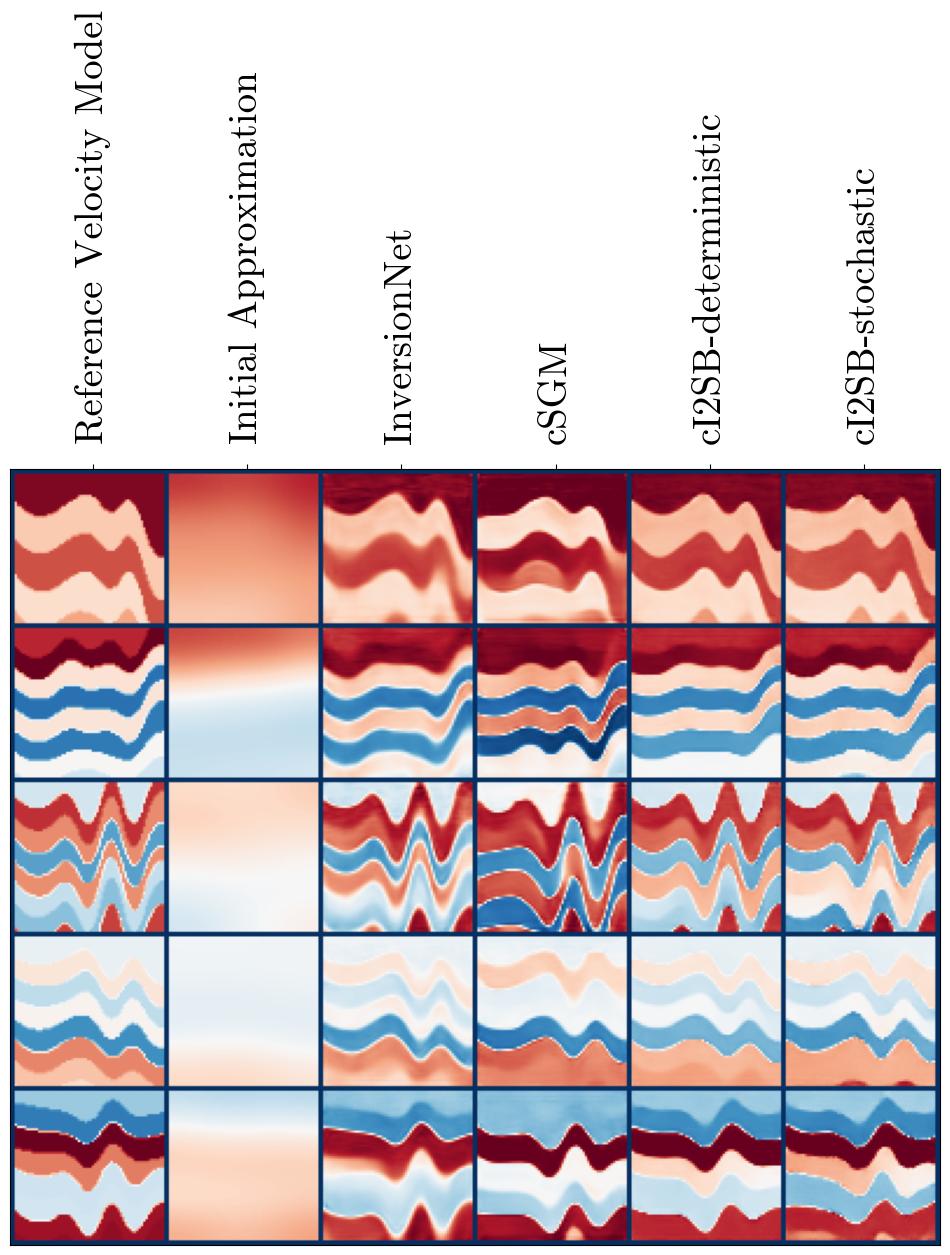}
        \subcaption{CurveVel\_B}
        \label{fig:i2sb_seismic_small_framework_comparison_curvevel_b}
    \end{minipage}
    \hfill 
    \begin{minipage}[b]{0.45\textwidth}
        \centering
        \includegraphics[width=\textwidth]{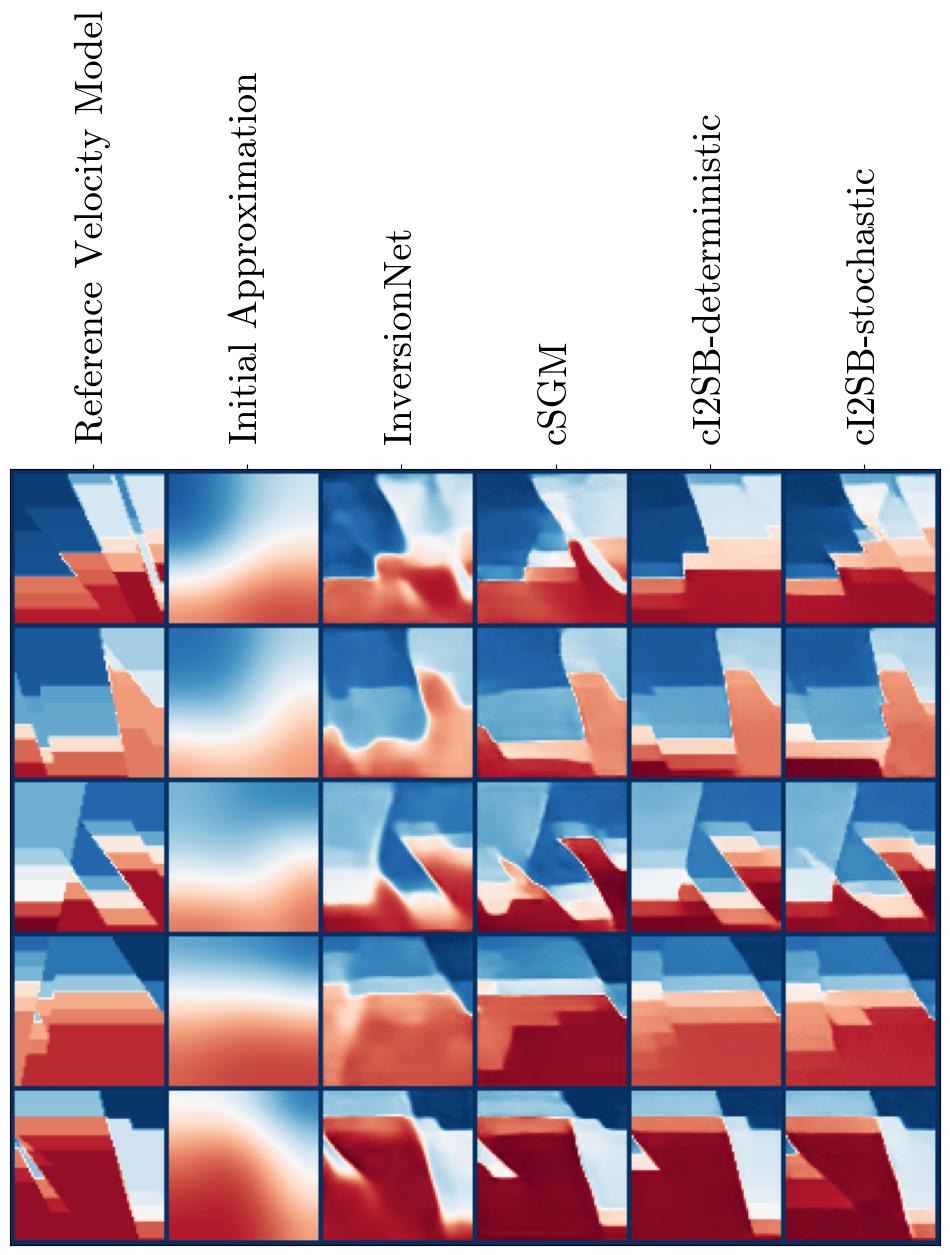}
        \subcaption{FlatFault\_B}
        \label{fig:i2sb_seismic_small_framework_comparison_flatfault_b}
    \end{minipage}

    \medskip

    \begin{minipage}[b]{0.45\textwidth}
        \centering
        \includegraphics[width=\textwidth]{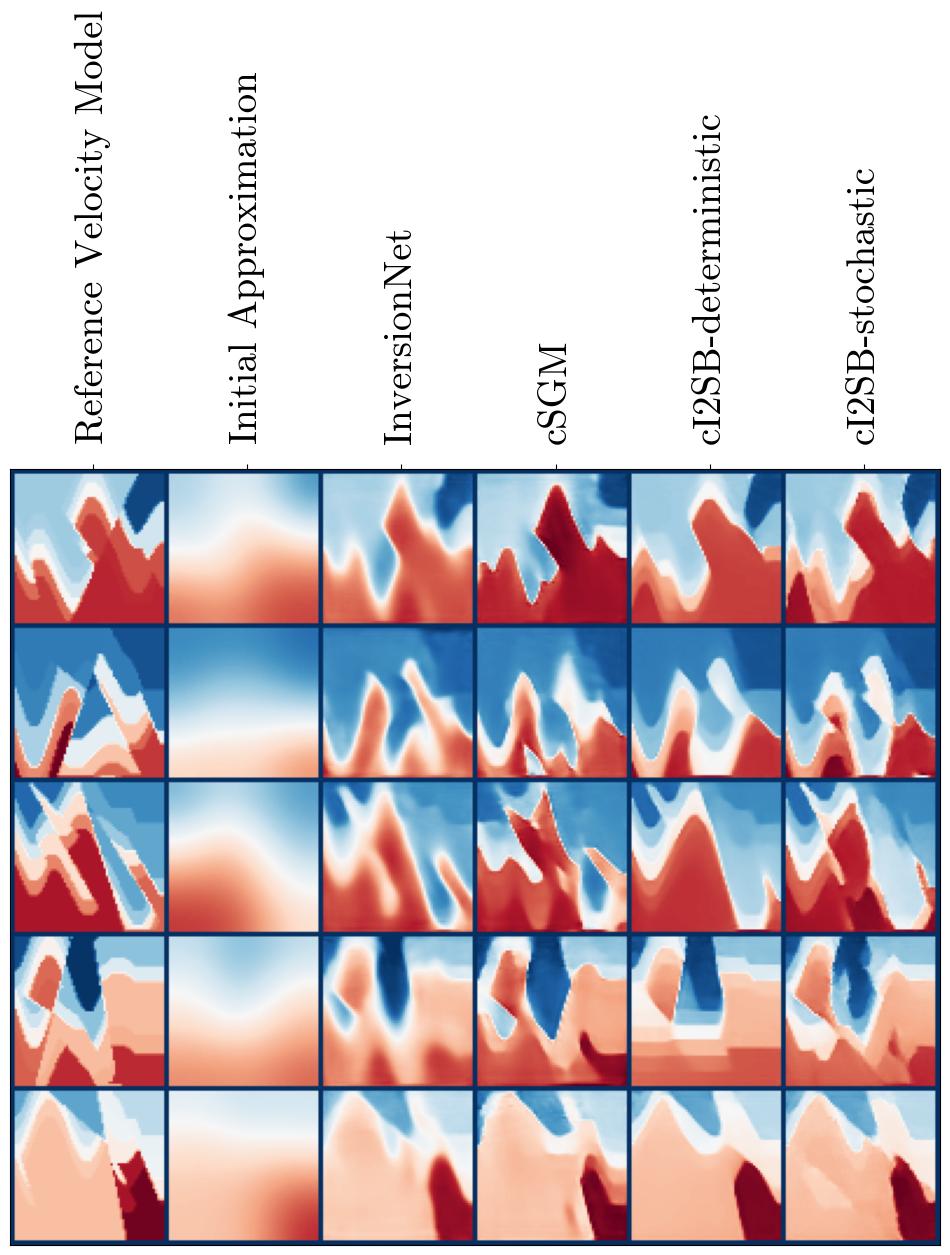}
        \subcaption{CurveFault\_B}
        \label{fig:i2sb_seismic_small_framework_comparison_curvefault_b}
    \end{minipage}
    \hfill 
    \begin{minipage}[b]{0.45\textwidth}
        \centering
        \includegraphics[width=\textwidth]{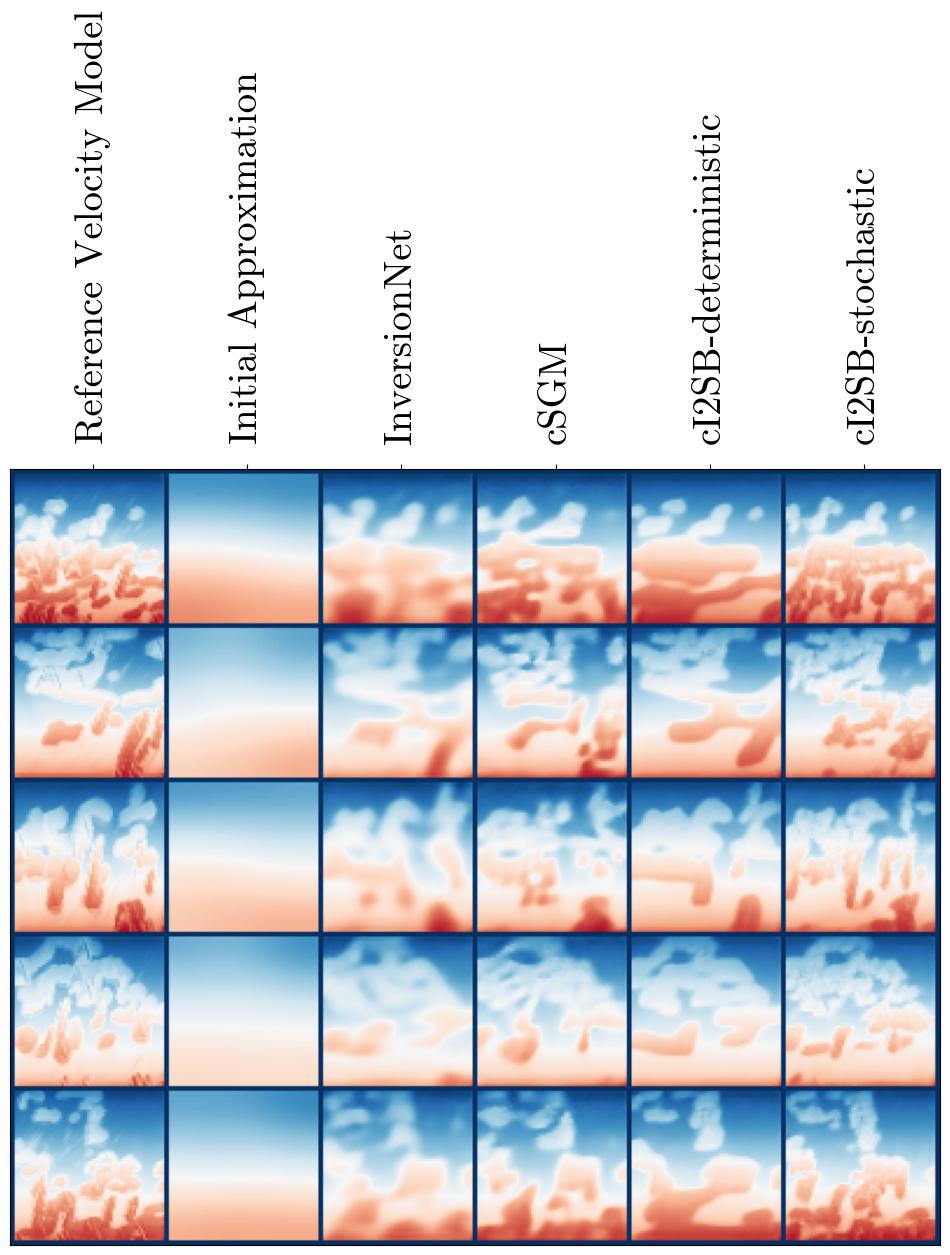}
        \subcaption{Style\_B}
        \label{fig:i2sb_seismic_small_framework_comparison_style_b}
    \end{minipage}

    \caption{ Velocity models obtained through deep-learning based seismic inversion in comparison with ground
    truth velocity models. The backbone neural network is \textbf{ci2sb-seismic-small}.}
    \label{fig:i2sb_seismic_small_framework_comparison}

\end{figure}

\begin{figure}[p]

    \begin{minipage}[b]{0.45\textwidth}
        \centering
        \includegraphics[width=\textwidth]{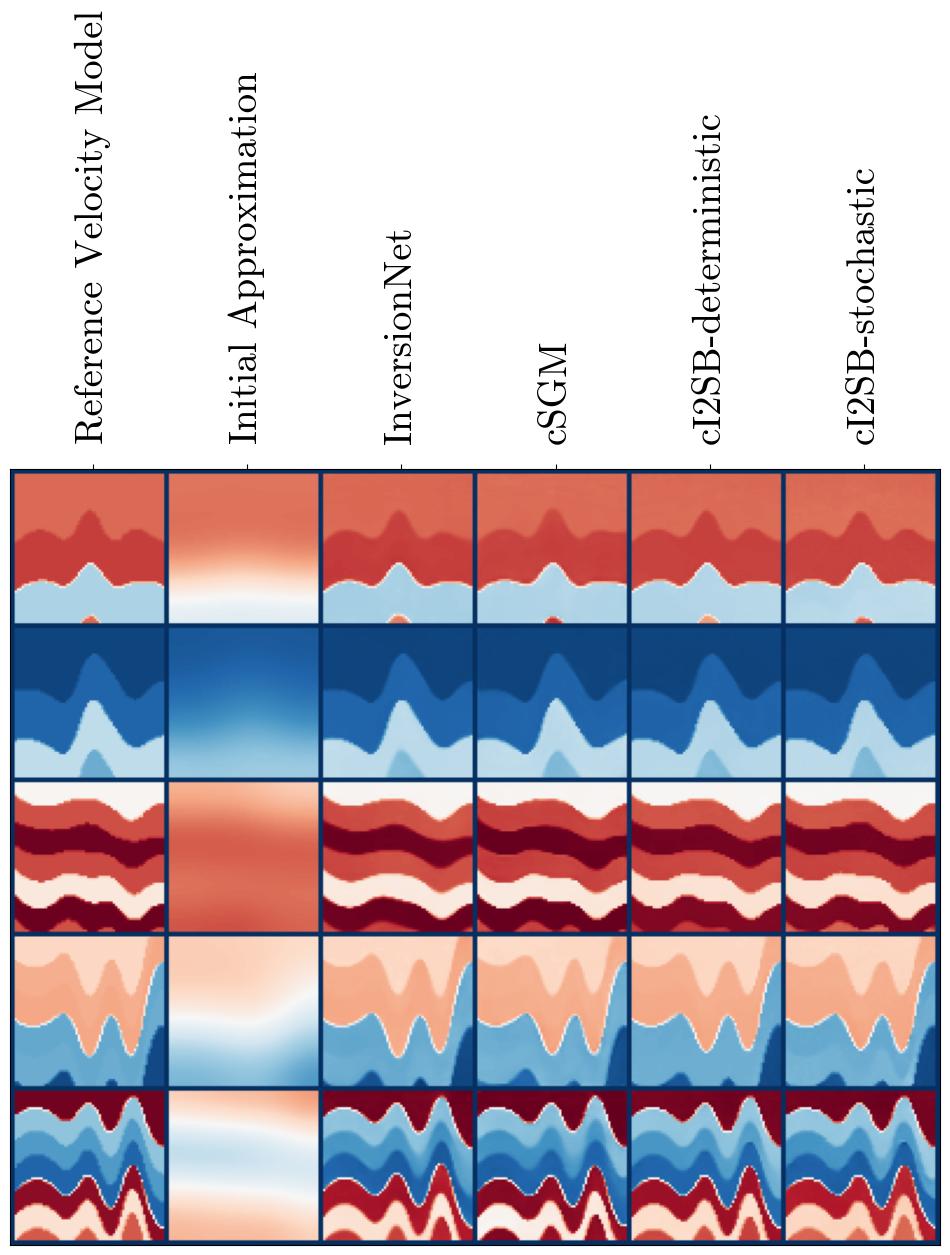}
        \subcaption{CurveVel\_B}
        \label{fig:i2sb_seismic_large_framework_comparison_curvevel_b}
    \end{minipage}
    \hfill 
    \begin{minipage}[b]{0.45\textwidth}
        \centering
        \includegraphics[width=\textwidth]{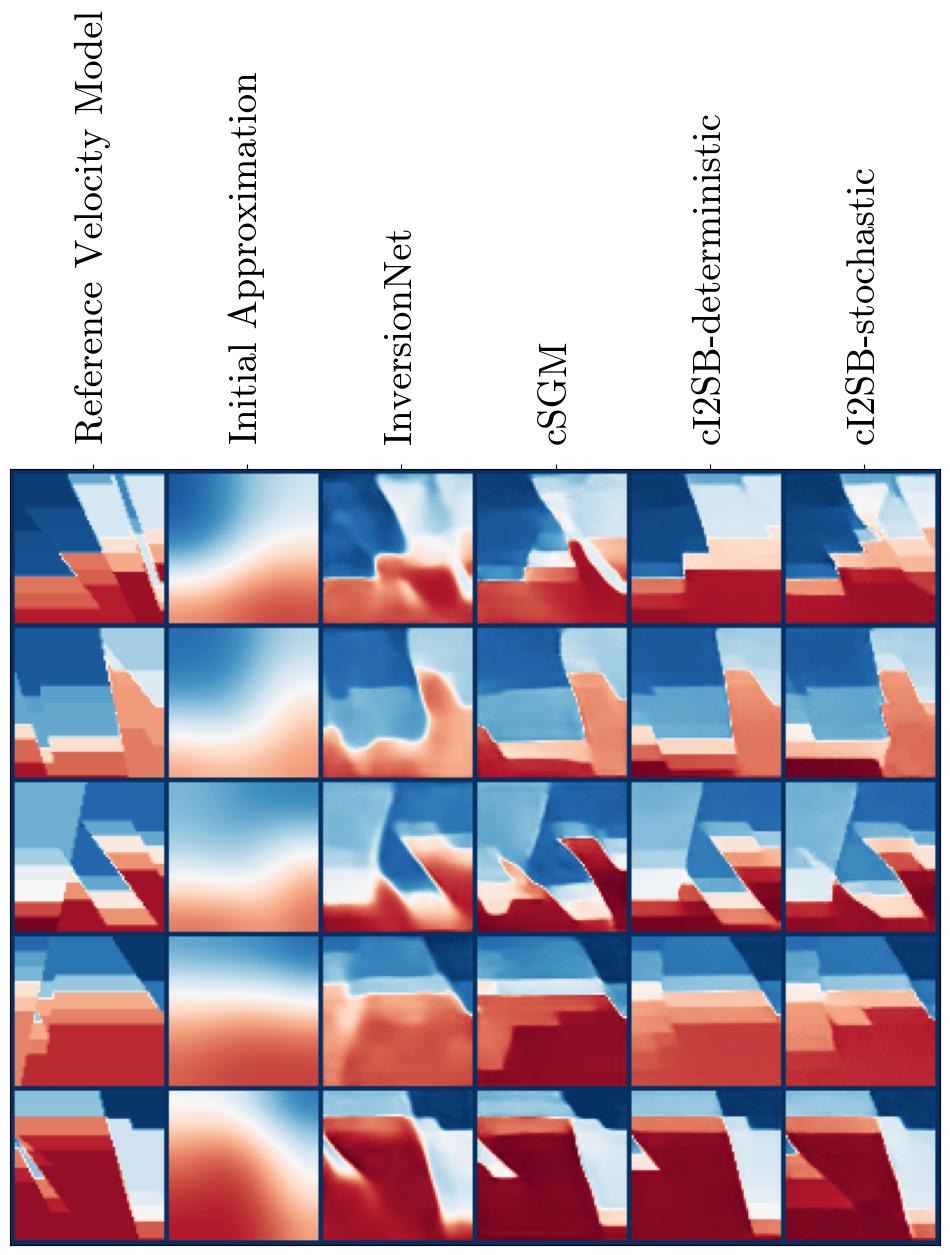}
        \subcaption{FlatFault\_B}
        \label{fig:i2sb_seismic_large_framework_comparison_flatfault_b}
    \end{minipage}

    \medskip

    \begin{minipage}[b]{0.45\textwidth}
        \centering
        \includegraphics[width=\textwidth]{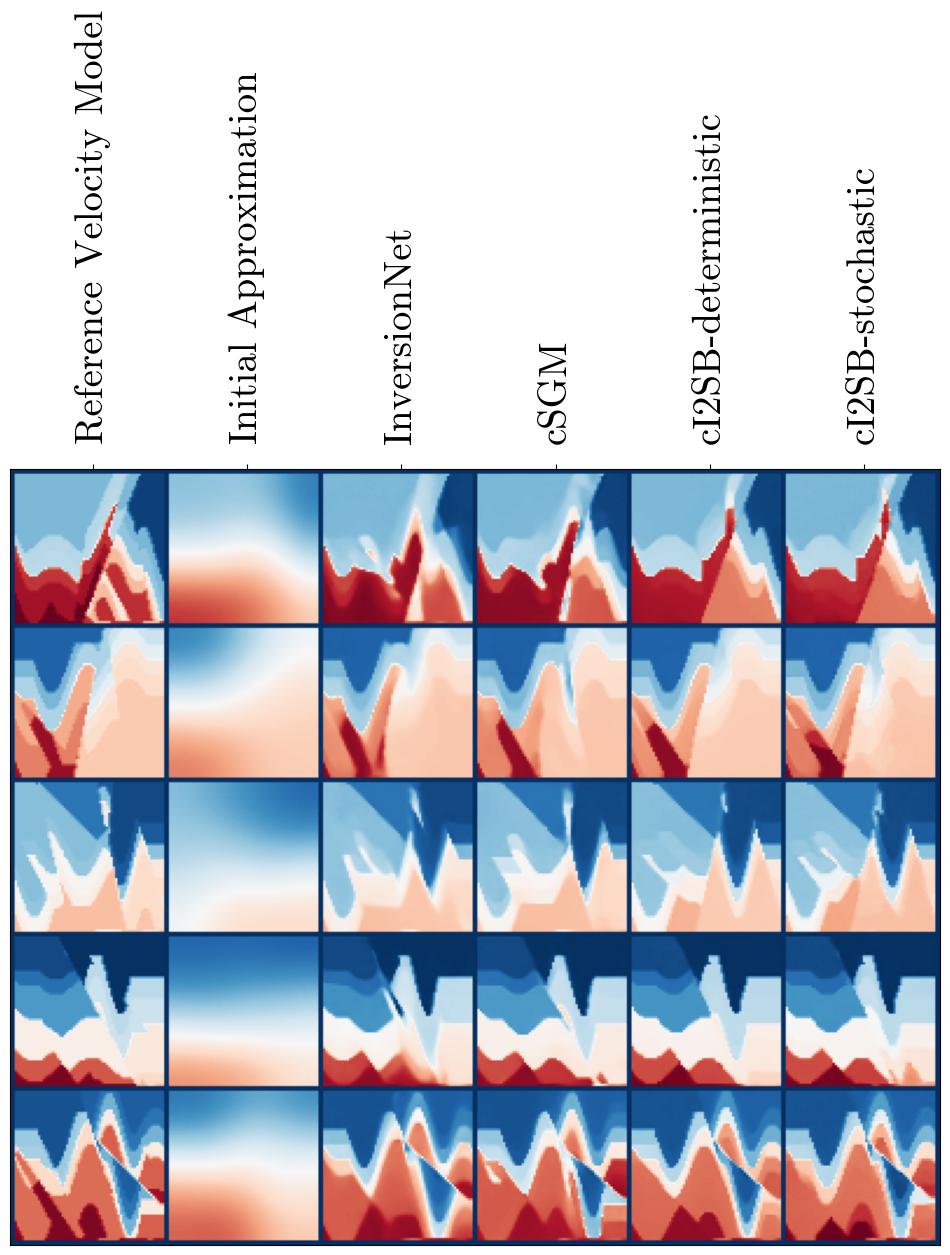}
        \subcaption{CurveFault\_B}
        \label{fig:i2sb_seismic_large_framework_comparison_curvefault_b}
    \end{minipage}
    \hfill 
    \begin{minipage}[b]{0.45\textwidth}
        \centering
        \includegraphics[width=\textwidth]{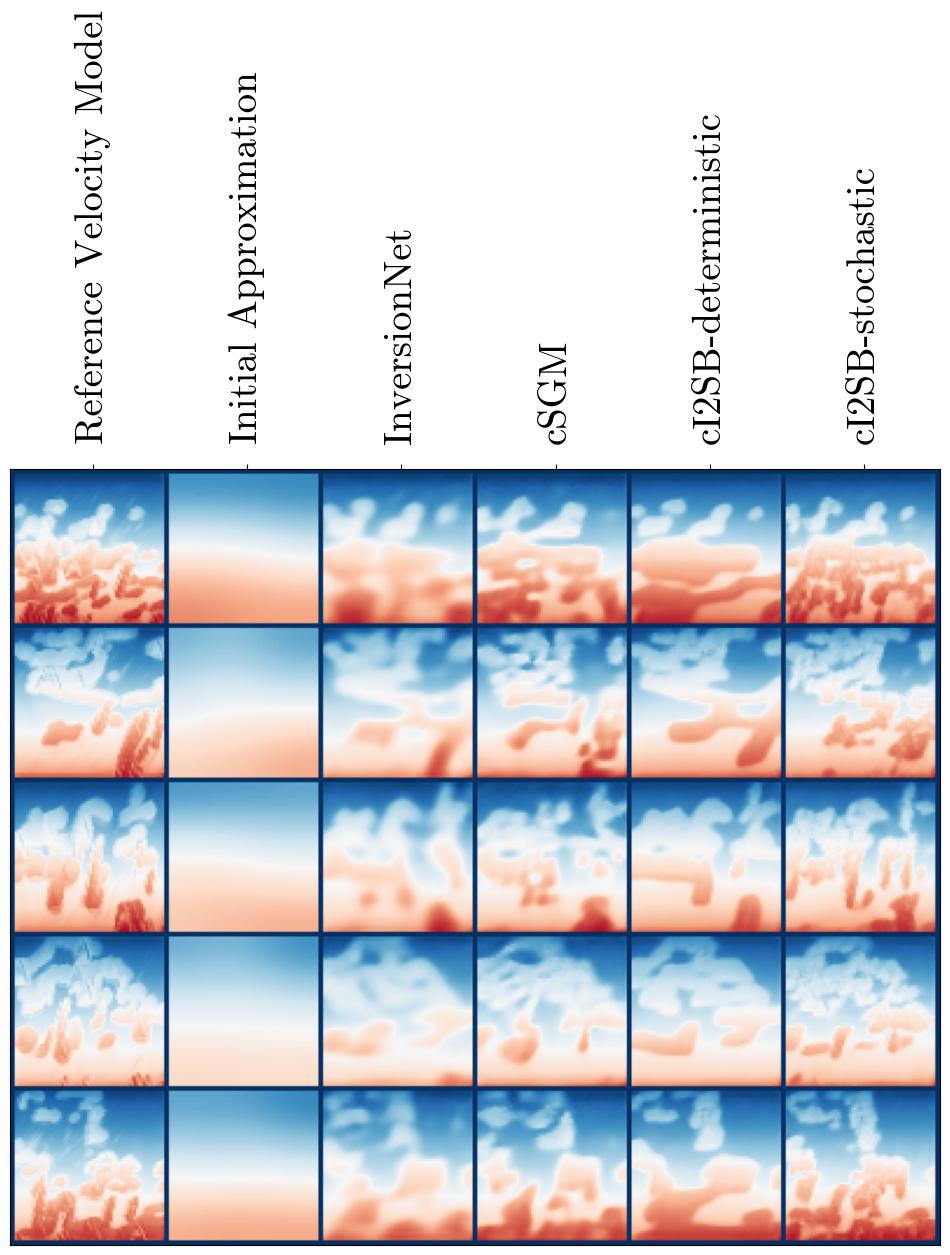}
        \subcaption{Style\_B}
        \label{fig:i2sb_seismic_large_framework_comparison_style_b}
    \end{minipage}

    \caption{ Velocity models obtained through deep-learning based seismic inversion in comparison with ground
    truth velocity models. The backbone neural network is \textbf{ci2sb-seismic-large}.}
    \label{fig:i2sb_seismic_large_framework_comparison}

\end{figure}

\begin{figure}[h!]
    \centering
    
    \begin{minipage}[b]{0.45\textwidth}
        \centering
        \includegraphics[width=\textwidth]{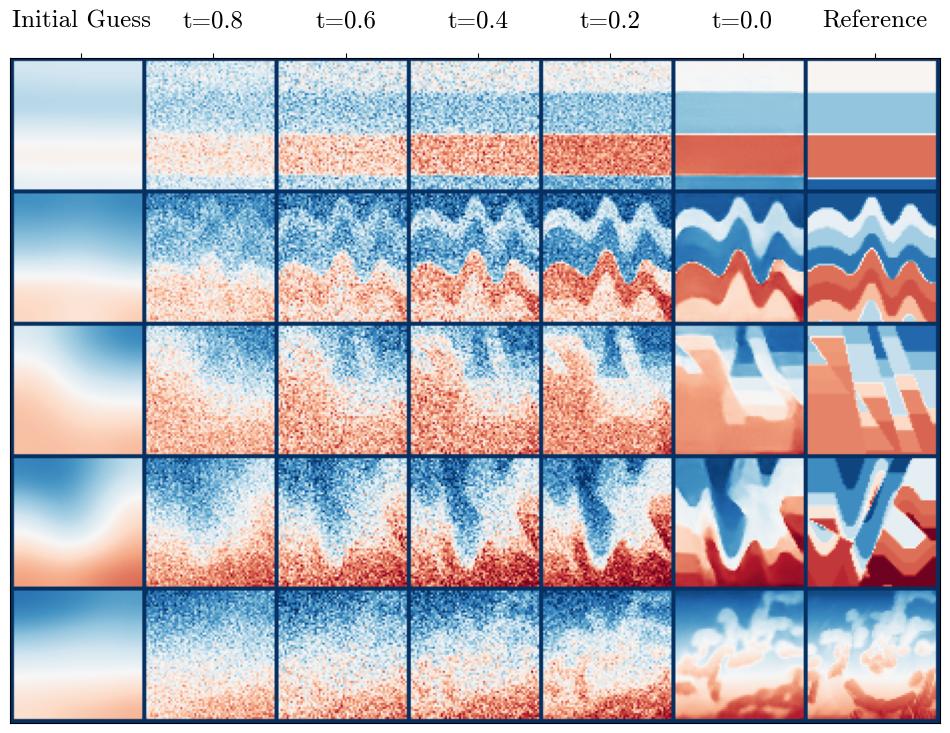}
        \subcaption{\textbf{i2sb-seismic-small}}
        \label{fig:i2sb_seismic_small_sde_trajectories}
    \end{minipage}
    \hfill
    \begin{minipage}[b]{0.45\textwidth}
        \centering
        \includegraphics[width=\textwidth]{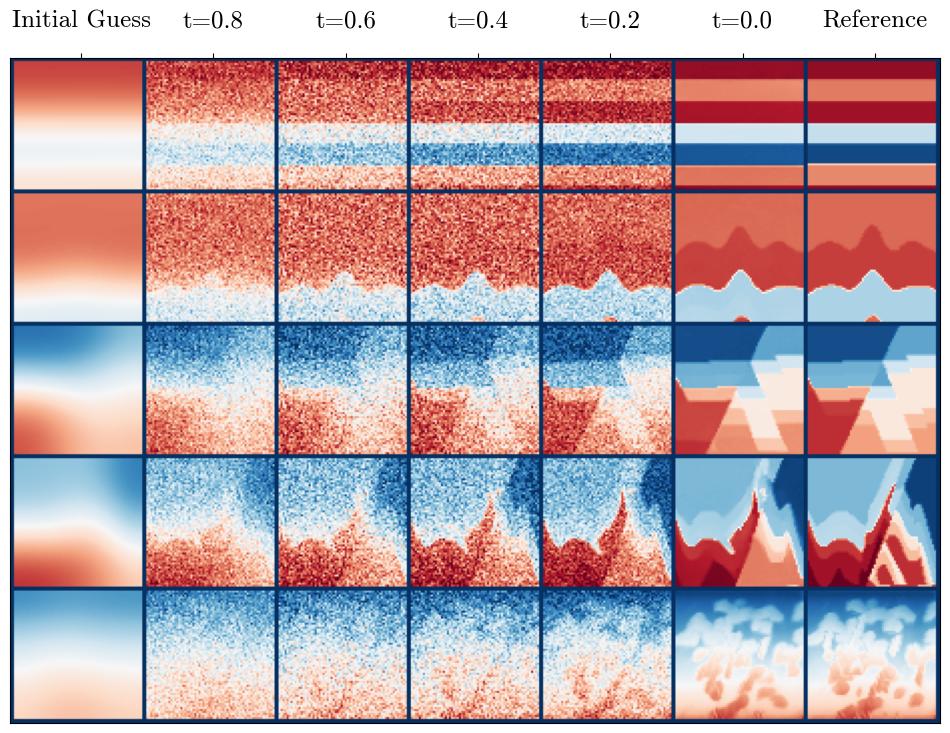}
        \subcaption{\textbf{i2sb-seismic-large}}
        \label{fig:i2sb_seismic_large_sde_trajectories}
    \end{minipage}

    \caption{
        Generative process of c\(\text{I}^2\text{SB}\) for model instances trained on the entirety of the OpenFWI collection.
    }
    \label{fig:i2sb_seismic_sde_trajectories}
\end{figure}

Tables \ref{tab:i2sb-seismic-small_performance_comparison} and \ref{tab:i2sb-seismic-large_performance_comparison} 
provide a performance assessment of \textbf{ci2sb-seismic-small} and \textbf{ci2sb-seismic-large} model instances 
trained as c\(\text{I}^2\text{SB}\) compared to counterpart frameworks. 
For completeness, we also provide results of stochastic sampling with the c\(\text{I}^2\text{SB}\) model 
while averaging the metrics values over 50 different random seeds. 
Figures \labelcref{fig:i2sb_seismic_small_framework_comparison,fig:i2sb_seismic_large_framework_comparison}
showcases difference between inversion procedures applied to a batch of data from 
validation subsets of CurveVel-B, FlatFault-B, CurveFault-B and Style-B datasets. Figure 
\labelcref{fig:i2sb_seismic_sde_trajectories} displays instances of 
stochastic c\(\text{I}^2\text{SB}\) sampling trajectories. The following takeaways summarize the observed data:  
\begin{itemize}
    \item Our reimplementation of InversionNet is a valid reference point, surpassing the performance of 
    state-of-the art solution \cite{jin2024empirical} provided the additional information given by smooth velocity models.
    \item The \textbf{ci2sb-seismic-small} model instance trained in a purely supervised fashion on 
    average performs better than both cSGM and c\(\text{I}^2\text{SB}\) with respect to the selected evaluation criteria. 
    However, our reimplementation of InversionNet is incapable of reproducing fine details 
    observed on velocity fields reconstructed with diffusion-based frameworks.  
    \item Parameter scaling favours the proposed method, as \textbf{ci2sb-seismic-large} model instance 
    trained as conditional \(\text{I}^2\text{SB}\) reduces the performance gap observed earlier,
    while maintaining texture quality that better aligns with human perception. Furthermore, stochastic 
    sampling scheme for c\(\text{I}^2\text{SB}\) was able to reproduce texture-level velocity field variations  
    characteristic of Style-B dataset. Concerning the studies that 
    utilize OpenFWI for model performance evaluation, this result, to our knowledge, is novel.
    \item In all studied cases training scheme based on c\(\text{I}^2\text{SB}\) outperforms the one based
    on cSGM by statistically significant margin.   
\end{itemize}

\begin{table}[h!]
    \caption{
        Survey of \textbf{ci2sb-seismic-small} model performance 
        within the scope of different data-driven seismic inversion 
        frameworks.
    }
    \label{tab:i2sb-seismic-small_performance_comparison}
    \centering
    \resizebox{0.9\textwidth}{!}{
        \begin{tabular}{lcccccccccccc}
            \toprule  
            & \multicolumn{3}{c}{InversionNet} 
            & \multicolumn{3}{c}{cSGM}
            & \multicolumn{3}{c}{OT-ODE sampling with c\(\text{I}^2\text{SB}\)} 
            & \multicolumn{3}{c}{Stochastic sampling with c\(\text{I}^2\text{SB}\)} 
            \\
            \cmidrule(lr){2-4} \cmidrule(lr){5-7} \cmidrule(lr){8-10} \cmidrule(lr){11-13} 
            & MAE $\downarrow$ & MSE $\downarrow$ & SSIM $\uparrow$ 
            & MAE $\downarrow$ & MSE $\downarrow$ & SSIM $\uparrow$
            & MAE $\downarrow$ & MSE $\downarrow$ & SSIM $\uparrow$
            & MAE $\downarrow$ & MSE $\downarrow$ & SSIM $\uparrow$  
            \\
            \midrule
            FlatVel\_A & \num{1.78e-3} & \num{8.36e-4} & \num{0.9782} & \num{3.96e-2} & \num{4.06e-3} & \num{0.9442} & \num{2.07e-2} & \num{1.25e-3} & \num{0.9826} & \num{3.43e-2} & \num{3.69e-3} & \num{0.9564} \\
            FlatVel\_B & \num{3.57e-2} & \num{4.26e-3} & \num{0.9484} & \num{9.27e-2} & \num{2.55e-2} & \num{0.8419} & \num{5.24e-2} & \num{1.25e-2} & \num{0.9150} & \num{7.23e-2} & \num{2.04e-2} & \num{0.8731} \\
            CurveVel\_A & \num{3.66e-2} & \num{3.79e-3} & \num{0.9224} & \num{6.23e-2} & \num{1.11e-2} & \num{0.8529} & \num{4.31e-2} & \num{5.89e-3} & \num{0.9227} & \num{6.29e-2} & \num{1.25e-2} & \num{0.8708} \\
            CurveVel\_B & \num{7.47e-2} & \num{1.75e-3} & \num{0.8597} & \num{1.36e-1} & \num{5.48e-2} & \num{0.7184} & \num{9.71e-2} & \num{3.30e-2} & \num{0.8102} & \num{1.20e-1} & \num{4.77e-2} & \num{0.7563} \\
            FlatFault\_A & \num{1.72e-2} & \num{1.01e-3} & \num{0.9825} & \num{3.99e-2} & \num{4.90e-3} & \num{0.9613} & \num{2.07e-2} & \num{1.44e-3} & \num{0.9839} & \num{2.94e-2} & \num{3.24e-3} & \num{0.9701} \\
            FlatFault\_B & \num{5.68e-2} & \num{8.69e-3} & \num{0.8586} & \num{1.01e-1} & \num{2.67e-2} & \num{0.7469} & \num{7.37e-2} & \num{1.55e-2} & \num{0.8240} & \num{9.44e-2} & \num{2.42e-2} & \num{0.7611} \\
            CurveFault\_A & \num{2.37e-2} & \num{2.07e-3} & \num{0.9681} & \num{4.38e-2} & \num{7.18e-3} & \num{0.9333} & \num{3.01e-2} & \num{3.37e-3} & \num{0.9666} & \num{4.04e-2} & \num{6.35e-3} & \num{0.9439} \\
            CurveFault\_B & \num{8.86e-2} & \num{1.89e-3} & \num{0.7778} & \num{1.42e-1} & \num{4.73e-2} & \num{0.6300} & \num{1.21e-1} & \num{3.40e-2} & \num{0.7083} & \num{1.44e-1} & \num{4.97e-2} & \num{0.6336} \\
            Style\_A & \num{3.80e-2} & \num{3.27e-3} & \num{0.9370} & \num{6.74e-2} & \num{1.01e-2} & \num{0.8562} & \num{5.10e-2} & \num{5.91e-3} & \num{0.9100} & \num{7.40e-2} & \num{1.25e-2} & \num{0.8531} \\
            Style\_B & \num{5.17e-2} & \num{5.34e-3} & \num{0.8011} & \num{8.00e-2} & \num{1.34e-2} & \num{0.6733} & \num{5.89e-2} & \num{7.81e-3} & \num{0.7748} & \num{7.74e-2} & \num{1.30e-2} & \num{0.6853} \\
        
            \bottomrule
        \end{tabular}
    }
\end{table}

\begin{table}[h!]
    \caption{
        Survey of \textbf{ci2sb-seismic-large} model performance 
        within the scope of different data-driven seismic inversion 
        frameworks.
    }
    \label{tab:i2sb-seismic-large_performance_comparison}
    \centering
    \resizebox{0.9\textwidth}{!}{
        \begin{tabular}{lcccccccccccc}
            \toprule  
            & \multicolumn{3}{c}{InversionNet} 
            & \multicolumn{3}{c}{cSGM}
            & \multicolumn{3}{c}{OT-ODE sampling with c\(\text{I}^2\text{SB}\)} 
            & \multicolumn{3}{c}{Stochastic sampling with c\(\text{I}^2\text{SB}\)} 
            \\
            \cmidrule(lr){2-4} \cmidrule(lr){5-7} \cmidrule(lr){8-10} \cmidrule(lr){11-13} 
            & MAE $\downarrow$ & MSE $\downarrow$ & SSIM $\uparrow$ 
            & MAE $\downarrow$ & MSE $\downarrow$ & SSIM $\uparrow$
            & MAE $\downarrow$ & MSE $\downarrow$ & SSIM $\uparrow$
            & MAE $\downarrow$ & MSE $\downarrow$ & SSIM $\uparrow$    
            \\
            \midrule
            FlatVel\_A & \num{3.87e-3} & \num{4.19e-5} & \num{0.9990} & \num{6.57e-3} & \num{1.11e-4} & \num{0.9961} & \num{6.62e-3} & \num{1.56e-4} & \num{0.9991} & \num{6.75e-3} & \num{1.57e-4} & \num{0.9977} \\
            FlatVel\_B & \num{1.03e-3} & \num{8.06e-4} & \num{0.9916} & \num{2.01e-2} & \num{2.31e-3} & \num{0.9787} & \num{1.72e-2} & \num{3.01e-3} & \num{0.9796} & \num{2.07e-2} & \num{3.65e-3} & \num{0.9745} \\
            CurveVel\_A & \num{1.26e-2} & \num{9.53e-4} & \num{0.9803} & \num{1.61e-2} & \num{1.46e-3} & \num{0.9713} & \num{1.25e-2} & \num{1.04e-3} & \num{0.9825} & \num{1.60e-2} & \num{1.48e-3} & \num{0.9752} \\
            CurveVel\_B & \num{3.40e-2} & \num{7.56e-3} & \num{0.9475} & \num{4.26e-2} & \num{1.01e-2} & \num{0.9320} & \num{3.50e-2} & \num{9.21e-3} & \num{0.9420} & \num{4.16e-2} & \num{1.20e-2} & \num{0.9269} \\
            FlatFault\_A & \num{5.38e-3} & \num{1.39e-4} & \num{0.9968} & \num{8.05e-3} & \num{3.03e-4} & \num{0.9933} & \num{6.62e-3} & \num{1.56e-4} & \num{0.9973} & \num{8.62e-3} & \num{2.48e-3} & \num{0.9955} \\
            FlatFault\_B & \num{2.76e-2} & \num{3.96e-3} & \num{0.9356} & \num{3.29e-2} & \num{5.42e-3} & \num{0.9154} & \num{2.53e-2} & \num{3.70e-3} & \num{0.9467} & \num{3.14e-2} & \num{4.85e-3} & \num{0.9321} \\
            CurveFault\_A & \num{7.76e-3} & \num{3.64e-4} & \num{0.9932} & \num{1.07e-2} & \num{6.72e-4} & \num{0.9898} & \num{8.86e-3} & \num{3.58e-4} & \num{0.9944} & \num{1.13e-2} & \num{5.48e-3} & \num{0.9913} \\
            CurveFault\_B & \num{6.48e-2} & \num{1.41e-2} & \num{0.8343} & \num{7.07e-2} & \num{1.74e-2} & \num{0.8121} & \num{6.06e-2} & \num{1.36e-2} & \num{0.8542} & \num{7.06e-2} & \num{1.72e-2} & \num{0.8241} \\
            Style\_A & \num{2.28e-2} & \num{1.52e-3} & \num{0.9681} & \num{2.91e-2} & \num{2.35e-3} & \num{0.9537} & \num{2.61e-2} & \num{2.04e-3} & \num{0.9672} & \num{3.50e-2} & \num{3.60e-3} & \num{0.9480} \\
            Style\_B & \num{3.41e-2} & \num{2.86e-3} & \num{0.8832} & \num{4.35e-2} & \num{4.93e-3} & \num{0.8260} & \num{3.35e-2} & \num{3.18e-3} & \num{0.8904} & \num{4.24e-2} & \num{4.83e-3} & \num{0.8477} \\
        
            \bottomrule
        \end{tabular}
    }
\end{table}

\newpage

\section{Discussion} \label{sec:discussion}

{
\subsection{Objective reweighting}

The use of sampling scheme \ref{alg:guided_conditional_i2sb_sampling} for \textbf{ci2sb-seismic-small} model instance trained with alg. \ref{alg:guided_conditional_i2sb_training} 
while $w_\text{cond}$ is set 0 leads to the inferior performance compared to both InversionNet, cSGM, and the same model instance trained with alg. \ref{alg:conditional_i2sb_training}.
However, applying baseline classifier-free guidance to cSGM does not affect its performance in the same way, as the decay in values of quality metrics 
for such framework is less significant (fig. \ref{fig:impact_of_objective_reweighting}).
Therefore, we believe that the reasoning for performance drop lies in prior distribution at terminal diffusion point shifting from pure gaussian 
to the mixture of the ones induced with degradation operator.

To rationalize the observed phenomenon, we revisit the joint c\(\text{I}^2\text{SB}\) training scheme. At each step neural network is asked to reconstruct the initial 
point $\vect{c}_0$ of diffusion bridge given either noisy sample $\vect{c}_t$, or $\vect{c}_t$ combined with conditional 
input $\vect{d}_{\text{obs}}$. By model design, $\vect{c}_t$ is close to $\vect{c}_0$, meanwhile the mapping between $\vect{d}_{\text{obs}}$ and $\vect{c}_0$
might be arbitrarily complex. As the target variable for score matching is independent of $\vect{d}_{\text{obs}}$, the natural course of action
for neural network when $w_\text{cond} = 0$ is to \textit{pay less attention to the conditional input}, inferring the regression target mostly from $\vect{c}_t$.
Indeed, in the case above the gap between evaluation metrics calculated with velocity models obtained through conditional and unconditional sampling is minor. 
However, as seen in fig. \ref{fig:impact_of_objective_reweighting}, explicitly nudging joint c\(\text{I}^2\text{SB}\) training procedure towards conditional evaluation via 
objective reweighting solves the issues, bringing model's performance in line with other solutions at the 
cost of the quality of unconditional evaluation. 
}

{
\subsection{Sample fidelity and regression metrics}

The number of NFEs required to obtain samples with high image fidelity represents a critical 
hyperparameter in diffusion-based inference, determining its total computational cost.
Since better performance at low NFE regimes is the desired property of such algorithms, 
we study impact of utilizing more NFEs for conditional stochastic \(\text{I}^2\text{SB}\) sampling 
on the value of regression metrics proposed for model evaluation. 
Results of our findings for the \textbf{ci2sb-seismic-large} model instance described 
in \ref{subsubsection:model} are presented with tables \ref{tab:i2sb_performance_nfe_p1} and 
\ref{tab:i2sb_performance_nfe_p2}.

\begin{table}[h!]
    \caption{The dependency of validation metrics estimated c\(\text{I}^2\text{SB}\) from the amount of NFEs, part 1}
    \label{tab:i2sb_performance_nfe_p1}
    \centering
    \resizebox{0.9\textwidth}{!}{
        \begin{tabular}{lccccccccc}
            \toprule  
            & \multicolumn{3}{c}{NFE=1} 
            & \multicolumn{3}{c}{NFE=2}
            & \multicolumn{3}{c}{NFE=5}  
            \\
            \cmidrule(lr){2-4} \cmidrule(lr){5-7} 
            \cmidrule(lr){8-10} 
            & MAE $\downarrow$ & MSE $\downarrow$ & SSIM $\uparrow$ 
            & MAE $\downarrow$ & MSE $\downarrow$ & SSIM $\uparrow$
            & MAE $\downarrow$ & MSE $\downarrow$ & SSIM $\uparrow$
            \\
            \midrule
            FlatVel\_A & \num{4.22e-3} & \num{4.24e-5} & \num{0.9989} & \num{4.35e-3} & \num{4.74e-5} & \num{0.9991} & \num{4.74e-3} & \num{6.02e-5} & \num{0.9991} \\
            FlatVel\_B & \num{1.63e-2} & \num{1.71e-2} & \num{0.9841} & \num{1.46e-2} & \num{1.84e-2} & \num{0.9845} & \num{1.54e-2} & \num{1.99e-3} & \num{0.9836} \\
            CurveVel\_A & \num{1.36e-2} & \num{9.90e-4} & \num{0.9807} & \num{1.38e-2} & \num{1.06e-4} & \num{0.9806} & \num{1.47e-2} & \num{1.26e-3} & \num{0.9783} \\
            CurveVel\_B & \num{3.87e-2} & \num{8.32e-3} & \num{0.9432} & \num{3.64e-2} & \num{8.99e-3} & \num{0.9427} & \num{3.81e-2} & \num{1.02e-3} & \num{0.9364} \\
            FlatFault\_A & \num{5.49e-3} & \num{1.28e-4} & \num{0.9969} & \num{5.93e-3} & \num{1.39e-4} & \num{0.9970} & \num{6.60e-3} & \num{1.71e-4} & \num{0.9973} \\
            FlatFault\_B & \num{2.75e-2} & \num{3.57e-3} & \num{0.9429} & \num{2.82e-2} & \num{3.90e-3} & \num{0.9406} & \num{3.06e-2} & \num{4.62e-3} & \num{0.9326} \\
            CurveFault\_A & \num{7.84e-3} & \num{3.35e-4} & \num{0.9935} & \num{8.35e-3} & \num{3.55e-4} & \num{0.9937} & \num{9.38e-3} & \num{4.34e-4} & \num{0.9930} \\
            CurveFault\_B & \num{6.36e-2} & \num{1.30e-2} & \num{0.8507} & \num{6.40e-2} & \num{1.37e-2} & \num{0.8472} & \num{6.84e-2} & \num{1.57e-2} & \num{0.8321} \\
            Style\_A & \num{2.62e-2} & \num{2.01e-3} & \num{0.9643} & \num{2.83e-2} & \num{2.33e-3} & \num{0.9608} & \num{3.20e-2} & \num{3.01e-3} & \num{0.9534} \\
            Style\_B & \num{3.55e-2} & \num{3.04e-3} & \num{0.8813} & \num{3.72e-2} & \num{3.50e-3} & \num{0.8726} & \num{4.03e-2} & \num{4.29e-3} & \num{0.8564} \\
        
            \bottomrule
        \end{tabular}
    }
\end{table}

\begin{table}[h!]
    \caption{The dependency of validation metrics estimated with c\(\text{I}^2\text{SB}\) from the amount of NFEs, part 2}
    \label{tab:i2sb_performance_nfe_p2}
    \centering
    \resizebox{0.9\textwidth}{!}{
        \begin{tabular}{lccccccccc}
            \toprule  
            & \multicolumn{3}{c}{NFE=10} 
            & \multicolumn{3}{c}{NFE=20} 
            & \multicolumn{3}{c}{NFE=50} 
            \\
            \cmidrule(lr){2-4} \cmidrule(lr){5-7} \cmidrule(lr){8-10} 
            & MAE $\downarrow$ & MSE $\downarrow$ & SSIM $\uparrow$ 
            & MAE $\downarrow$ & MSE $\downarrow$ & SSIM $\uparrow$
            & MAE $\downarrow$ & MSE $\downarrow$ & SSIM $\uparrow$
            \\
            \midrule
            FlatVel\_A & \num{5.04e-3} & \num{6.80e-5} & \num{0.9989} & \num{5.42e-3} & \num{7.71e-5} & \num{0.9986} & \num{5.86e-3} & \num{8.28e-5} & \num{0.9981} \\
            FlatVel\_B & \num{1.61e-2} & \num{2.08e-3} & \num{0.9826} & \num{1.69e-2} & \num{2.19e-3} & \num{0.9817} & \num{1.76e-2} & \num{2.37e-3} & \num{0.9807} \\
            CurveVel\_A & \num{1.54e-2} & \num{1.40e-3} & \num{0.9765} & \num{1.59e-2} & \num{1.48e-3} & \num{0.9751} & \num{1.66e-2} & \num{1.57e-3} & \num{0.9732} \\
            CurveVel\_B & \num{3.97e-2} & \num{1.10e-2} & \num{0.9317} & \num{4.09e-2} & \num{1.16e-2} & \num{0.9286} & \num{4.32e-2} & \num{1.21e-2} & \num{0.9255} \\
            FlatFault\_A & \num{7.04e-3} & \num{1.90e-4} & \num{0.9965} & \num{7.49e-3} & \num{2.11e-4} & \num{0.9962} & \num{8.08e-3} & \num{2.33e-4} & \num{0.9954} \\
            FlatFault\_B & \num{3.24e-2} & \num{5.08e-3} & \num{0.9278} & \num{3.37e-2} & \num{5.42e-3} & \num{0.9238} & \num{3.46e-3} & \num{5.64e-3} & \num{0.9210} \\
            CurveFault\_A & \num{9.92e-3} & \num{4.76e-4} & \num{0.9925} & \num{1.04e-2} & \num{5.15e-4} & \num{0.9919} & \num{1.10e-2} & \num{5.53e-4} & \num{0.9909} \\
            CurveFault\_B & \num{7.16e-2} & \num{1.71e-2} & \num{0.8219} & \num{7.41e-2} & \num{1.83e-2} & \num{0.8137} & \num{7.63e-2} & \num{1.93e-2} & \num{0.8068} \\
            Style\_A & \num{3.44e-2} & \num{3.50e-3} & \num{0.9480} & \num{3.61e-2} & \num{3.86e-3} & \num{0.9440} & \num{3.77e-2} & \num{4.16e-3} & \num{0.9403} \\
            Style\_B & \num{4.29e-2} & \num{4.79e-3} & \num{0.8457} &  \num{4.40e-2} & \num{5.17e-3} & \num{0.8371} & \num{4.59e-2} & \num{5.57e-3} & \num{0.8271} \\
        
            \bottomrule
        \end{tabular}
    }
\end{table}

Figure \labelcref{fig:i2sb_nfe_grid_datasets} depicts batches from validation 
subsets of CurveVel\_B, FlatFault\_B, CurveFault\_B, and Style\_B datasets respectively, 
along with reconstructed data obtained through c\(\text{I}^2\text{SB}\) sampling procedure 
in different NFE regimes. For illustrative purposes we extend the range for NFEs up to 
1000. As the evidence shows, despite the noticeable difference in perceptual 
quality, utilizing more NFEs per inference run does not necessarily lead to higher value of
suggested evaluation metrics. On the contrary, the average value of MAE and MSE 
on the validation set increases with additional neural network calls, 
whereas SSIM steadily decreases. This observation aligns with the perception-distortion
tradeoff \cite{blau2018perception} -- the characteristic previously observed in various 
 generative modelling frameworks, including bridge-like models 
\cite{delbracio2023inversion}.

We believe that perception-distortion tradeoff achieved with c\(\text{I}^2\text{SB}\) inference 
could be credited to bias amplification in sampling scheme, as velocity model estimates through denoising timesteps utilize
neural network with the shared set of parameters. Figure \ref{fig:i2sb_nfe_grid_slices} illustrates the phenomenon 
through vertical slices of velocity models presented in Figure \ref{fig:i2sb_nfe_grid_datasets} 
at fixed grid nodes, revealing that consecutive \(\text{I}^2\text{SB}\) iterations can significantly 
shift velocity profiles of predicted $\vect{c}_0$ away from ground truth. In comparison, one-step
estimates of $\vect{c}_0$ with the same network given the timestep $t$ and the intermediate point $\vect{c}_t$ 
sampled with \eqref{eq:sb_posterior_analytical_form} are increasingly more accurate at the beginning of bridge
(fig. \ref{fig:i2sb_large_single_shot_evaluations_different_timesteps}).
Further research could be directed towards mitigating such issue, however, it is out of the scope of the current study.  

In addition, we present empirical evidence to support the claim that Schrödinger Bridge-based models achieve high perceptual quality of samples 
with fewer NFEs compared to SGM-based models. Namely, we illustrate several batches of reconstructed velocity fields from validation subset 
with NFEs ranging from 1 to 1000 (fig \ref{fig:ddpm_nfe_grid}). The difference in perceptual quality is most noticeable 
for samples from Style-B subset, since in this case inference process for cSGM is unable to capture the texture-level details until NFE parameter 
is set to 100 or higher. In comparison, high-frequency detailing of velocity fields reconstructed with c\(\text{I}^2\text{SB}\) is observable 
starting from NFE=5. 

\subsection{Classifier-free Guidance in application to c\(\text{I}^2\text{SB}\)}

Our experiments demonstrate that inference procedure \ref{alg:guided_conditional_i2sb_sampling} for c\(\text{I}^2\text{SB}\) model trained with algorithm 
\ref{alg:guided_conditional_i2sb_training} provides an additional degree of control over the variance of output 
samples through the guidance scale parameter $\eta$. To demonstrate that, we display velocity models recovered from distribution of initial guesses 
obtained through $p_{\text{prior}}$ with the same reference velocity model depending on value of $\eta$ (fig. \ref{fig:guidance_scale_grids}). 

The diversity in reconstructed velocity models shows an inverse correlation with $\eta$ values - for instance, setting $\eta=0.2$ 
results in more diverse output compared to $\eta=0.8$. As the value of guidance scale approaches $\eta=1$, i.e., for purely conditional inference, 
modes of output distributions "collapse", resulting in nearly constant reconstruction outcome regardless of initial guess. 
At the same time, with $\eta$ set to $0$, which is equivalent to \(\text{I}^2\text{SB}\) sampling, model output exhibits the most significant 
degree of variance. Hence, we substantiate our claim about guidance scale acting as an equivalent of inverse temperature 
for GAN output manipulation.

\subsection{Limitations}

From the perspective of data-related issues, the major open challenges in the field of deep learning-based FWI 
are the inherent noise in seismic data and transferability of models between different basins of geological structures and acquisition geometries
\cite{mousavi2024applications}. The first issue can be partially addressed with the suggested framework, since the 
joint prior distribution can be selected with noisy observations and smoothed velocty models in mind. 
However, the implementation of classifier-free guidance relies on the constant shape of the conditioning vector,
thus, only exacerbating the second issue, as the representation of seismograms greatly varies depending on acquisition setup. 

Another limiting factor of the developed framework is its reliance on the explicit design of distortion operator 
$\mathcal{D}_k^{\gamma}$. Hence, the quality of inversion is subject to degradation if inference distribution of inital guesses 
differs from the one used for model training. To illustrate the case, we repeat the c\(\text{I}^2\text{SB}\) evaluation 
procedure for two different operators $\hat{\mathcal{D}}_k^{\gamma}$ and $\tilde{\mathcal{D}}_k^{\gamma}$:
for the first one $k$ belongs in the integer range from 16 to 24, whereas for the second one the same parameter varies 
from 0 to 8 (table \ref{tab:ci2sb-generalization-ood}). The results of the proposed experiment reveal 
that the inferior value of metrics on the out-of-distribution smooth models is related to the distribution shift per se. Indeed, initial 
guesses obtained through the second distortion operator are arguably more informative than those used during parameter tuning, yet 
c\(\text{I}^2\text{SB}\)-based inversion performs better when smooth models align with samples from training distribution.
Thus, despite the developed procedure having improved theoretical grounding compared to previous iterations of data-driven algorithms, 
it stil lacks robustness of classical waveform inversion pipelines. 

\begin{table}[h!]
    \centering
    \resizebox{0.9\textwidth}{!}{
        \begin{tabular}{lC{3cm}C{3cm}C{3cm}C{3cm}C{3cm}C{3cm}}
            \toprule  
            & \multicolumn{3}{c}{c\(\text{I}^2\text{SB}\) with initial guesses obtained through $\hat{\mathcal{D}}_k^{\gamma}$} 
            & \multicolumn{3}{c}{c\(\text{I}^2\text{SB}\) with initial guesses obtained through $\tilde{\mathcal{D}}_k^{\gamma}$} 
            \\
            \cmidrule(lr){2-4}\cmidrule(lr){5-7}
            & MAE $\downarrow$ & MSE $\downarrow$ & SSIM $\uparrow$ 
            & MAE $\downarrow$ & MSE $\downarrow$ & SSIM $\uparrow$ 
            \\
            \midrule
            FlatVel\_A & \num{2.24e-2} & \num{1.84e-3} & \num{0.9853} & \num{1.12e-2} & \num{6.63e-4} & \num{0.9952} \\
            FlatVel\_B & \num{4.91e-2} & \num{1.50e-2} & \num{0.9265} & \num{5.97e-2} & \num{1.37e-2} & \num{0.9292} \\
            CurveVel\_A & \num{3.02e-2} & \num{3.40e-3} & \num{0.9672} & \num{2.51e-2} & \num{3.50e-3} & \num{0.9669} \\
            CurveVel\_B & \num{6.27e-2} & \num{2.07e-3} & \num{0.8989} & \num{8.85e-2} & \num{2.72e-3} & \num{0.8855} \\
            FlatFault\_A & \num{1.92e-2} & \num{1.11e-3} & \num{0.9932} & \num{1.66e-2} & \num{8.21e-4} & \num{0.9939} \\
            FlatFault\_B & \num{4.39e-2} & \num{7.60e-3} & \num{0.9155} & \num{3.93e-2} & \num{6.91e-3} & \num{0.9321} \\
            CurveFault\_A & \num{2.25e-2} & \num{1.53e-3} & \num{0.9887} & \num{1.94e-2} & \num{1.18e-3} & \num{0.9899} \\
            CurveFault\_B & \num{8.65e-2} & \num{2.35e-2} & \num{0.7984} & \num{9.07e-2} & \num{2.63e-2} & \num{0.8233} \\
            Style\_A & \num{4.72e-2} & \num{6.97e-3} & \num{0.9319} & \num{3.83e-2} & \num{4.27e-3} & \num{0.9515} \\
            Style\_B & \num{4.69e-2} & \num{5.62e-3} & \num{0.8415} & \num{3.95e-2} & \num{4.11e-3} & \num{0.8638} \\
        
            \bottomrule
        \end{tabular}
    }
    \caption{
        c\(\text{I}^2\text{SB}\) inversion results for out-of-distribution initial guesses. 
    }
    \label{tab:ci2sb-generalization-ood}
\end{table}

\section{Conclusion}

We develop a variation to the recently proposed data-driven approach to acoustic waveform inversion with diffusion 
models. The theoretical backbone of our method is a special case of conditional Schrödinger Bridge problem which 
admits reformulation that makes the training process tractable. 
Our method demonstrates significant improvements over the previous results on the subject, 
achieving comparable performance to supervised-learning methods with only a few NFEs while producing samples that better 
align with human perceptual quality. Furthermore, the proposed training procedure enables an additional degree of control
over output variance through the guidance scale hyperparameter.

\printbibliography

\newpage

\section*{Supplementary figures}

\begin{figure}[h!]

    \begin{minipage}[t]{\textwidth} 
        \begin{minipage}[b]{\textwidth}
            \centering
            \includegraphics[width=0.8\textwidth]{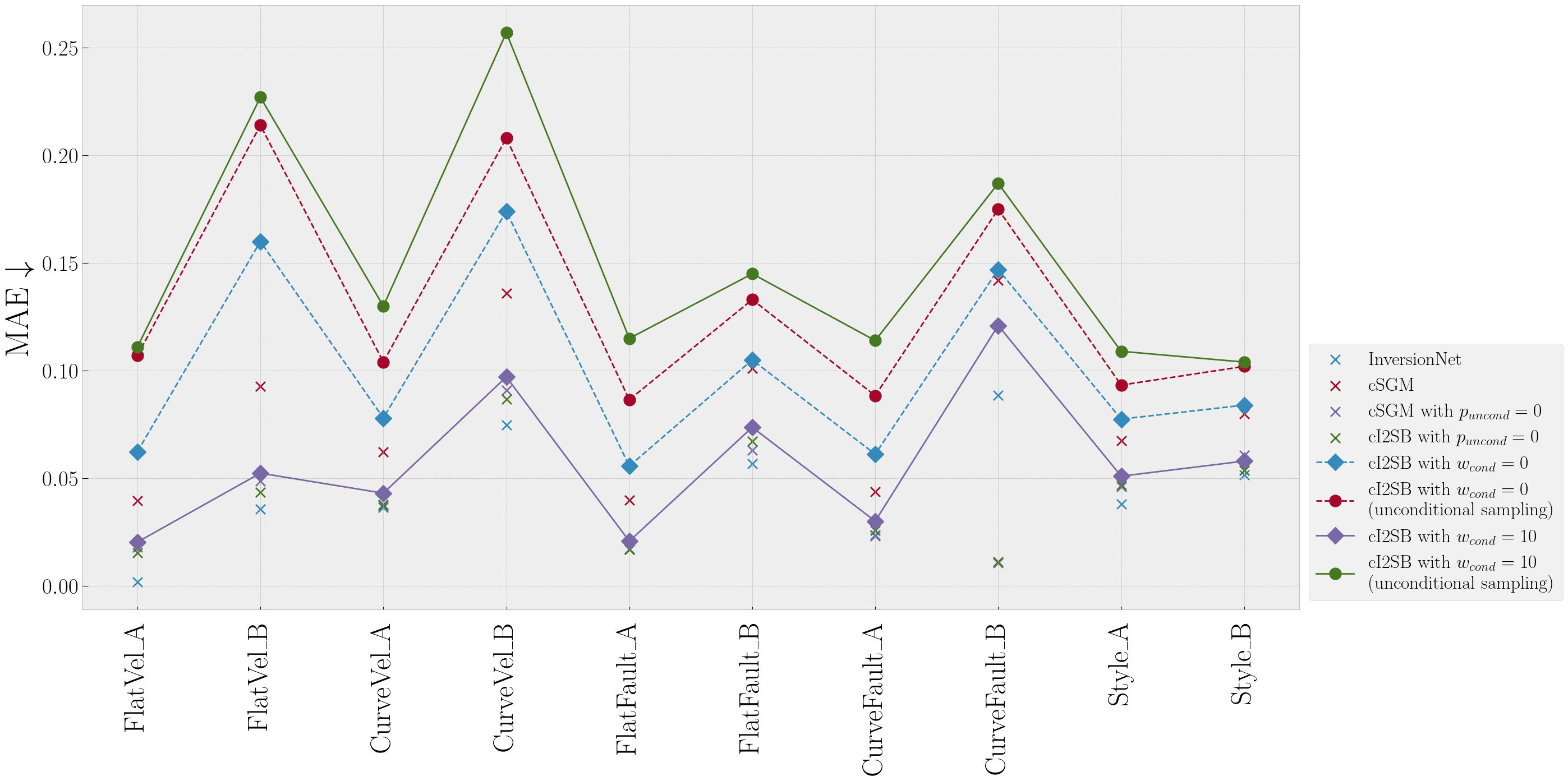}
            \subcaption{}
            \label{fig:impact_of_objective_reweighting_mae}
        \end{minipage}
        \begin{minipage}[b]{\textwidth}
            \centering
            \includegraphics[width=0.8\textwidth]{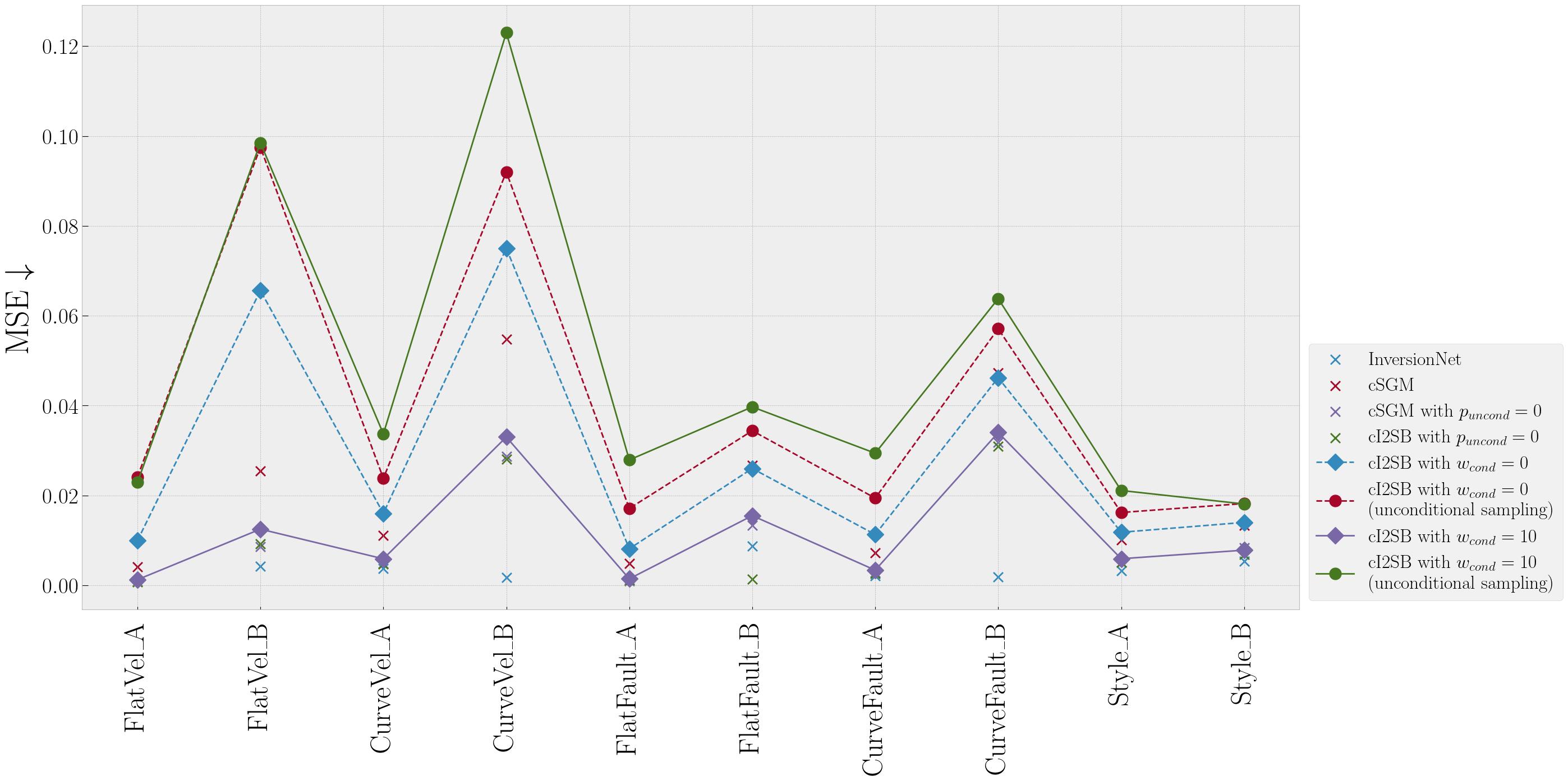}
            \subcaption{}
            \label{fig:impact_of_objective_reweighting_mse}
        \end{minipage}
        \begin{minipage}[b]{\textwidth}
            \centering
            \includegraphics[width=0.8\textwidth]{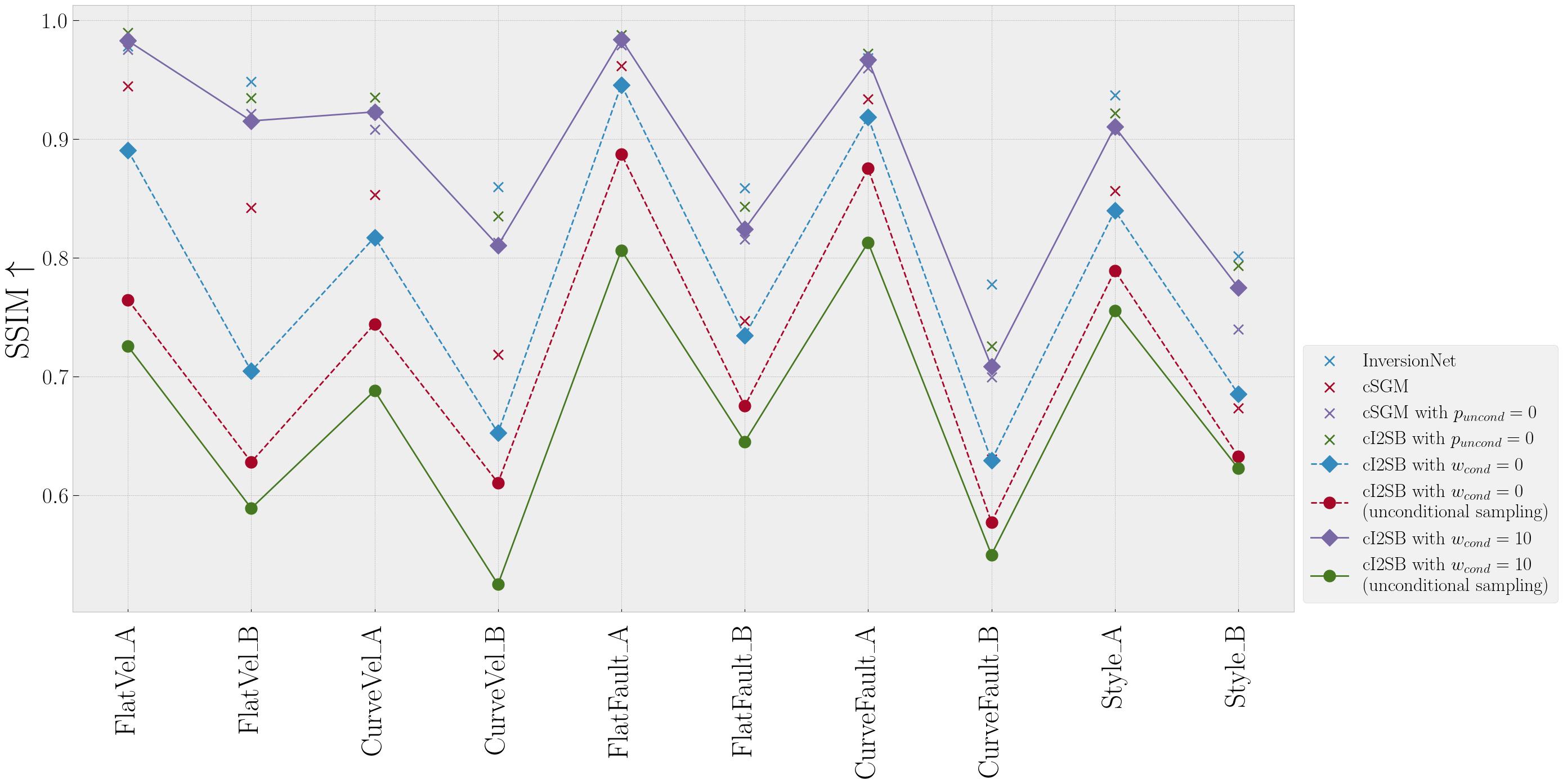}
            \subcaption{}
            \label{fig:impact_of_objective_reweighting_ssim}
        \end{minipage}
    \caption{
        Performance comparison for introduced frameworks with different sets of hyperparameters. 
        Setting a higher value of loss weighting factor $w$ for c\(\text{I}^2\text{SB}\) 
        is beneficial for inference in purely conditional mode 
        (solid line with diamond markers for $w_{\text{cond}}=10$, 
        dashed line with diamond markers for $w_{\text{cond}}=0$)
    }

    \label{fig:impact_of_objective_reweighting}
    \end{minipage}
    
\end{figure}

\newpage 

\begin{figure}[h!]

    \centering

    \begin{minipage}[b]{0.7\textwidth}
        \centering
        \includegraphics[width=\textwidth]{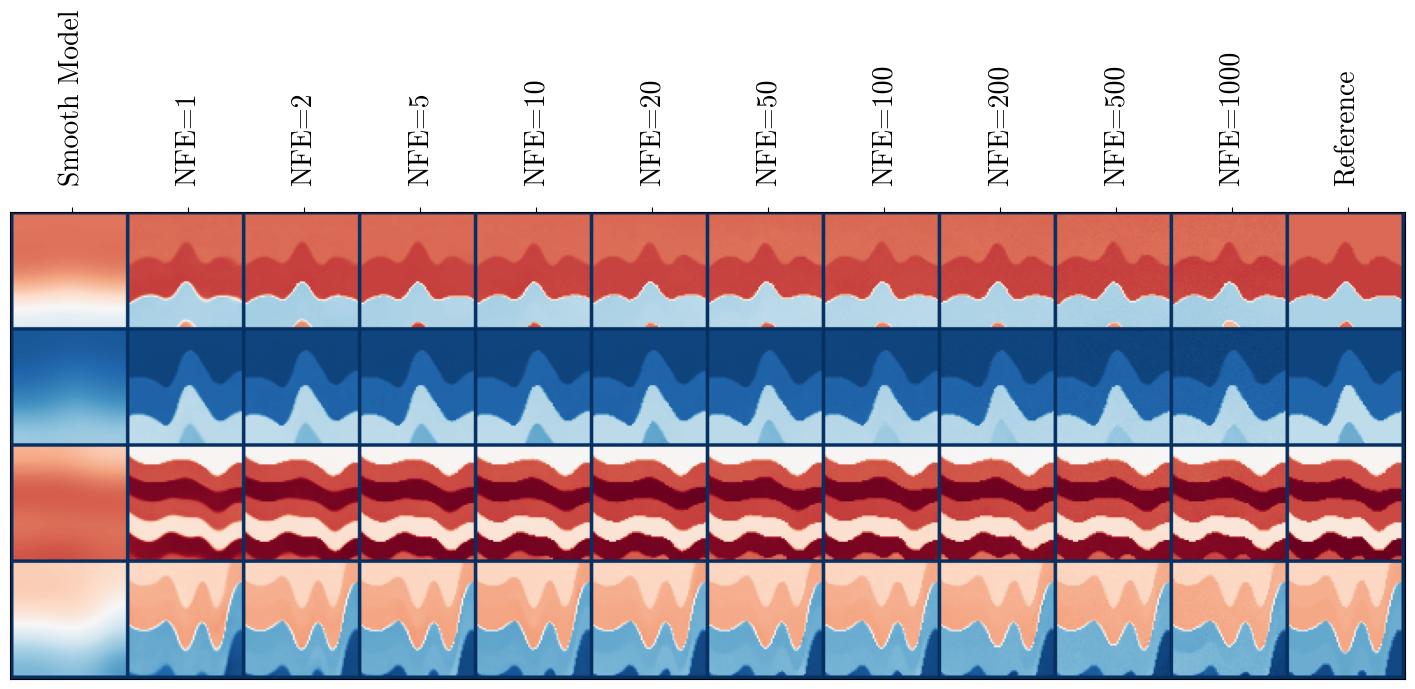}
        \subcaption{CurveVel\_B}
        \label{fig:nfe_grid_curvevel_b}
    \end{minipage}
    
    \medskip 
    
    \begin{minipage}[b]{0.7\textwidth}
        \centering
        \includegraphics[width=\textwidth]{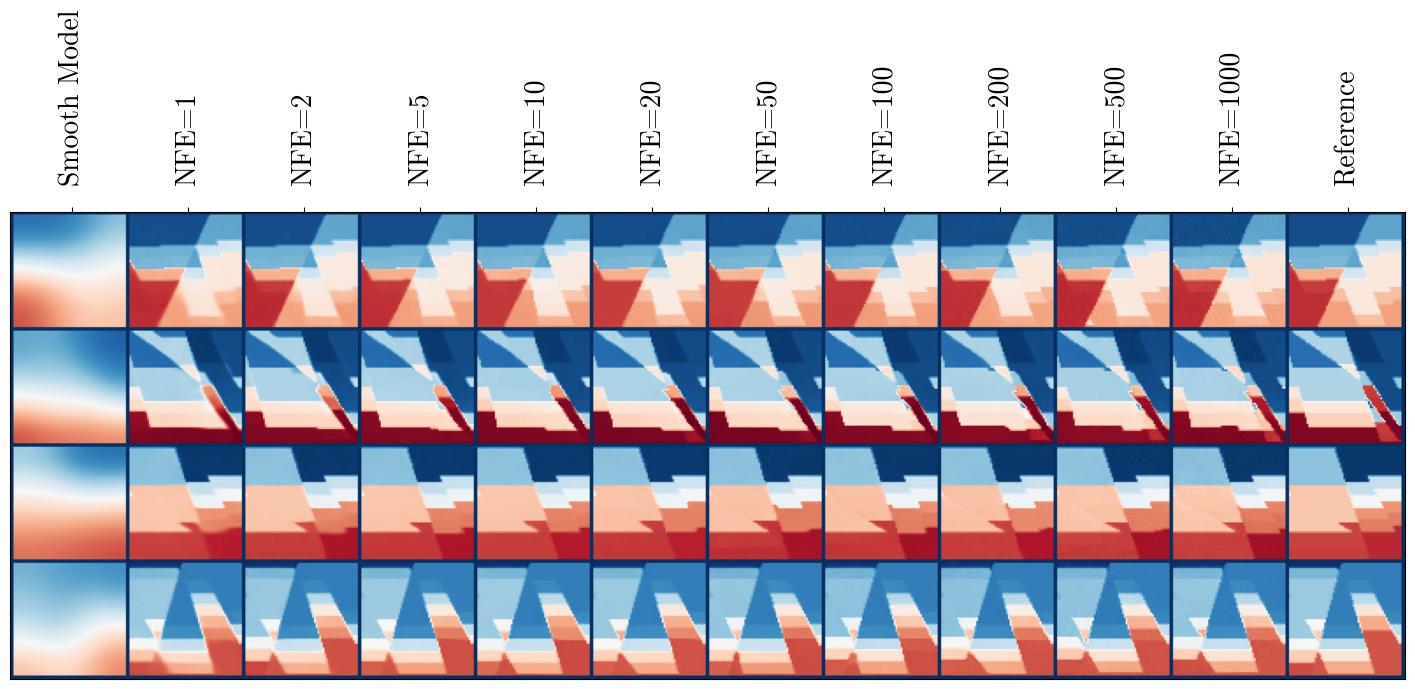}
        \subcaption{FlatFault\_B}
        \label{fig:nfe_grid_flatfault_b}
    \end{minipage}

    \medskip

    \begin{minipage}[b]{0.7\textwidth}
        \centering
        \includegraphics[width=\textwidth]{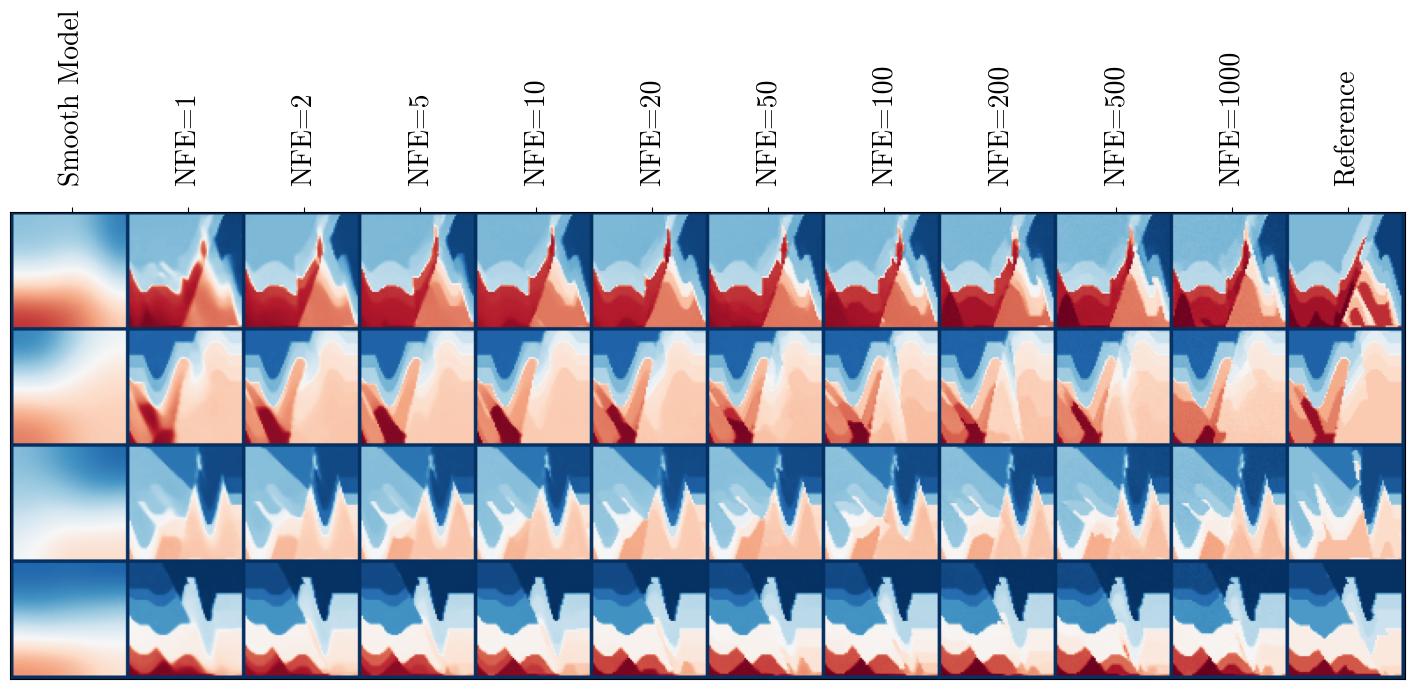}
        \subcaption{CurveFault\_B}
        \label{fig:nfe_grid_curvefault_b}
    \end{minipage}
    
    \medskip 
    
    \begin{minipage}[b]{0.7\textwidth}
        \centering
        \includegraphics[width=\textwidth]{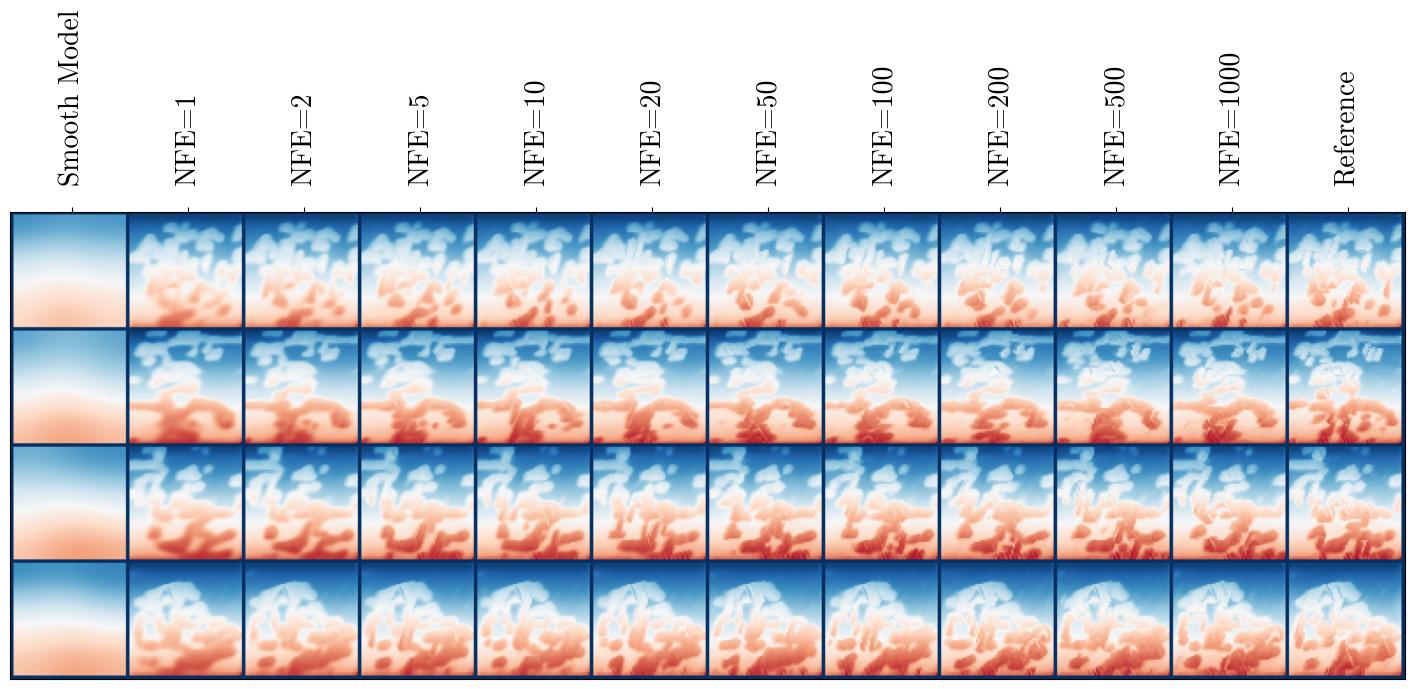}
        \subcaption{Style\_B}
        \label{fig:nfe_grid_style_b}
    \end{minipage}

    \caption{Velocity models reconstructed with c\(\text{I}^2\text{SB}\) for varying number of NFEs}
    \label{fig:i2sb_nfe_grid_datasets}

\end{figure}

\newpage

\begin{figure}[p]

    \begin{minipage}[b]{0.45\textwidth}
        \centering
        \includegraphics[width=\textwidth]{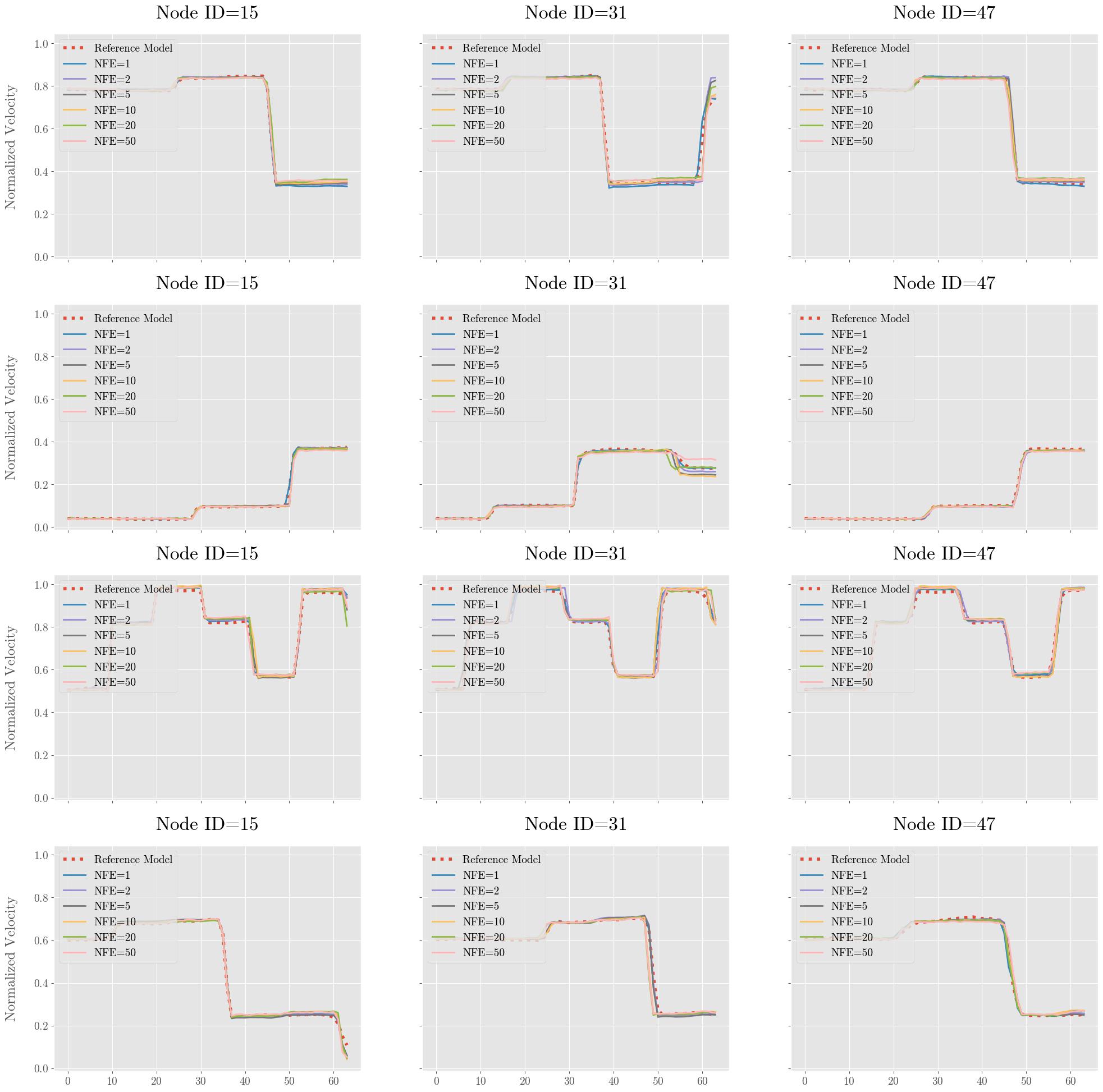}
        \subcaption{CurveVel\_B}
        \label{fig:nfe_grid_slices_curvevel_b}
    \end{minipage}
    \hfill 
    \begin{minipage}[b]{0.45\textwidth}
        \centering
        \includegraphics[width=\textwidth]{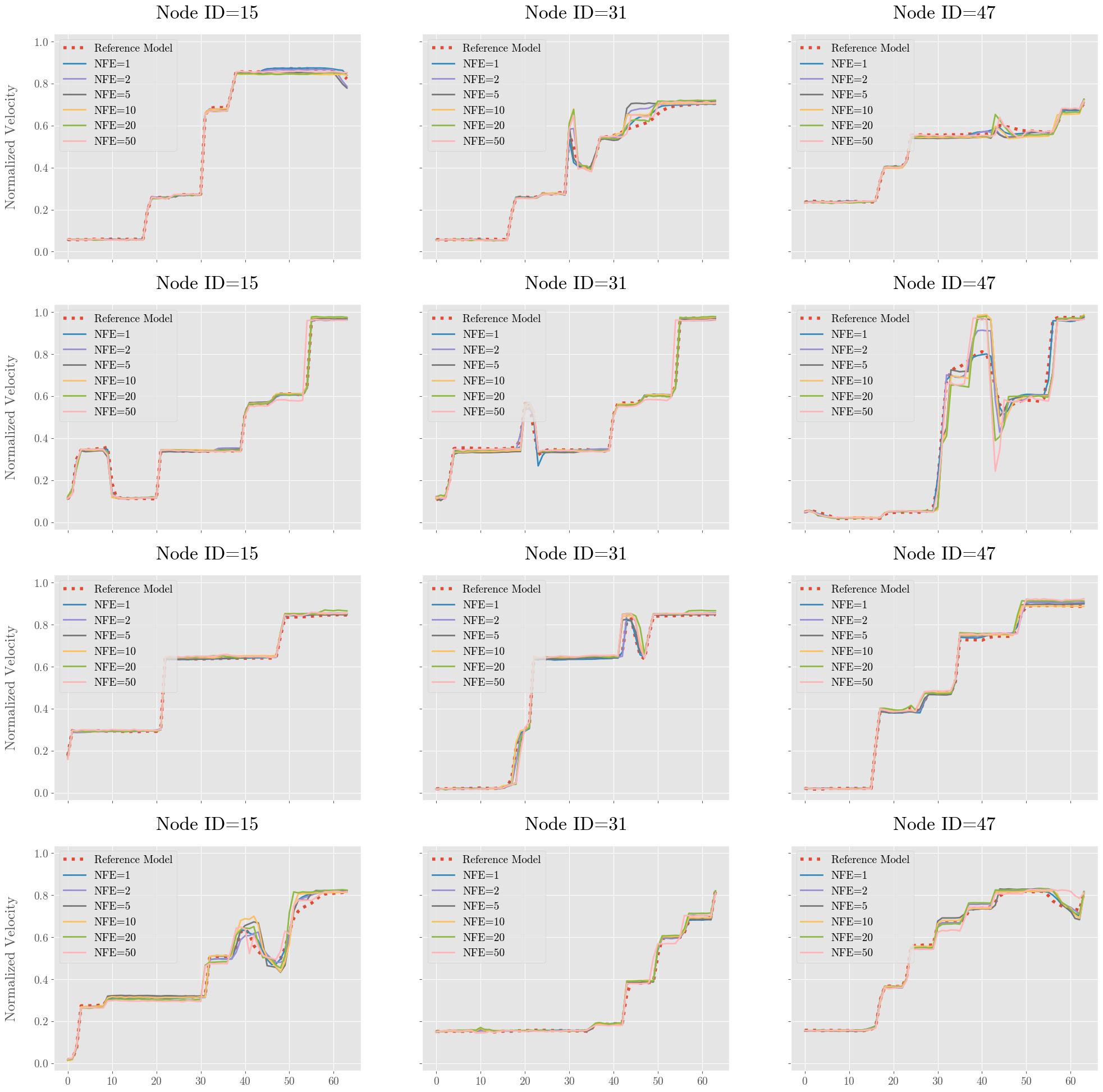}
        \subcaption{FlatFault\_B}
        \label{fig:nfe_grid_slices_flatfault_b}
    \end{minipage}

    \medskip

    \begin{minipage}[b]{0.45\textwidth}
        \centering
        \includegraphics[width=\textwidth]{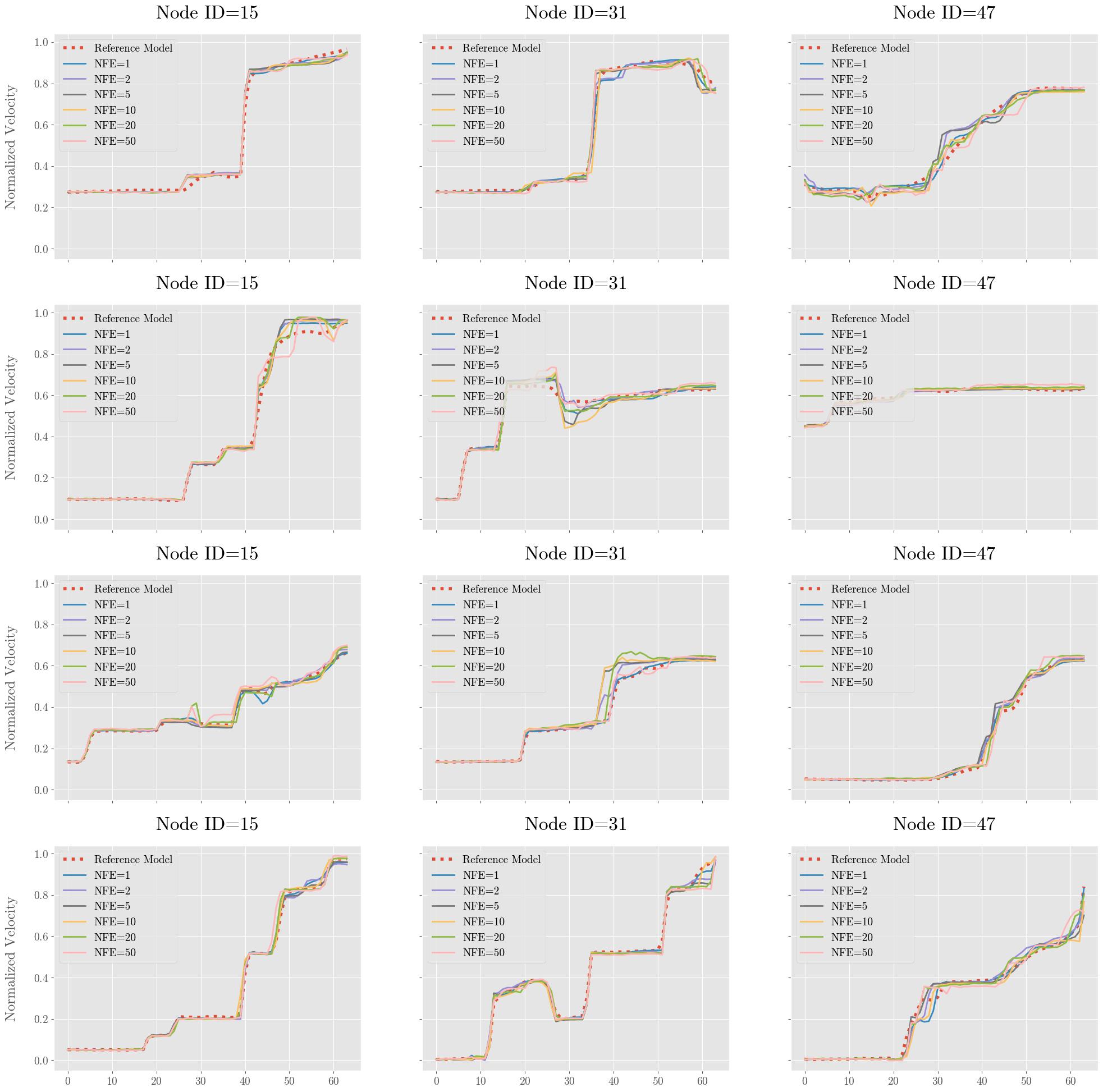}
        \subcaption{CurveFault\_B}
        \label{fig:nfe_grid_slices_curvefault_b}
    \end{minipage}
    \hfill 
    \begin{minipage}[b]{0.45\textwidth}
        \centering
        \includegraphics[width=\textwidth]{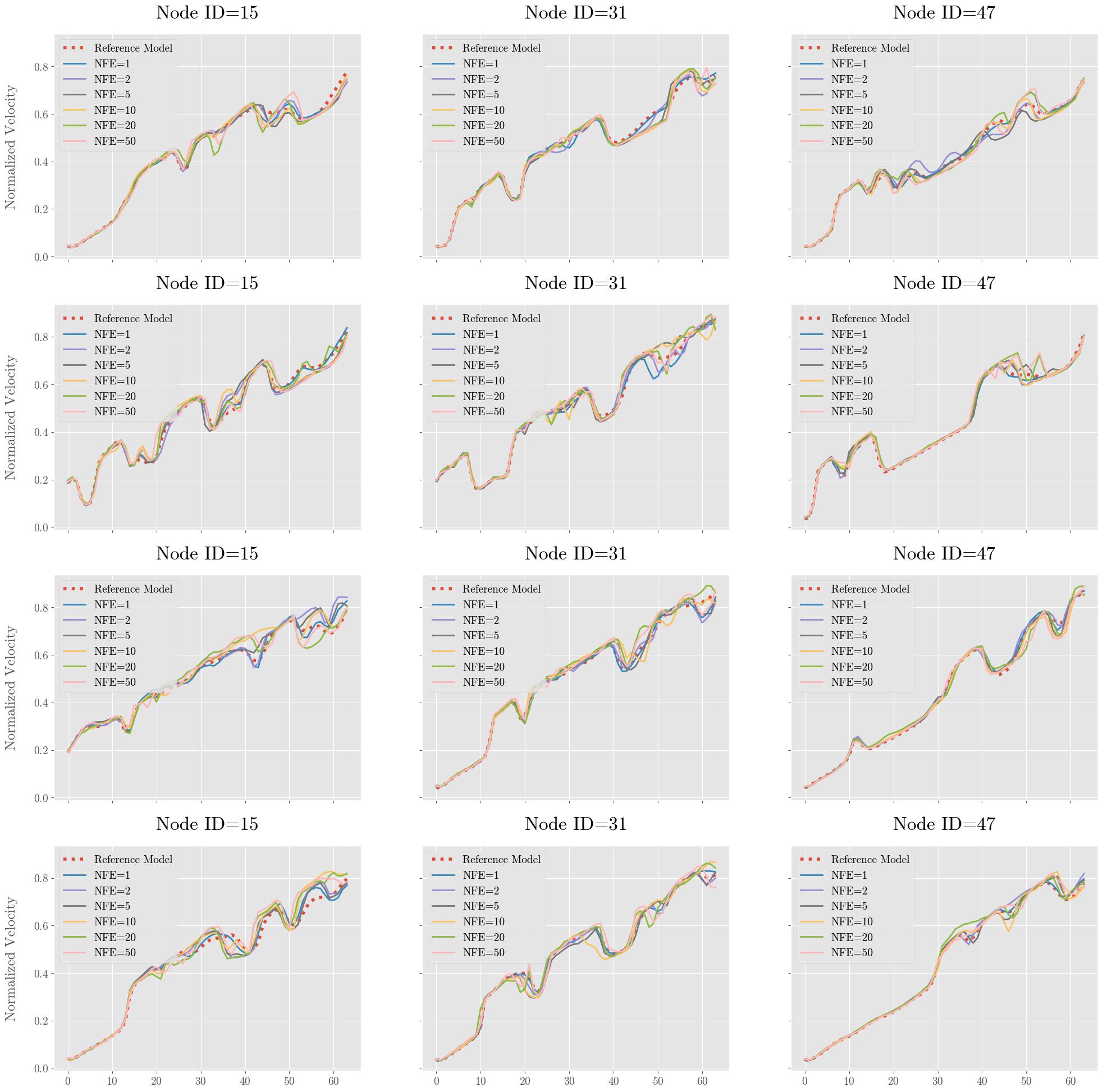}
        \subcaption{Style\_B}
        \label{fig:nfe_grid_slices_style_b}
    \end{minipage}

    \caption{Reconstructed velocity model slices for varying number of NFEs}
    \label{fig:i2sb_nfe_grid_slices}

\end{figure}

\begin{figure}[p]

    \centering
    \includegraphics[width=\textwidth]{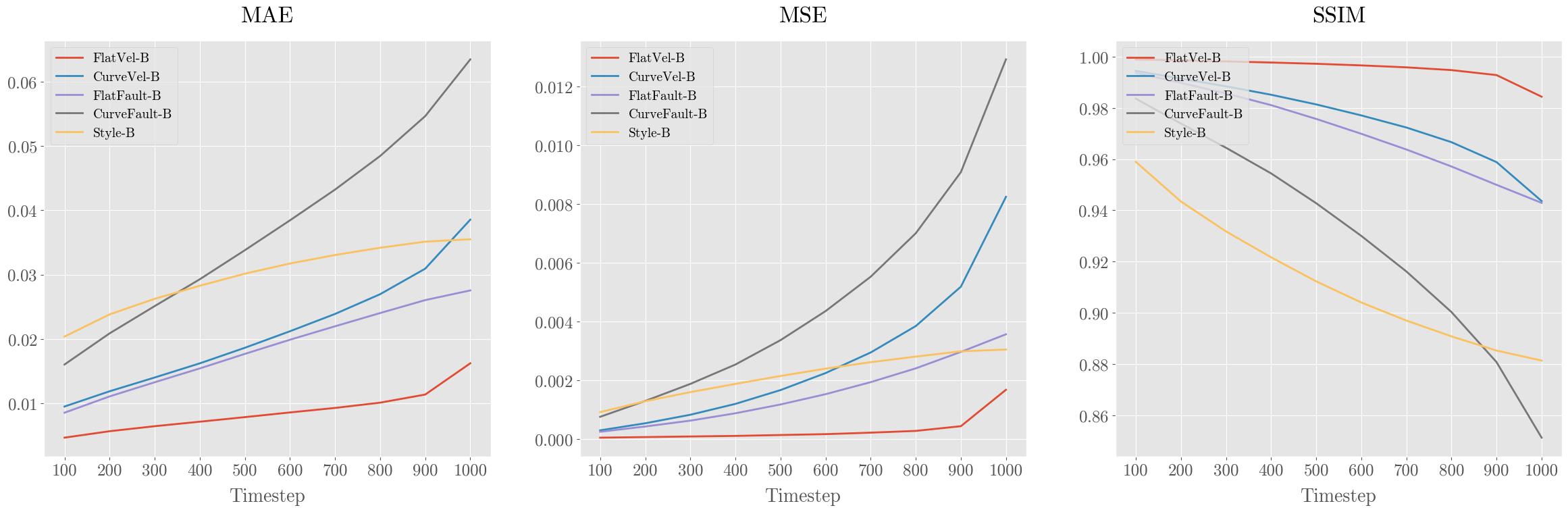}
    \caption{
        Dependency of validation metrics from timestep $n$ for single-point estimates with \textbf{ci2sb-seismic-large} 
        model given the point $\vect{c}_n$ from bridge \eqref{eq:sb_posterior_analytical_form} and timestep value.
    }
    \label{fig:i2sb_large_single_shot_evaluations_different_timesteps}

\end{figure}

\newpage

\begin{figure}[p]

    \centering

    \begin{minipage}[b]{0.7\textwidth}
        \centering
        \includegraphics[width=\textwidth]{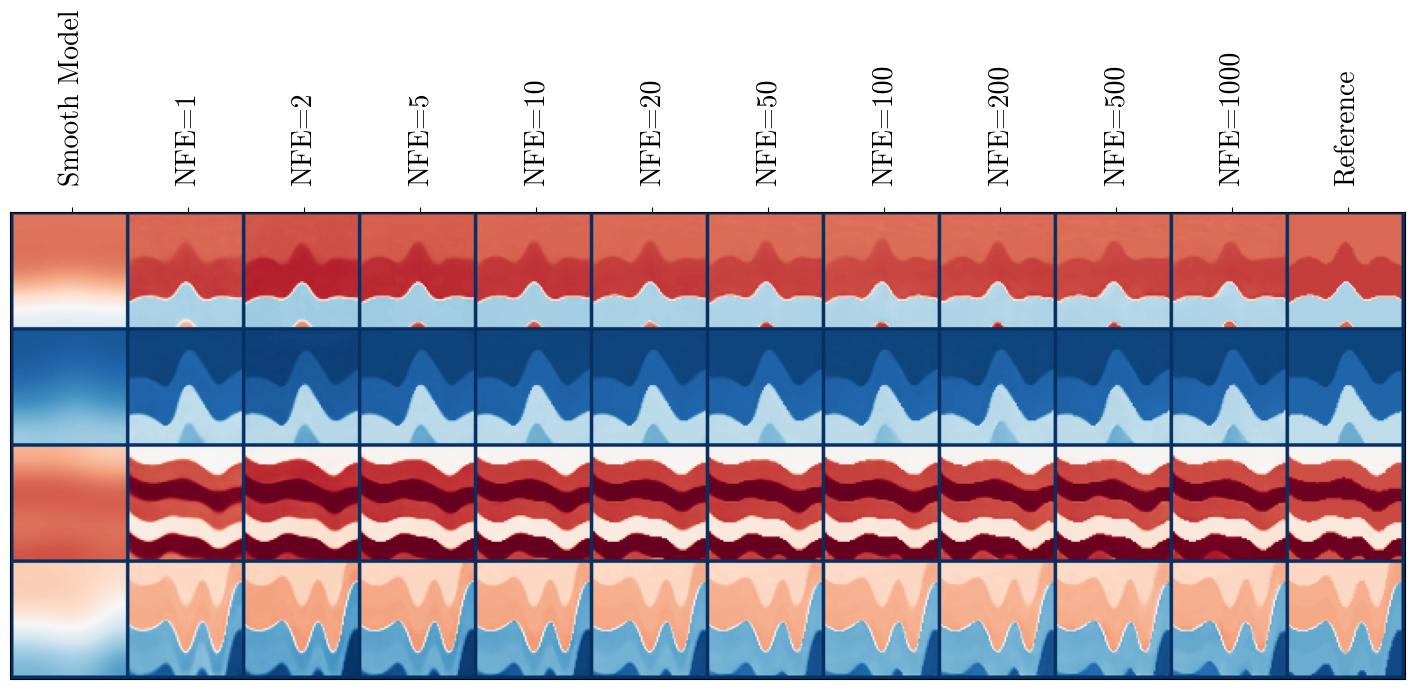}
        \subcaption{CurveVel\_B}
        \label{fig:nfe_grid_ddpm_curvevel_b}
    \end{minipage}
    
    \medskip
   
    \begin{minipage}[b]{0.7\textwidth}
        \centering
        \includegraphics[width=\textwidth]{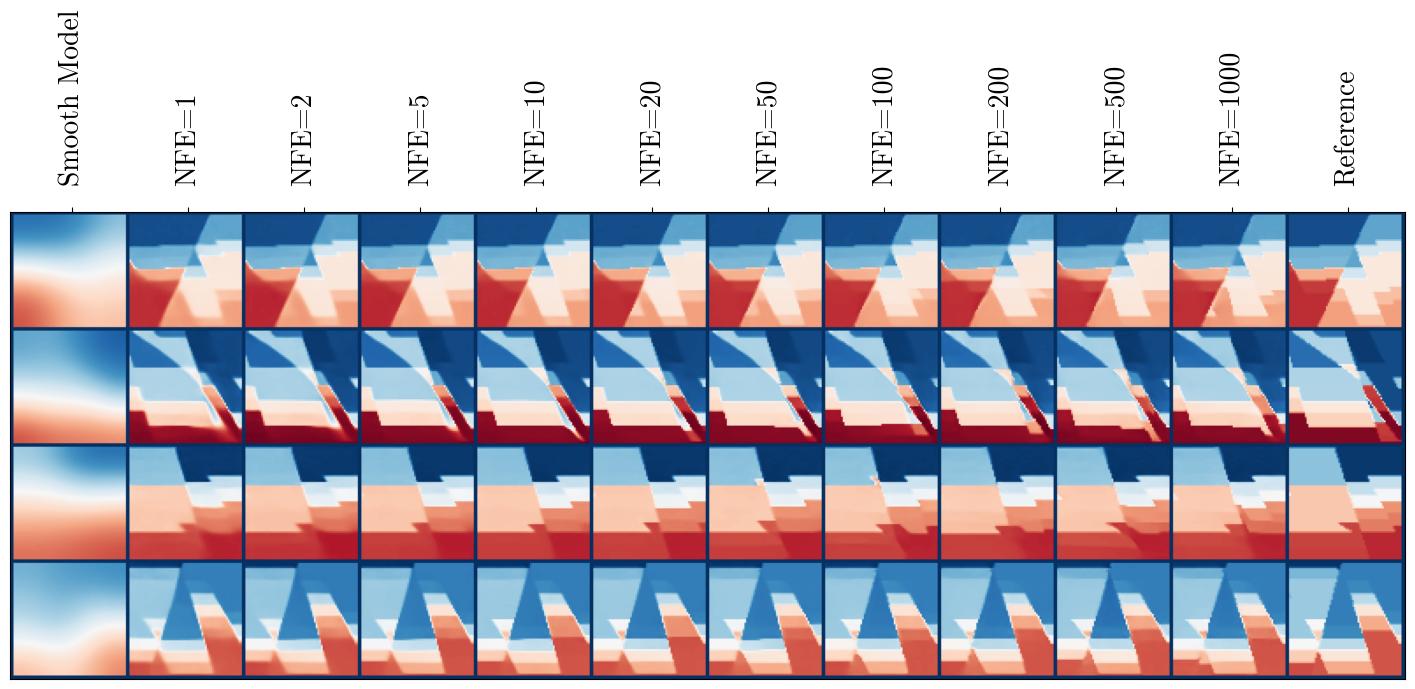}
        \subcaption{FlatFault\_B}
        \label{fig:nfe_grid_ddpm_flatfault_b}
    \end{minipage}

    \medskip

    \begin{minipage}[b]{0.7\textwidth}
        \centering
        \includegraphics[width=\textwidth]{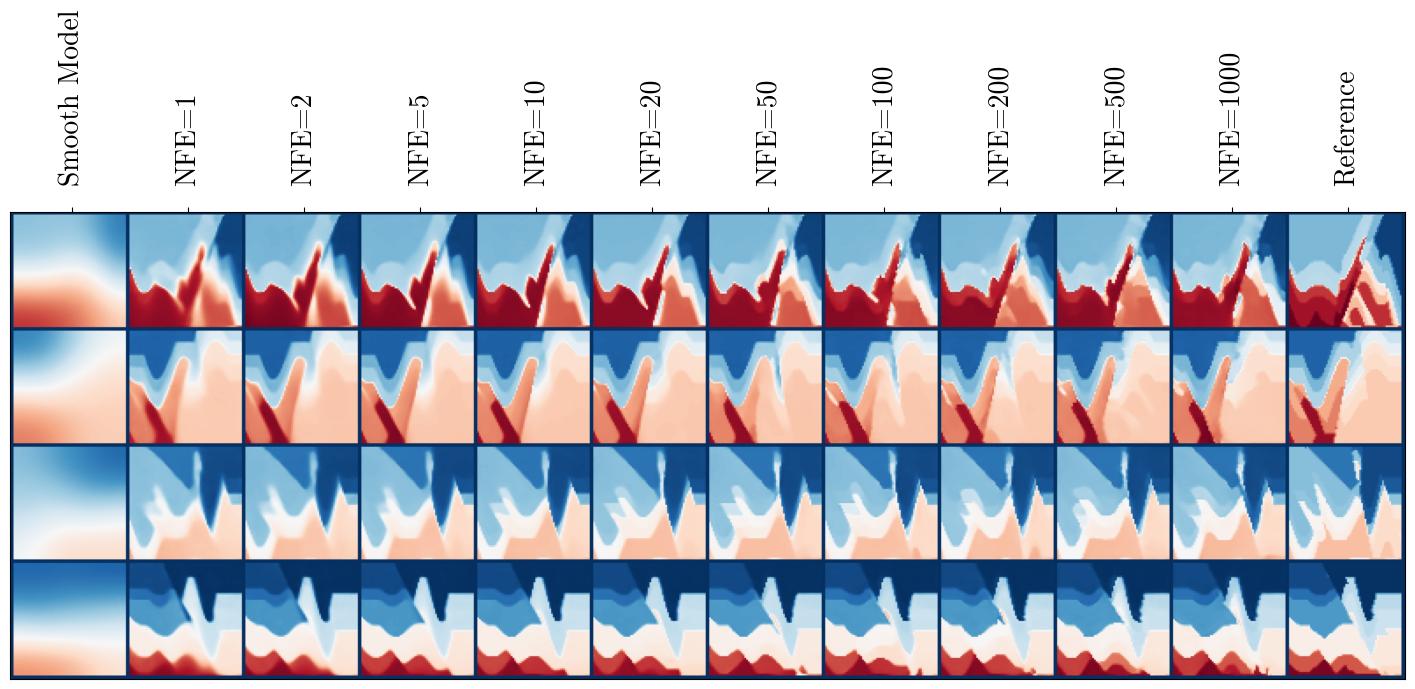}
        \subcaption{CurveFault\_B}
        \label{fig:nfe_grid_ddpm_curvefault_b}
    \end{minipage}
    \hfill 
    \begin{minipage}[b]{0.7\textwidth}
        \centering
        \includegraphics[width=\textwidth]{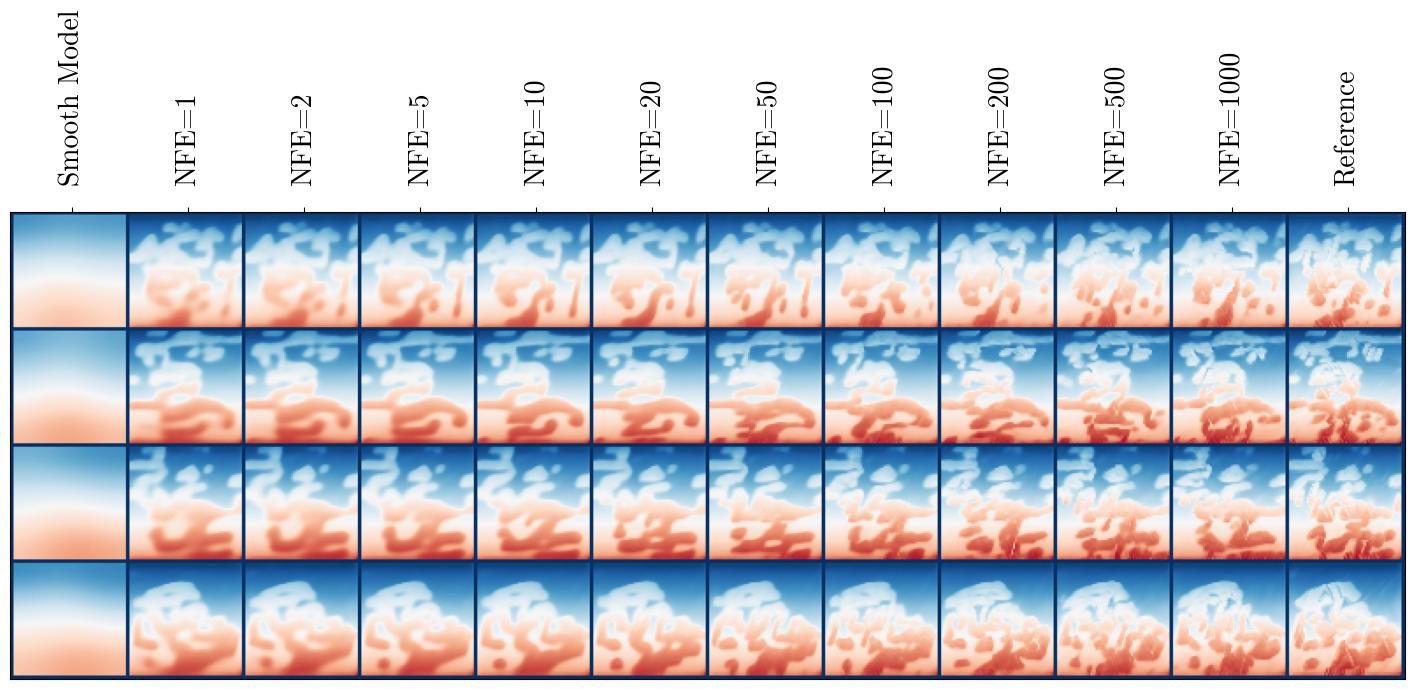}
        \subcaption{Style\_B}
        \label{fig:nfe_grid_ddpm_style_b}
    \end{minipage}

    \caption{Velocity model reconstructed with cSGM for varying number of NFEs}
    \label{fig:ddpm_nfe_grid}

\end{figure}

\newpage

\begin{figure}[h!]
    
    \centering

    \begin{minipage}[b]{0.5\textwidth}
        \centering
        \includegraphics[width=\textwidth]{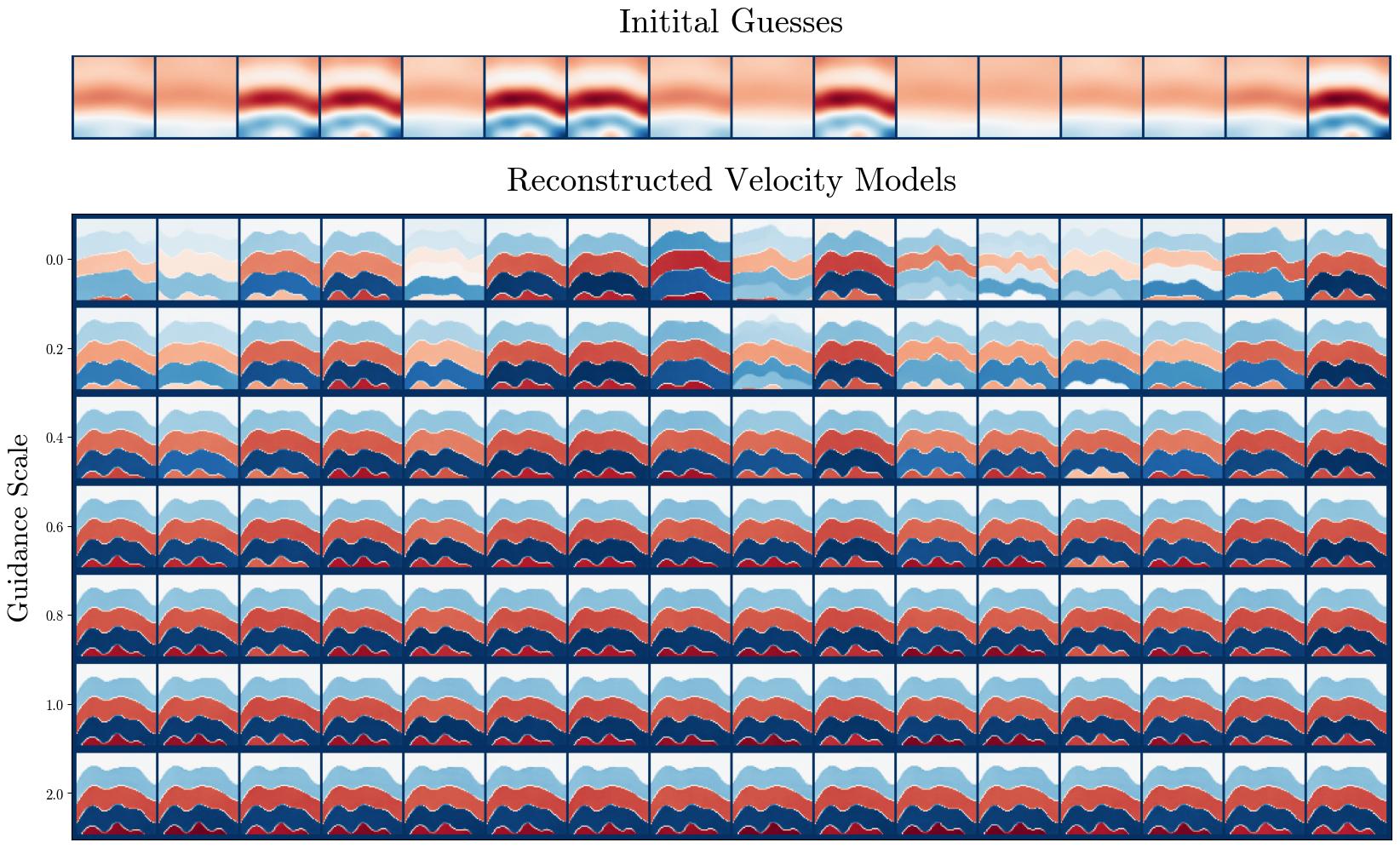}
        \subcaption{CurveVel\_B}
        \label{fig:guidance_scale_grid_curvevel_b}
    \end{minipage}
    
    \medskip
    
    \begin{minipage}[b]{0.5\textwidth}
        \centering
        \includegraphics[width=\textwidth]{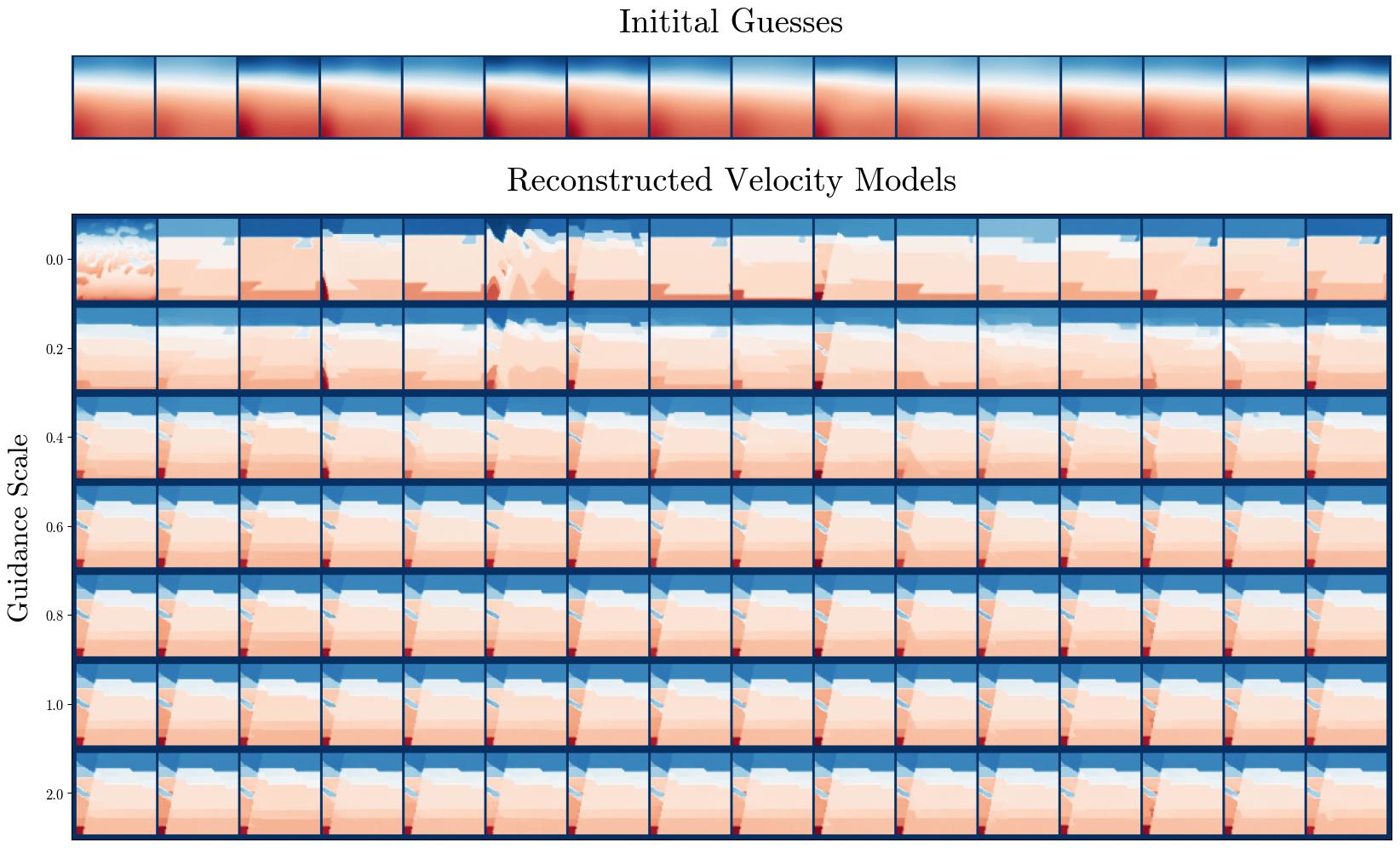}
        \subcaption{FlatFault\_B}
        \label{fig:guidance_scale_grid_flatfault_b}
    \end{minipage}
    
    \medskip
    
    \begin{minipage}[b]{0.5\textwidth}
        \centering
        \includegraphics[width=\textwidth]{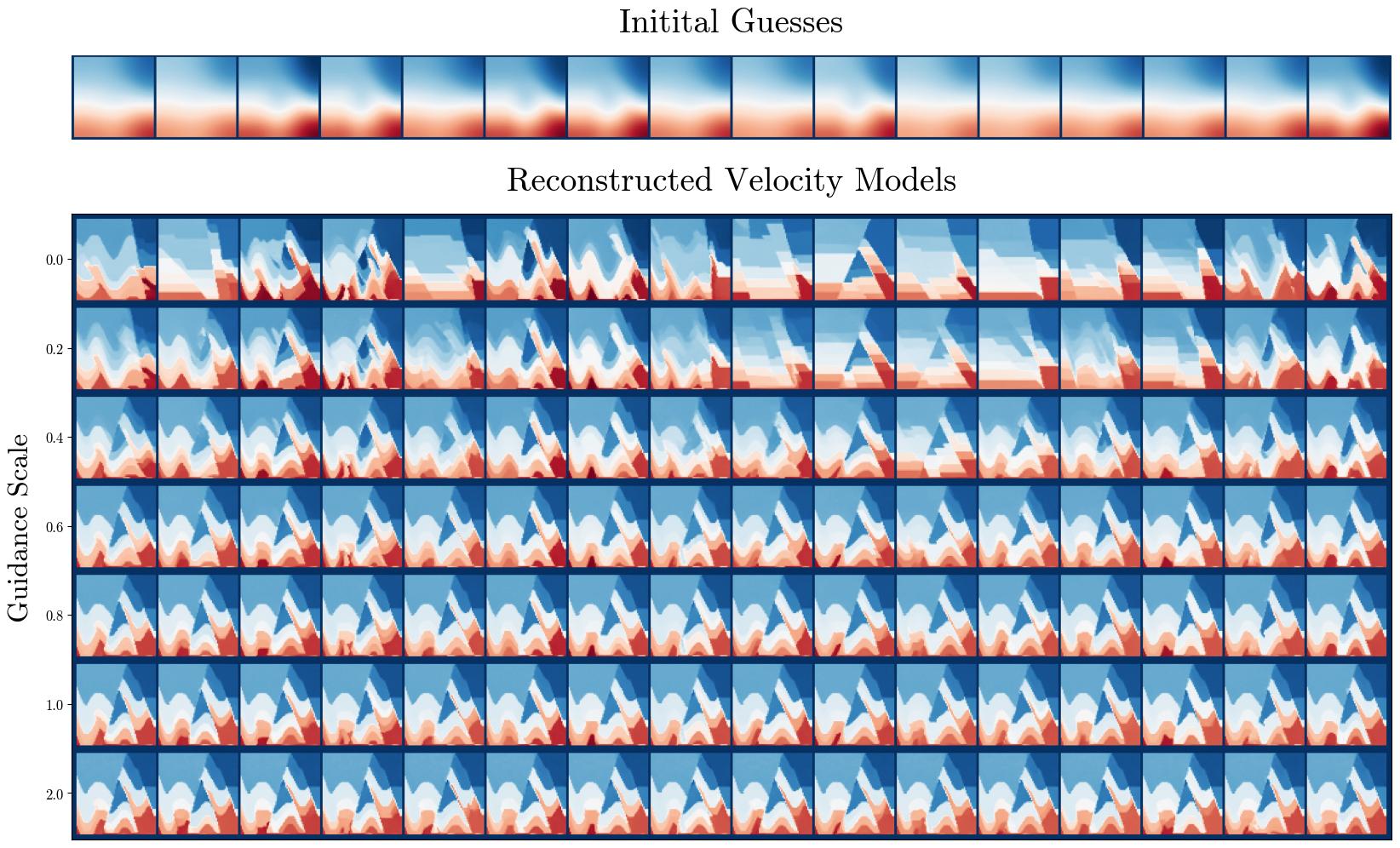}
        \subcaption{CurveFault\_B}
        \label{fig:guidance_scale_grid_curvefault_b}
    \end{minipage}
    
    \medskip
    
    \begin{minipage}[b]{0.5\textwidth}
        \centering
        \includegraphics[width=\textwidth]{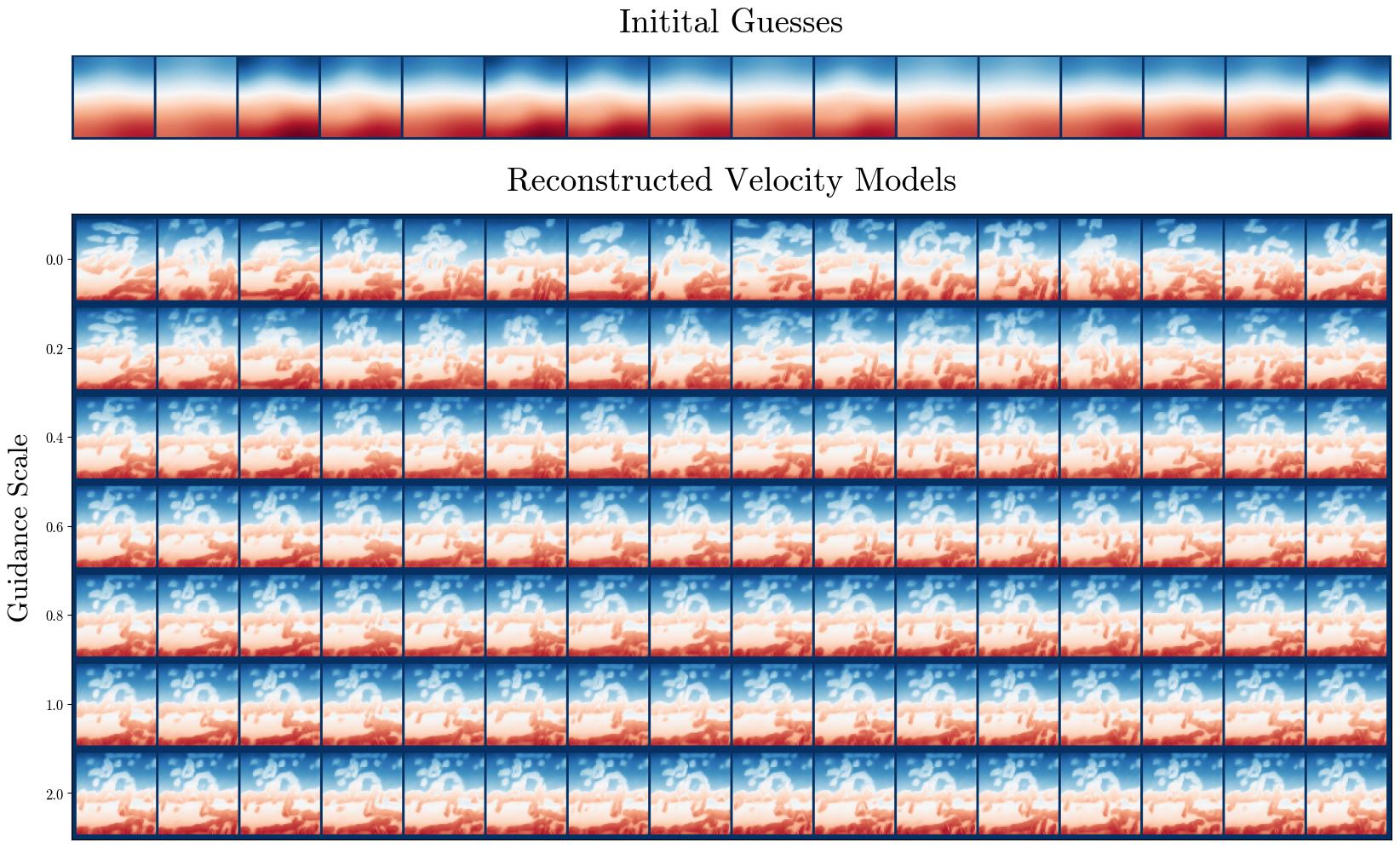}
        \subcaption{Style\_B}
        \label{fig:guidance_scale_grid_style_b}
    \end{minipage}
    \caption{Velocity models reconstructed with c\(\text{I}^2\text{SB}\) for varying values of guidance scale}
    \label{fig:guidance_scale_grids}

\end{figure}

\end{document}